\documentclass[10pt,journal,compsoc]{IEEEtran}

\usepackage{cite}
\usepackage{amsmath,amssymb,amsfonts}
\usepackage{algorithmic}
\usepackage{graphicx}
\usepackage{hyperref}
\usepackage{amsthm}
\usepackage{tabularx}

\usepackage{tabu}
\usepackage{rotating}
\usepackage{array}
\usepackage{multirow}
\usepackage{booktabs}
\usepackage{mathrsfs}
\usepackage{ragged2e}
\usepackage{enumitem}
\usepackage{orcidlink}
\usepackage[ruled,linesnumbered,vlined]{algorithm2e}
\usepackage{verbatim}
\usepackage[labelfont=bf,textfont={bf}]{caption}
\usepackage{xcolor}
\newtheorem{define}{Definition}[section]

\newtheorem{proposition}{Proposition}[section]

\newcommand*{\rowstyle}[1]{
  \gdef\@rowstyle{#1}%
  \@rowstyle\ignorespaces%
}

\definecolor{red}{rgb}{0,0,0}
\definecolor{red1}{rgb}{0,0,0}
\definecolor{red2}{rgb}{0,0,0}

\usepackage{fancyhdr}
\usepackage[english]{babel}

\usepackage{lettrine}
\usepackage{color}
\usepackage{tabularx}
\usepackage{rotating}
\definecolor{Gray}{gray}{0.95}
\usepackage{colortbl}
\usepackage{threeparttable}
\usepackage{subfig}
\usepackage{pifont}
\usepackage{makecell}
\usepackage{float}
\usepackage{tikz}
\usetikzlibrary{bayesnet}
\usetikzlibrary{arrows}

\newcommand\copyrighttext{%
  \footnotesize \textcopyright 2025 IEEE. Personal use of this material is permitted. Permission from IEEE must be obtained for all other uses, in any current or future media, including reprinting/republishing this material for advertising or promotional purposes, creating new collective works, for resale or redistribution to servers or lists, or reuse of any copyrighted component of this work in other works. 
  DOI: 10.1109/TPAMI.2025.3626068}
\newcommand\copyrightnotice{%
\begin{tikzpicture}[remember picture,overlay]
\node[anchor=south,yshift=5pt] at (current page.south) {\fbox{\parbox{\dimexpr\textwidth-\fboxsep-\fboxrule\relax}{\copyrighttext}}};
\end{tikzpicture}%
}

\makeatletter
\def\hlinewd#1{%
  \noalign{\ifnum0=`}\fi\hrule \@height #1 \futurelet
   \reserved@a\@xhline}
\makeatother

\begin{document}

\newcommand{\THEMODEL}{DynamicPAE}

\title{\THEMODEL: Generating Scene-Aware Physical Adversarial Examples in Real-Time}

\author{~Jin Hu~\orcidlink{0009-0000-5564-9331},
        Xianglong Liu~\orcidlink{0000-0002-7618-3275}, \IEEEmembership{Senior Member, IEEE},
        Jiakai Wang \orcidlink{0000-0001-5884-3412}, 
        Junkai Zhang \orcidlink{0009-0005-8840-234X}, \\
        Xianqi Yang,
        Haotong Qin \orcidlink{0000-0001-7391-7539},
        Yuqing Ma \orcidlink{0000-0003-1936-9396},
        and Ke Xu
\IEEEcompsocitemizethanks{
\IEEEcompsocthanksitem Jin Hu, Xianglong Liu, Junkai Zhang, Yuqing Ma, and Ke Xu are with State Key Laboratory of Complex \& Critical Software Environment, Beihang University, Beijing, China. E-mail: \{hujin, xlliu, zhangjunkai, mayuqing, kexu\}@buaa.edu.cn. Jin Hu, Xianglong Liu, and Ke Xu are also with Zhongguancun Lab.
\IEEEcompsocthanksitem Jiakai Wang and Xianqi Yang are with Zhongguancun Lab, Beijing, China. E-mail: \{wangjk, yangxq\}@mail.zgclab.edu.cn.
\IEEEcompsocthanksitem Haotong Qin is with ETH Zurich, E-mail: haotong.qin@pbl.ee.ethz.ch
\\
}
\thanks{Manuscript received Oct. 10, 2024 (Corresponding author: Xianglong Liu.)}
}

\markboth{Journal of IEEE Transactions on Pattern Analysis and Machine Intelligence}%
{Shell \MakeLowercase{\textit{et al.}}: Bare Demo of IEEEtran.cls for Computer Society Journals}


\IEEEtitleabstractindextext{%
\begin{abstract}
\justifying
Physical adversarial examples (PAEs) are regarded as ``whistle-blowers'' of real-world risks in deep-learning applications, thus worth further investigation.
However, current PAE generation studies show limited adaptive attacking ability to diverse and varying scenes, revealing the \textcolor{red}{urgent requirement of dynamic PAEs that are generated in real time and conditioned on the observation from the attacker.}
\textcolor{red}{
The key challenge in generating dynamic PAEs is learning the sparse relation between PAEs and the observation of attackers under the noisy feedback of attack training.
To address the challenge, we present DynamicPAE, the first generative framework that enables scene-aware real-time physical attacks.
Specifically, to address the noisy feedback problem that obfuscates the exploration of scene-related PAEs, we introduce the residual-guided adversarial pattern exploration technique.
We first introduce the limited feedback information restriction to model the training degeneracy problem under noisy feedback.
Then, residual-guided training, which relaxes the attack training with a reconstruction task, is proposed to enrich the feedback information, thereby achieving a more comprehensive exploration of PAEs.}
\textcolor{red}{To address the alignment problem between the trained generator, which represents the learned relation, and the real-world scenario, we introduce the distribution-matched attack scenario alignment, consisting of the conditional-uncertainty-aligned data module and the skewness-aligned objective re-weighting module.}
\textcolor{red}{The former aligns the training environment with the incomplete observation of the real-world attacker.
The latter facilitates consistent stealth control across different attack targets by balancing the objectives with the skewness indicator.}
Extensive digital and physical evaluations demonstrate the superior attack performance of DynamicPAE, attaining a \textcolor{red}{2.07}× boost (\textcolor{red}{58.8}\% average AP drop under attack) on representative object detectors (e.g., \textcolor{red}{DETR}) over state-of-the-art static PAE generating methods.
Overall, our work opens the door to end-to-end modeling of dynamic PAEs.
\end{abstract}

\begin{IEEEkeywords}
Adversarial Example, Physical Adversarial Attack, Object Detection, Dynamic Physical Adversarial Example
 \end{IEEEkeywords}
}


\maketitle

\IEEEdisplaynontitleabstractindextext

\IEEEpeerreviewmaketitle

\copyrightnotice

\IEEEraisesectionheading{\section{Introduction}\label{sec:isntro}}

\IEEEPARstart{N}{umerous} intelligent applications in real-world scenarios have been landed in recent years, such as autonomous driving~\cite{chen2023end}, healthcare~\cite{esteva2019guide} and intelligent assistant~\cite{zhao2023surveyllm}.
However, adversarial examples (AEs), which are specially designed for misleading machine learning models~\cite{szegedy2013intriguing}, have long been a challenge for deep learning applications~\cite{peck24advpami}.
Among them, the physical-world adversarial examples (PAEs) are attracting broader attention due to their feasibility in the real world and de facto threats to business AI systems~\cite{wang2021universal,wang2021dual, Wang2023ADSys,liu2023x}.
Besides revealing the risks, the research on AEs (PAEs) also deepens the understanding of deep neural networks and reveals their defects in industrial and scientific applications~\cite{ding2022survey, zhang2022experimental}. Therefore, modeling AEs, especially PAEs, is worth further \textcolor{red2}{investigation} to alleviate the trustworthy issues (\emph{e.g.}, interpretability, security, and robustness) of deep learning models and related applications.

\begin{figure}[t]
    \centering
    \includegraphics[width=\linewidth]{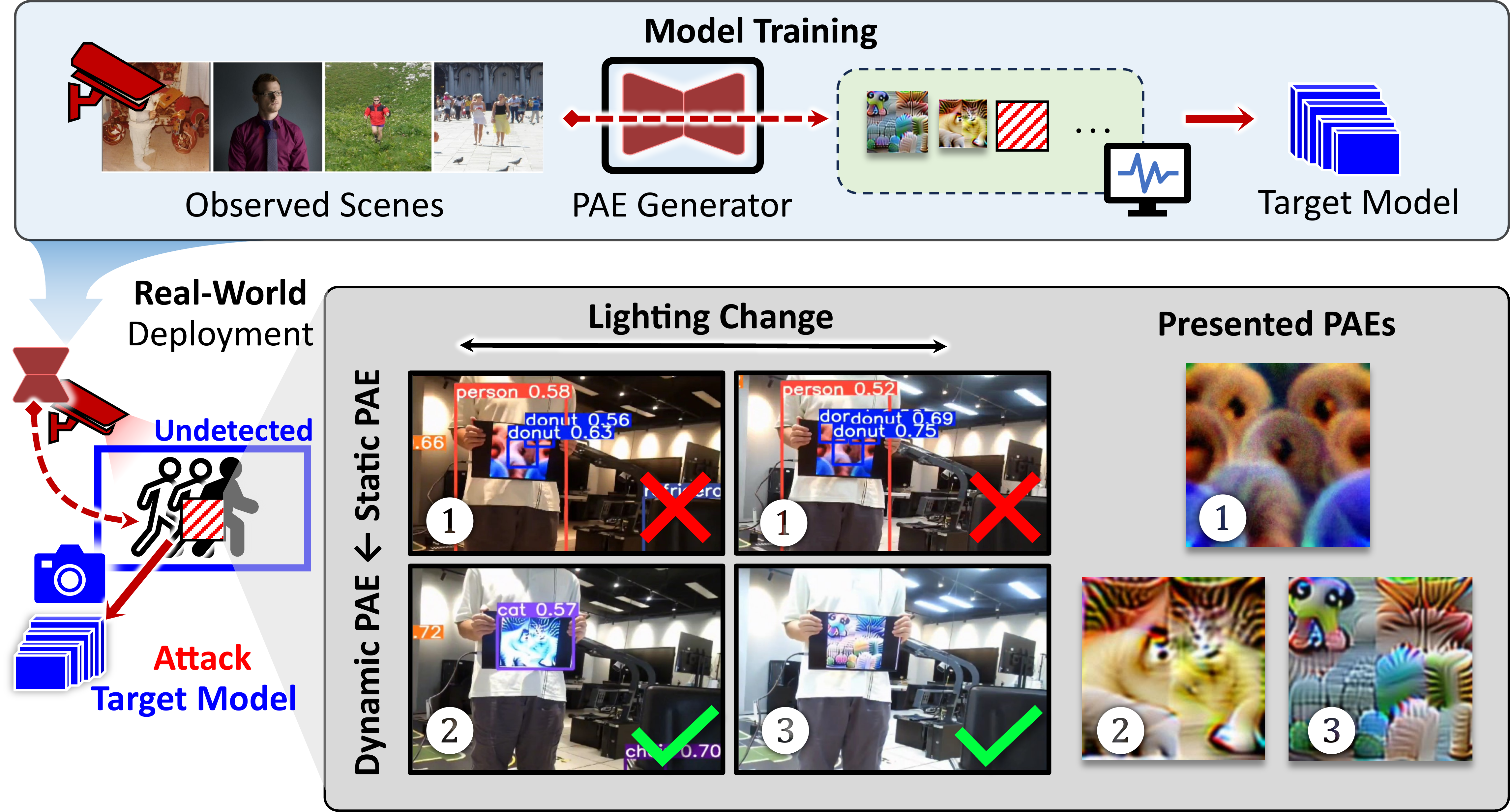}
    \caption{DynamicPAE framework trains real-time scene-aware physical adversarial example (PAE) generators and establishes a novel form of PAE, \emph{i.e.}, \textit{dynamic PAE}. \textcolor{red}{After training, the generator generates highly effective PAEs based on the characteristics of the current scene.}}
    \label{fig:intro}
    \vspace{-1em}
\end{figure}

\textcolor{red}{Generally, PAEs might target varying real-world conditions in application scenarios, such as moving objects, dynamic lighting, and ever-changing backgrounds.}
However, mainstream PAE methods address the PAE generation problem by treating it as a static issue, either attempting to generalize the PAE across all simulated physical scenes~\cite{eykholt2018robust} or requiring retraining each time to adapt to new settings~\cite{zhu2023understanding}. This approach results in inadequate adaptability or generation efficiency.
On the contrary, we define scene-aware and real-time generated PAEs as \textbf{dynamic PAEs}, a problem that has yet to be addressed in the field.
Related attempts are limited to determining patch locations only~\cite{liu2019perceptual}, simulated control of a few states~\cite{sun2024embodied}, optimization for each clustered scene in the lab~\cite{chahe2023dynamic}, and resisting dynamic fluctuations~\cite{guesmi2024dap} rather than enabling dynamic responses.

\textcolor{red}{This work focuses on constructing the dynamic PAE generating framework to establish the novel form of PAE, \emph{i.e.}, dynamic PAE.x
As shown in Figure~\ref{fig:intro}, we aim to construct the real-time dynamic PAE generator through an end-to-end training and deployment paradigm, achieving highly effective PAE generation by adapting to the current scene in real time.
This can be regarded as a new generative learning problem that takes the observation from the attacker as the input and the corresponding PAE as the output.
However, compared to conventional generative learning tasks that extract knowledge from existing data, this task requires creating unknown adversarial data that is effective in an uncertain world, and the generator shall learn the key relations between the observation of the attacker and suitable PAEs.
Such relation may be sparse, and the optimization feedback of the attack task may be noisy, which poses the following challenges:}
\ding{182} \textcolor{red}{The suitable scene-related PAEs are hard to discover, presented as the optimization degeneracy problem in training dynamic PAE generators.
The problem arises from the noisy gradient feedback~\cite{athalye2018obfuscated} in attack training that hinders the exploration of the potentially sparse distributions~\cite{ma2018characterizing} of PAEs, and the noise originates from the necessary randomness injection simulating the real-world uncertainty.}
\ding{183} \textcolor{red}{Aligning the trained generator, which represents the learned relations, with the real-world scenario is challenging. The observation of the real-world attacker is incomplete, and the behavior of victim models is inconsistent.
These characteristics can cause difficulties in balancing universality, stealth, and attack performance when adapting the generator to real-world scenarios.}

To address the dynamic PAE generation problem, we present the \THEMODEL\ framework that \textcolor{red}{guides the design and the training of the dynamic PAE generator}.
As shown in Figure~\ref{fig:intro}, the framework trains the scene-aware generator in the end-to-end paradigm and significantly improves the attack performance under varying conditions.
\textcolor{red}{Specifically, \textbf{residual-guided adversarial pattern exploration} is proposed to address the training degeneracy problem of the dynamic PAE generator.}
\textcolor{red}{We first propose the limited feedback information restriction model to analyze the training degeneracy problem and  support the construction of training techniques.}
\textcolor{red}{Inspired by deep residual learning~\cite{he2016deep}, we present the construction of the residual task that redefines the optimization problem, enriches the guidance information by relaxing the training with an auxiliary task, and leads the optimization toward \textcolor{red}{the deep exploration of the relation between PAEs and scenes}.}
\textcolor{red}{\textbf{Distribution-matched attack scenario alignment} is proposed to align the generator with the real-world scenario.}
Specifically, the conditional-uncertainty-aligned data model is proposed \textcolor{red}{to align the training environment with the incomplete observation of the attacker and therefore balance the trade-off between the universality and attack performance}. The skewness-based weight controller is proposed to align the \textcolor{red}{stealthiness of generated PAEs automatically by the novel skewness indicator, achieving the consistent and balanced generating objective control across different attack targets.}
By solving the key training and task modeling issues, the \THEMODEL\ framework bridges the gap between generative neural networks and dynamic PAE generation.

We construct a comprehensive digital benchmark on patch attacks, an example of simulated adversarial testing in the autonomous driving scenario, and a physical-world attack prototype system for object detection to evaluate the proposed framework and demonstrate its effectiveness.
Our contribution is summarized as follows:
\begin{itemize}
\item We propose the novel dynamic physical adversarial example (PAE), which is generated conditioned on the observation from the attacker in real-time. We propose the \THEMODEL\ framework that opens the end-to-end modeling of this widely impactful task.
\item We propose the \textcolor{red}{\textit{residual-guided adversarial pattern exploration} to address the training degeneracy problem and explore suitable PAEs under noisy feedback. We model the problem by the limited feedback information restriction.
We propose the residual-guided training technique that relax the attack training with a reconstruction task to enrich the feedback information.
It guides the optimization away from degenerate solutions, leading to a more comprehensive and stable exploration of feasible PAEs.}
\item We propose the \textcolor{red}{\textit{distribution-matched attack scenario alignment} to address the challenges of aligning the generator with the real-world scenario}. It consists of (1) the conditional-uncertainty-aligned data module that \textcolor{red}{models the incomplete observation of the attacker and aligns the training environment with it;} (2) the skewness-aligned objective re-weighting module that enables consistent stealthiness control across different attack targets by automatically re-weighting losses using the skewness indicator.
\item Extensive experiments across diverse settings demonstrate the superiority of \THEMODEL, \emph{i.e.}, \textcolor{red}{\textbf{2.07}}$\times$ of \textit{average AP drop after attack} compared to existing PAEs and \textbf{12ms} \textit{inference latency}. \textcolor{red}{In particular, the improvement is more significant when attacking robust models.} Evaluations on the simulation platform (\emph{i.e.}, CARLA) and physical-world-deployed devices further validate the effectiveness of our proposed method.
\end{itemize}

The rest of this paper is organized as follows: Section 2 presents the background and the definition of the dynamic PAE problem. Section 3 introduces the key components of the \THEMODEL\ framework. Section 4 details the experimental evaluation, including the settings, the results, and the analysis. Section 5 discusses the related works of the dynamic attack problem and optimization techniques. \textcolor{red}{Section 6 discusses the potential application, limitation and future works, and section 7 concludes the paper.}
\section{Preliminaries}

\subsection{Backgrounds}

\textbf{Adversarial Examples (AEs)} are a type of specially designed data for fooling deep-learning models~\cite{DBLP:journals/corr/GoodfellowSS14}.
Formally, given a visual recognition model $\mathcal{F}$, a benign input $x$, and its corresponding ground truth $y$, the AE $\delta$ is designed to satisfy:
\begin{equation}
    y \neq \mathcal{F}(x \oplus \boldsymbol\delta),\quad \lVert \boldsymbol\delta\rVert _p \leq \varepsilon,
\end{equation}
where $\lVert  \cdot \rVert _p$ indicates the distance metric under the $p$-norm and could be replaced by other metrics,  $\varepsilon$ is a small constant that controls the magnitude of the adversarial attack.

For the AEs to reveal real vulnerabilities, a key issue is whether AEs retain validity in the physical world.
In the presence of diverse observations of the physical world by intelligent models \& systems, effective PAEs need to overcome the perturbation to manipulate the target model.
This significantly reduces the attack success rate of typical AEs, especially for potent AEs that are iterative optimized~\cite{DBLP:conf/iclr/KurakinGB17a}.
Currently, the key method is to simulate the physical world with the \textit{Expectation over Transformation (EoT)}~\cite{EOT} paradigm to reflect physical world randomness~\cite{wolfram1985origins}. Also, randomness injection, or domain randomization, is crucial in constructing reality simulation models~\cite{dai2022domain} that prevent the optimizer from exploiting the sim-to-real error of the simulator~\cite{2021pamisim2real}.

In the past decade~\cite{pami24paedecade}, algorithms supporting PAE generation have mainly focused on applying the iterative optimization with \textit{EoT} paradigm to find a universal PAE that is robust under all simulated scenes~\cite{zhao2019seeing,huang2024towards,zhu2023understanding, wei2023unified, guesmi2024dap}.
On the contrary, the solutions based on the inference of non-iterative models, \emph{e.g.}, conditional generative networks~\cite{xiao2018generating}, have been validated only in the digital or randomness-free scenarios.
However, the iterative paradigm does not support scene-adaptive PAEs because it requires inefficient repetitive optimization~\cite{wei2022simultaneously, liu2023x}. It is difficult to fully achieve the capability to present the corresponding PAE by current observation,\textit{ i.e.,} \textit{scene-aware}, for the attackers. \textcolor{red1}{We overcome the obstacle by advancing the training technology of efficient generative neural networks for this task.}

In terms of specific tasks, a typical PAE is patch-form~\cite{advpatch,shunchang22crowdcounting,zhang23CAPatch}, and object detection models are often employed for evaluating the PAEs~\cite{eykholt2018physical, Hu2021NaturalisticPA, TSEA}.
This task is not easy since current detection models are well-optimized.

\subsection{Definitions}

\label{sec:problem_form}
Next, we present the key problem definition, and the key notations are provided in Table~\ref{table:notations}.
For attacks in the dynamic and physical worlds, the key difference is the uncontrollable nature of the attack process and the limited perceptibility of the surrounding environment.
We formulate the attack scene as the current state of the world $\mathbf X \in \mathbb{X}$, the transformation parameters $\boldsymbol \theta \in \boldsymbol{\Theta}$ of the attack injection operation $\oplus$, and the input obtained by the attacker containing \textit{limited information} about $\mathbf X$ and $\boldsymbol \theta$, denoted as the physical context data $\mathbf{P}_\mathrm{X} \in \mathcal{P}$.
To achieve practical \textit{real-time} generation, we focus on the challenge of constructing $\mathcal{G}$ as a generative neural network that models the mapping from the physical observation
$\mathbf{P}_\mathrm{X}$ and the corresponding physical AE $\boldsymbol \delta$.
The problem is formulated as follows.
\begin{define}
\textbf{The dynamic physical adversarial attack problem} is defined as finding a PAE generator $\mathcal G$ that maps the observed physical context data $\mathbf{P}_\mathrm{X} \in \mathcal{P}$ into the PAE $\boldsymbol \delta \in \mathbb D$, formulated as:
    \begin{equation}
    \textit{find } \mathcal G  \text{  s.t.  } \mathcal{F}(\mathbf X \oplus (\mathcal G(\mathbf{P}_\mathrm{X}), \boldsymbol \theta )) \in Y_{adv},
    \label{eq:attack_prob}
\end{equation}
where $Y_{adv}$ is the range of successful attacks defined with the result of victim model(s) $\mathcal{F}$, $\oplus$ is the attack injection operation that updates the state of the world $\mathbf X \in \mathbb X$ with the PAE $\boldsymbol \delta = \mathcal G(\mathbf{P}_\mathrm{X})$, and $\mathcal G$ is the PAE generating network or the iterative algorithm.
\end{define}

\begin{table}[t]
    \caption{Glossary of Key Notations.}\label{table:notations}
    \resizebox{\linewidth}{!}{%
    \begin{tabular}{p{1.6cm}|p{6.0cm}}  
    \toprule
      \textbf{Notation} & \textbf{Description}\\
      \midrule
      $\boldsymbol \delta$& Physical Adversarial Example (PAE)\\ 
      $\mathcal G; \mathcal F$& Dynamic PAE Generator; victim/target model\\
      $\mathbf X; \mathcal X$& State of the world; training dataset\\
      $\mathbf{P}_\mathrm{X}; \mathcal P$ & Observation of $\mathbf X$; range of variable $P$\\
    $\mathcal Y$&Range of $\mathcal F$'s output;\\
    $\mathcal Y_{adv}$&Range of successful attacks in $\mathcal Y$\\
 $\mathbf Z; \mathcal Z$&Latent embedding; latent space\\
      $ \oplus; \boldsymbol \theta$& Attack injection operation; parameter of $\oplus$\\
      $\mathrm{Enc}; \mathrm{Dec}$& Encoder network; decoder network\\
      $H;  I$& Shannon entropy; mutual information\\
      $\text{skewness}(\cdot)$& Skewness statistic\\
      $\lambda$ &  Sample-wise loss weight, encoded into $\mathcal{G}$\\
      $\alpha$ &  Task-wise loss weight\\
      $\mathscr L_{\boldsymbol{\lambda}}$ & Residual fusion operator parameterized by $\boldsymbol\lambda$\\
      $\mathcal L_\mathrm{Atk}; \mathcal L_\mathrm{Inv}$& Attack loss; invisibility loss;\\
      $\mathcal L_\mathrm{Reg};$ & Regularization loss\\
 \bottomrule
    \end{tabular}}
\end{table}

\section{Methodology}

\begin{figure*}[t]
    \centering
    \includegraphics[width=1\linewidth]{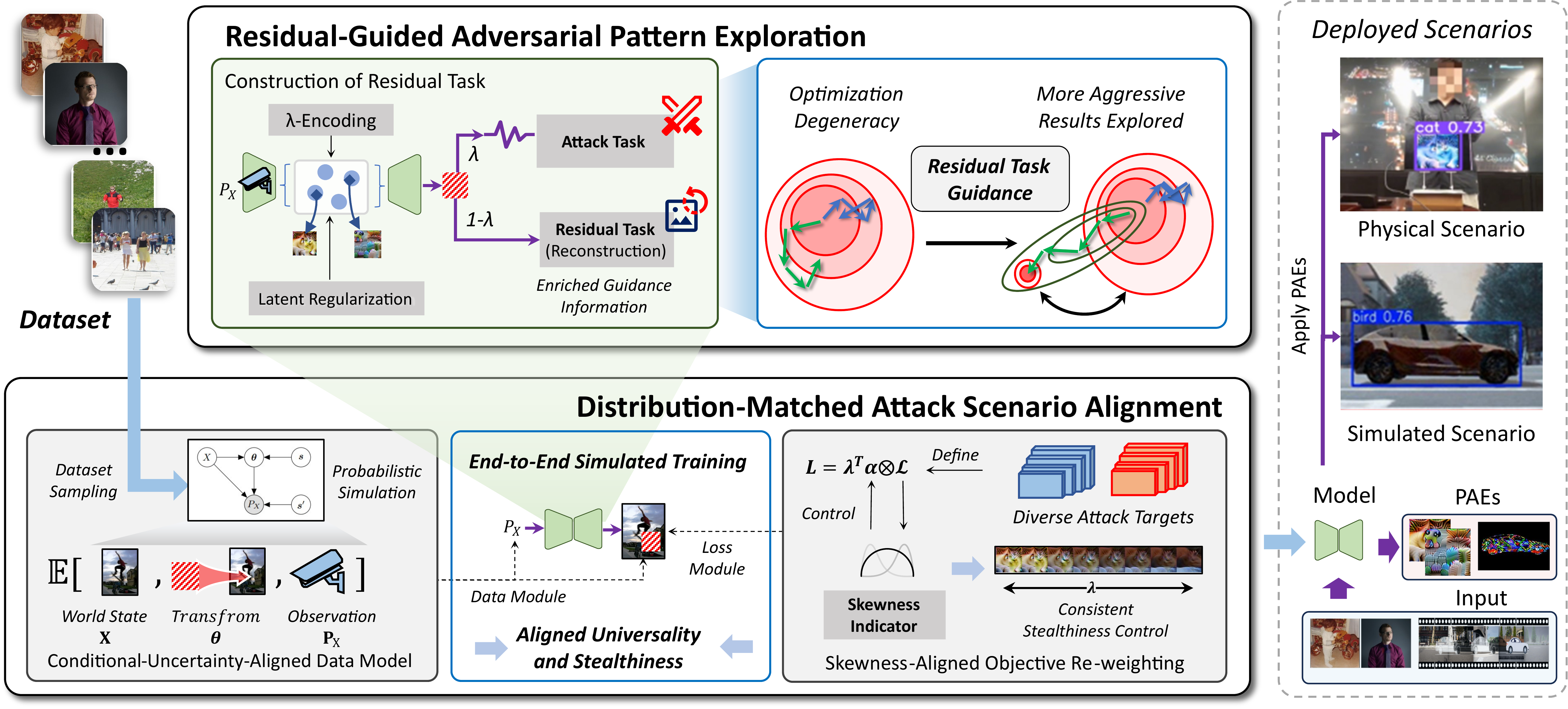}
    \vspace{-1em}
    \caption{\textcolor{red2}{The overview of the \THEMODEL\ framework, consisting of \textit{Residual Guided Adversarial Pattern Exploration} that solves the training difficulties and the design methodology of \textit{Distribution-Matched Attack Scenario Alignment} that aligns the generator to real-world constraints. The former enriches the guidance information with a residual-constructed reconstruction task, which decouples and diversifies the exploration of scene-conditional PAEs (shown as green \& blue arrows). The latter adapts a probabilistic simulation process to model the real-world uncertainty, and a skewness-aligned objective weight controller for automatically balancing the stealthiness and attack objectives.}}
    \vspace{-1em}
    \label{fig:method overview}
\end{figure*}

The overview of the \THEMODEL\ framework is shown in Figure~\ref{fig:method overview}.
\textcolor{red}{
To solve the dynamic PAE generation problem and achieve real-time performance, modeling the relation within the simplified and unified representation is crucial, which formulates the general neural architecture of the PAE generator as follows.
}
\begin{define}[Neural PAE Generator] The PAE generation module $\mathcal{G}$ is defined by the unified representation learning paradigm as:
\begin{equation}
    \mathcal G := \mathrm{Dec} \circ \mathrm{Enc}, \ \mathrm{Enc}: \mathcal P\to \mathcal Z \subseteq \mathbb{R}^d,
\end{equation}
where $\mathcal Z$ is the space of latent representation, and $\mathrm{Enc}$ and $\mathrm{Dec}$ are the encoder and the decoder, respectively.
\end{define}

\noindent
\textcolor{red}{
\THEMODEL\ framework provides the key guidance for the training and the design of $\mathcal{G}$.
The framework synergizes two key components:
residual-guided adversarial pattern exploration that tackles the training difficulties of $\mathcal{G}$; and distribution-matched attack scenario alignment that adapts $\mathcal{G}$ to the physical world,} thereby achieving the scene-aware dynamic adversarial attack in the real world.
\textcolor{red}{Based on these components, \THEMODEL\ establishes an end-to-end training pipline for $\mathcal{G}$.}

\subsection{\textcolor{red}{Residual-Guided Adversarial Pattern Exploration}}

\label{sec:aux_task}


\textcolor{red}{We first focus on the non-trivial training problem of the generator $\mathcal{G}$ under Eq.~\ref{eq:attack_prob}.
We found that the training degeneracy problem always occurs in this task (shown in the experiments in section~\ref{app:mode_vollapse}).
To characterize the key difficulty, we propose the \textit{limited feedback information restriction} model that accounts for the degeneracy of the generative training.}
Inspired by deep residual learning, we tackle the problem by constructing a \textbf{residual task} that relaxes the original attack training and breaks the \textit{limited feedback information restriction}.
\textcolor{red}{
We further propose the regularized latent encoding to tackle the \textit{latent evasion} problem and achieve stabilized multi-task training.
Afterward, the training escapes from degenerate solutions and allows a more stable and comprehensive exploration of feasible scene-related PAEs.
}

\emph{\textbf{$\bullet$ \textcolor{red}{Characterizing} the Degeneracy:}}
\textcolor{red}{We utilize an information-theoretic modeling approach to construct a concise model for the training problem and then establish a connection between noisy gradients and the trajectories of generated samples.}
Previous works have demonstrated that \textit{obfuscated gradient}~\cite{athalye2018obfuscated} is a mechanism of adversarial defense techniques, and adaptive attacks shall overcome it~\cite{kang2024diffattack}. Section~\ref{app:pdg} provides compelling evidence that the simulated transformation has a similar impact on PGD optimization, confirming the existence of the noisy gradient problem.
Intuitively, a small distortion in the environment $(\mathbf X, \boldsymbol\theta)$ may cause the gradient feedback of the attack task $\nabla_{\boldsymbol\delta} \mathcal L_\mathrm{Atk}$ changes significantly.
We formulate the \textbf{limited feedback information restriction} to model the problem based on the mutual information and entropy metrics on $\nabla_{\boldsymbol\delta} \mathcal L_\mathrm{Atk}$ and the (bottleneck) latent variable $\mathbf Z$:
\begin{equation}
    I(\nabla_{\boldsymbol\delta} \mathcal L_\mathrm{Atk}; \mathbf Z) / H(\nabla_{\boldsymbol\delta}  \mathcal L_\mathrm{Atk}) < \epsilon,
    \label{eq:information_bound}
\end{equation}
indicating that the uncertainty of $\nabla_{\boldsymbol\delta}  \mathcal L_\mathrm{Atk}$ is significantly higher than the uncertainty in $\nabla_{\boldsymbol\delta}  \mathcal L_\mathrm{Atk}$ reduced by $\mathbf Z$.

We then analyze the sample trajectory with this model to understand the optimization degeneration problem.
Specifically, considering the gradient descent process of deep learning, the trajectory of the generated PAEs $\boldsymbol \delta$ in training step $t$ can be modeled by:
\begin{equation}
\begin{aligned}
\boldsymbol \delta^{(t+1)} &= \boldsymbol\delta^{(t)} + \eta  \nabla_{\boldsymbol\delta} \mathcal L_\mathrm{Atk}^{(t)},\\
        \nabla_{\boldsymbol\delta} \mathcal L_\mathrm{Atk}^{(t)} &= f^{(t)}(\mathbf Z, \boldsymbol{\delta} ) + g^{(t)}(\boldsymbol{\delta}),
\end{aligned}
\end{equation}
where $\eta$ is the learning rate, $g^{(t)}$ is the noise introduced by random simulated transformation, and the term $f^{(t)}$ is the components correlated with $\mathbf Z$, representing the learnable information.
By \textit{rate-distortion theory}, the  signal power $\mathbb{E}[|f^{(t)}|^2]$ is significantly lower than the noise power $\mathbb E[|g^{(t)}|^2]$ if $I(\nabla_{\boldsymbol\delta} \mathcal L_\mathrm{Atk}^{(t)}; \mathbf Z)$ is relatively small according to Eq.~\ref{eq:information_bound}.
Based on the empirical fact of the existence of universal AE~\cite{moosavi2017universal}, the AE optimization contains isotropic components, and $\mathbb E[g^{(t)}]$ is large.
Thus, the optimization of PAE may converge to the local minimum determined by $g^{(t)}$ and $\mathbb E[g^{(t)}]$ even with sufficiently large steps $t$, which is coherent with the experiment results in section~\ref{app:pdg}.

\begin{figure}[h]
    \centering
    \includegraphics[width=1.0\linewidth]{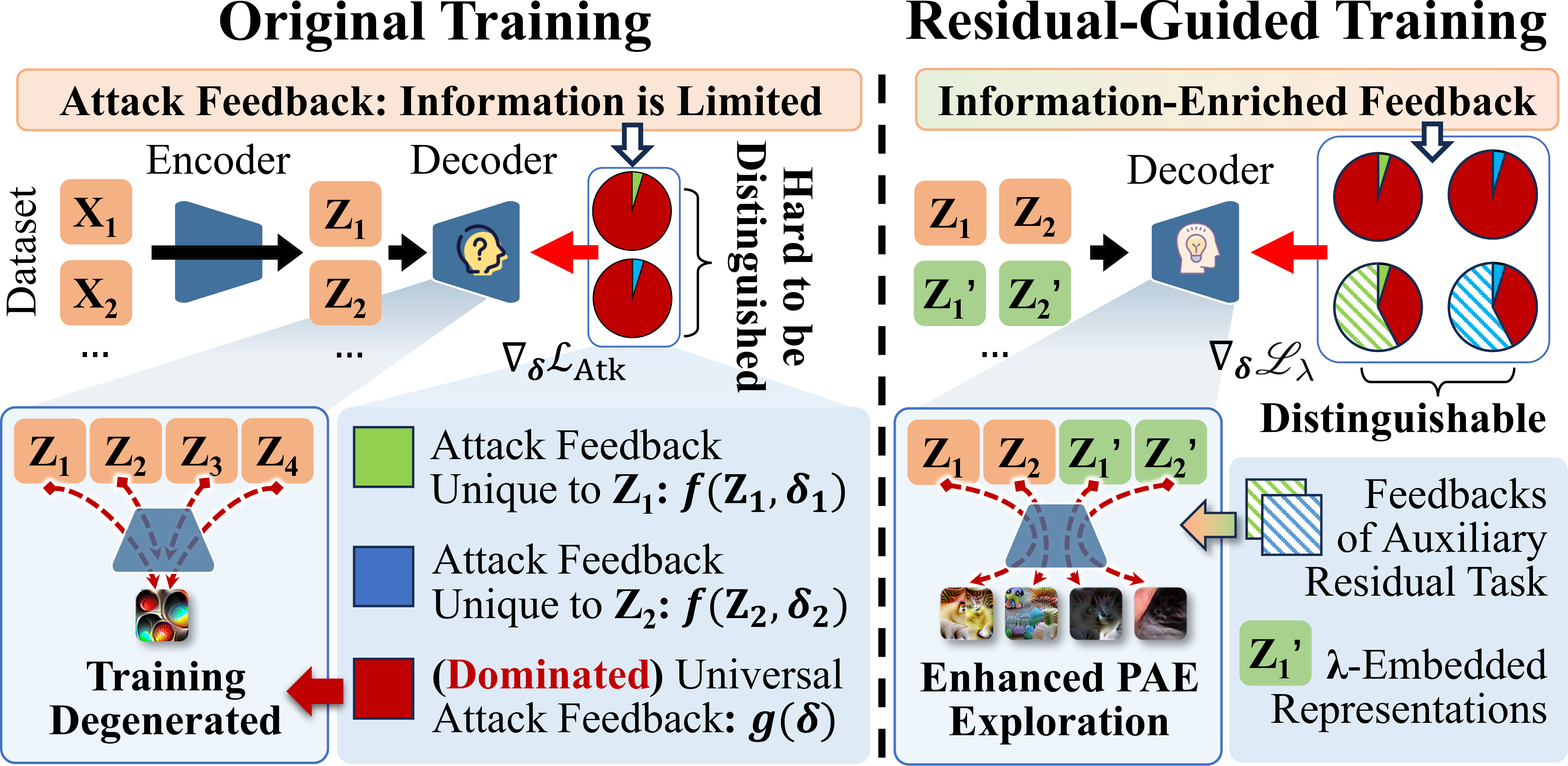}
    \caption{\textcolor{red1}{Training Problem and Solution Illustration.}}
    \label{fig:training_interpretation}
    \vspace{-0.5em}
\end{figure}

\noindent
\textcolor{red1}{
A more intuitive illustration is shown in Figure~\ref{fig:training_interpretation}.
Specifically, the original training method without the residual task is unsatisfactory due to the limited feedback information constraint. Our qualitative analysis shows that it leads to the degeneracy of training, \textit{i.e.} the generator only generates a single form of PAE,  which is independent of whether the optimizer includes techniques like momentum to escape from local optima.
The advancements in generative models, such as diffusion models~\cite{song19scoregen} and autoregressive models~\cite{tian2024visual}, have significantly improved the quality of complex data generation with better training task construction.
However, the degeneracy highlights the unique challenges inherent to the task of dynamic PAE training.
Therefore, we argue that addressing it requires a fundamental transformation of the training task itself.
As a solution, we propose the residual-guided training that breaks the limited feedback information constraint by redefining the training task, }\textcolor{red2}{leading to the decoupled and diversified exploration of PAEs.}


\emph{\textbf{$\bullet$ \textcolor{red1}{Exploration Enhancement with Residual Task:}}}
\textcolor{red1}{As shown in the right part of Figure~\ref{fig:training_interpretation}, we bypass the difficulty by redefining the training task to enrich the feedback information.}
Specifically, we relax the adversarial attack task with the additional task $\mathcal R$ with the conditional parameter $\lambda$, and modify the loss from $\mathcal L_\mathrm{Atk}(\boldsymbol \delta,\cdot)$ to $\mathscr  L_{ \lambda }(\mathcal L_\mathrm{Atk}, \mathcal{L}_{\mathcal R})$, defined as:
\begin{equation}
    \mathscr  L_{ \lambda }(\mathcal L_\mathrm{Atk}, \mathcal{L}_{\mathcal R}) := \lambda \mathcal L_\mathrm{Atk}(\boldsymbol \delta_\lambda, \cdot)+ (1 - \lambda)\mathcal L_\mathcal {R}(\boldsymbol \delta_\lambda,  \cdot).
    \label{eq:residual_task}
\end{equation}
We denote $\mathscr  L$ as the \textit{residual fusion} operator.
The modification aims to break the noisy gradient model in Eq.~\ref{eq:information_bound} by constructing $\mathcal L_\mathcal {R}$ to be more dependent on the learnable latent representation:
\begin{equation}
   \exists \lambda \text{ s.t. } I(\nabla_{\boldsymbol\delta_\lambda} \mathscr   L_{ \lambda }(\mathcal L_\mathrm{Atk}, \mathcal{L}_{\mathcal R}); \mathbf Z_\lambda)  \gg I(\nabla_{\boldsymbol\delta} \mathcal L_\mathrm{Atk}; \mathbf Z).
\end{equation}
Afterward, the imbalanced noise issue in SGD optimization could be solved, which mitigates the occurrence of degeneration.
Further, the optimal PAE solution remains identical when $\lambda = 0$.
Since the gradient takes a noiseless path, or a residual path, to the parameters of $\mathcal{G}$, we identify task $\mathcal R$ as the \textbf{\textit{auxiliary residual task}}.

\textcolor{red1}{From the perspective of the PAE generating space, the goal of the residual task in generating PAEs is to encourage the \textbf{exploration} of the global space of PAEs.}
Inspired by the construction of the denoising task in diffusion models to learn the entire gradient field~\cite{song19scoregen}, we designate local area reconstruction as the auxiliary residual task $\mathcal R$ ($\mathcal L_\mathcal R := \mathcal L_\mathrm{Inv}$) since it is directly related to the model input $\mathbf P_\mathrm{X}$, is well studied by generative models, and lets the model learns to generate PAEs at different magnitudes, possessing the utility of \textit{controllable stealthiness}.
Specifically, we adapt the mean square error (\textit{MSE}) as the objective quality metric and the \textit{LPIPS}~\cite{zhang2018perceptual} as the subjective quality metric, and integrate them as:
\begin{equation}
  \mathcal{L}_\mathrm{Inv} := \mathrm{MSE}(\mathbf X, \mathbf X') +  \mathrm{LPIPS}(\mathbf X, \mathbf X').
\end{equation}
where $\mathbf X$ and $\mathbf X' = \mathbf X \oplus \boldsymbol \delta$ are the global image captures before and after applying PAE $\boldsymbol \delta$, respectively.

To enable diverse sampling of $\lambda$ during training,
the magnitude of attack objectives for each batch of data $\boldsymbol{\lambda} = [\lambda_1, \lambda_2, ..., \lambda_b]$ is sampled by the distribution:
\begin{equation}
\label{eq:sample_lambda}
    \lambda_i \sim 
\begin{cases}
\mathrm \delta(0)& \text{i} \in [1,\  b / 4]\\
\mathrm \delta(1)& \text{i} \in (b / 4,\  b / 2]\\
\mathcal U(0, 1) & \text{i} \in (b / 2,\  b ]
\end{cases},
\end{equation}
where $\mathrm \delta$ denotes the Dirac distribution and $b$ denotes the batch size.
The distribution is constructed by sampling minimal, maximal, and random hyperparameters and is inspired by the sampling techniques in AutoML training~\cite{yu2019universally}.
To generate PAEs conditioned on $\lambda$, $\lambda$ is encoded into the latent representation $\mathbf Z$ in $\mathcal{G}$ by $\log$ embeddings and a $\mathrm{MLP}$ module modeling the non-linear transformation:
\begin{equation}
    \mathbf {\color{red}Z}_{\boldsymbol{\color{red}\lambda}} = \mathrm{MLP}(-\log ( \max \{\boldsymbol  \lambda, \exp(-10)\}) + \mathbf Z.
\end{equation}
\textcolor{red}{We denote the $\boldsymbol \lambda$-conditioned generator as $\mathcal{G}_{\boldsymbol \lambda}$, which generates PAEs $\boldsymbol{\delta}_{\boldsymbol \lambda}= [\boldsymbol\delta_{\lambda_1}, \boldsymbol\delta_{\lambda_2}, ... ,\boldsymbol\delta_{\lambda_b}]$ with conditioned magnitudes in each training batch,} and the same $\boldsymbol\lambda$ is applied to Eq.~\ref{eq:residual_task} as training objectives.


\label{sec:learning framework}

\label{sec:method_latents_evation}


\emph{\textbf{$\bullet$ Stabilizing Sample Exploration:}}
The particularity of the dynamic PAE generation also exists in the latent space of generator $\mathcal G$.
Empirically, we observed a non-trivial gradient explosion behavior in the generator during training.
The analysis is shown in the supplementary material, indicating the difference between PAE generation and conventional tasks.
We recognize it as the \textbf{latent evasion} problem, which stems from the process of exploring PAEs.

Specifically, due to the analyzed noisy gradient problem, the discovery of new conditional PAEs often requires multiple steps of fitting to emerge.
Meanwhile, the disentangled representations and weights within the model continuously deviate towards a certain direction throughout the multi-step fitting process, causing their norms to continually increase.
Therefore, the optimization speed of the model’s internal representations should be slowed down.
Thus, to tackle this problem, we incorporate a regularization loss on the latent representation of PAEs $\mathbf Z_{\boldsymbol\lambda}$ to stabilize the exploration, which together regularizes the PAEs with the total variation loss $\mathcal{L}_\mathrm{TV}$~\cite{rudin1992nonlinear} on the data space that eliminates the frequency discrepancy between PAEs and physically realizable images:
\begin{equation}
    \mathcal{L}_\mathrm{Reg} := \gamma  \cdot ||\mathbf Z_{\boldsymbol\lambda}||_2^2 + \beta \cdot \mathcal{L}_\mathrm{TV},
    \label{eq:reg}
\end{equation}
where $\gamma$ is the hyperparameter. Note that the latent regularization is more like \textit{BatchNorm}, which aims at tackling the optimization difficulties rather than preventing overfitting. We further compare the regularization with the classical VAE prior in experiments.


\subsection{\textcolor{red}{Distribution-Matched Attack Scenario Alignment}}

Next, we focus on the alignment problem between $\mathcal{G}$ and the real-world adversarial attack scenario.
Regarding the trade-off, the optimal attack performance is highly related to the universality and the stealthiness; the training environment of $\mathcal{G}$ shall align these properties with the requirements of the real-world scenario.
Therefore, we regard the alignment problem as the unbiased construction problem of the training data and objectives.
To solve the problem, we propose the \textit{Distribution-Matched Attack Scenario Alignment} as a design methodology, consisting of the conditional uncertainty alignment and the objective bias alignment.
The former controls the attack universality by aligning the sampled scenes' distribution and the real-world environment, while the latter controls the attack stealthiness by aligning the skewness statistics of the generator's loss distribution.

\begin{figure}[h]
    \centering
    \vspace{-0.5em}
       \begin{tikzpicture}

  \node[obs]                               (P) {$\mathbf{P}_\mathrm{X}$};
  \node[latent, above=of P]  (t) {$\boldsymbol{\theta}$};
  \node[latent, left=of t] (X) {$\mathbf X$};
    \node[latent, right=of t]  (s) {$\boldsymbol{s}$};
    \node[latent, right=of P]  (ss) {$\boldsymbol{s}'$};

  \edge {X,t,ss} {P} ; %
  \edge {s} {t} ; %
  \edge {X} {t} ; %


    \end{tikzpicture} 
    \caption{Conditional Probabilistic Model Generating Dynamic Physical Adversarial Attack Scenes}
    \vspace{-0.5em}
    \label{fig:prob_simulation}
\end{figure}

\emph{\textbf{$\bullet$ Conditional-Uncertainty-Aligned Data Model:}}
\textcolor{red}{Bringing EoT~\cite{EOT} and domain randomization~\cite{dai2022domain}, we align the training and real-world scenes by creating a conditional probabilistic model to generate both of the parameters of the attack injection process $\boldsymbol \theta$ and the attackers' observation $\mathbf{P}_\mathrm{X}$, as illustrated in Figure~\ref{fig:prob_simulation}, based on the problem formulation in Equation~\ref{eq:attack_prob}.}  In this model, $\boldsymbol \theta$ are generated by the current state of the world $\mathbf X$ and a random factor $\boldsymbol s$, which, along with $\mathbf X$ and another random factor $\boldsymbol s'$, jointly generate the physical context data $\mathbf{P}_\mathrm{X}$ that the attacker acquires.
The training is performed by minimizing the expected losses under the data model, where $\Omega$ and $\Theta$ are the probability distribution that generates $\mathbf{P}_\mathrm{X}$ and $\boldsymbol\theta$, respectively:
\begin{equation}
\label{eq:omega}
\begin{aligned}
        \min \mathbb{E}_{\mathbf{P}_\mathrm{X}\sim \Omega(\boldsymbol{\theta ,\mathbf X}), \boldsymbol \theta \sim  \Theta(\mathbf X), \mathbf X\sim \mathcal X}\left[ {\mathcal L\Big( \mathcal F \big( \mathbf X \oplus  ( \mathcal G (\mathbf{P}_\mathrm{X}), \boldsymbol \theta ) \big)  \Big) } \right],
\end{aligned}
\end{equation}
where $\mathcal L$ is the total loss function, which is defined as $\mathcal L_{\mathrm{Total}}$ in the following context.
For patch attacks, the attack injection $\oplus$ is formulated as:
\begin{equation}
\begin{aligned}
    \mathbf X \oplus (\boldsymbol \delta, \boldsymbol{\theta}) := & \mathbf X \otimes (1 - \boldsymbol{m}_\theta) + \\
    & \textit{AffineTransform}\left( \boldsymbol \delta, A_{\boldsymbol{\theta}}\right) \otimes \boldsymbol{m}_\theta, \
\end{aligned}
 \label{equ:gaussian_blur}
\end{equation}
where the binary mask $\boldsymbol{m}_\theta\in \{0, 1\}^{\mathrm H_1\times\mathrm  W_1}$ representing the configurable patch location corresponding to $\boldsymbol \theta$.
Considering the input of the attack device, the local feature corresponding to the stealthiness and the global context corresponding to the aggressiveness are included in $\mathbf{P}_\mathrm{X}$, as the attack objective is correlated with the global attention field of the model~\cite{wang2024generate}.
Specifically, the local feature is introduced by the local image content $\mathbf X_{local} \in \mathbb R^{\mathrm{H}_2\times \mathrm{W}_2\times 3}$ around the patch, while the global context corresponds to a global image $ \mathbf X_{global} \in \mathbb R^{\mathrm H_1\times \mathrm  W_1\times 3}$ representing the observed environment.
To simulate the distortion in the open world, \textit{the RandAugment}~\cite{nips20randaug} method is applied to $\mathbf X_{global}$ and $\boldsymbol m_\theta$ simultaneously.
\textit{Las Vegas algorithm} is performed to ensure the physical validity: the augmentation pipeline reruns if the result is invalid.
Thus, the distribution $\Omega$ that generates $\mathbf{P}_\mathrm{X}$ is formulated as:
\begin{equation}
\begin{aligned}
   \Omega(\boldsymbol{\theta},\mathbf X) &= \left[\mathbf X_{local}; \textit{RandAug}(\mathbf X_{global},\boldsymbol s'); \textit{RandAug}(\boldsymbol{m}_\theta, \boldsymbol s')\right],
    \label{eq:Y_augment}
\end{aligned}
\end{equation}
where $\boldsymbol s'$ is the generated random seed for \textit{RandAugment}.
$\Theta$ is constructed by sampling the object location according to the objects in $\mathbf X$, and $ \mathbf X$ is constructed by the training dataset of the target model.

\emph{\textbf{$\bullet$ Skewness-Aligned Objective Re-weighting:}}
The dynamic PAE generator may be deployed to evaluate and attack diverse victim models and corresponding tasks.
\textcolor{red1}{To balance the stealth and attack objectives in PAE exploration, achieve controllable stealthiness, and avoid manual adjustment of the optimization-related parameters when facing new targets, we further propose an objective re-weighting approach based on the \textbf{\textit{skewness}} measurement~\cite{groeneveld1984measuring}.
Specifically, we focus on adding and determining the additional relative weights $\alpha \in  (0, +\infty)$ of the \textit{residual task}, and define the optimization objective using the \textit{residual fusion} operator in Eq.~\ref{eq:residual_task} as: $\mathscr  L_{ \lambda }(\mathcal L_\mathrm{Atk}, \alpha \mathcal L_\mathrm{Inv}) = \boldsymbol{\lambda} \mathcal L_\mathrm{Atk} +(1 - \boldsymbol{ \lambda} ) \alpha \mathcal L_\mathrm{Inv}$.}

%

\begin{figure}[h]
    \centering
    \subfloat[Different Skewness]{
    \includegraphics[width=0.44\linewidth]{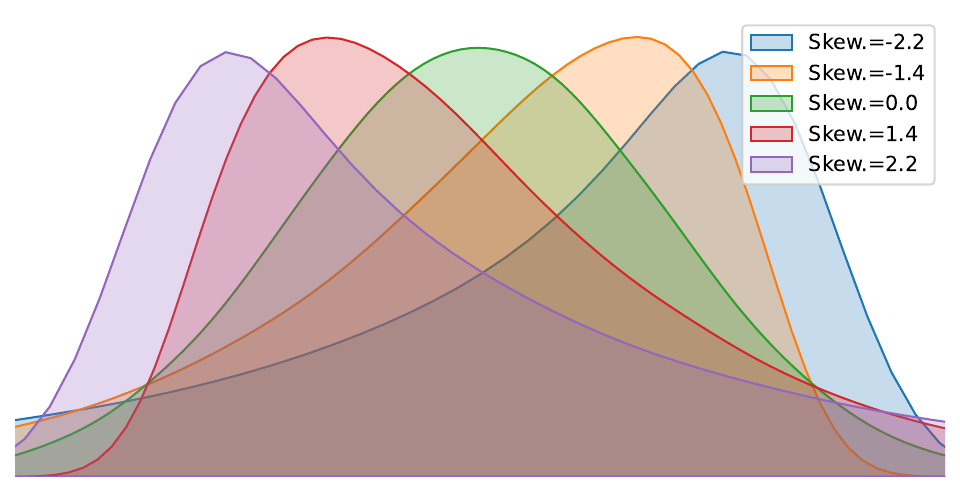}
    \label{fig:skewness}
    }\subfloat[Relation with Loss Weights]{
    \includegraphics[width=0.51\linewidth]{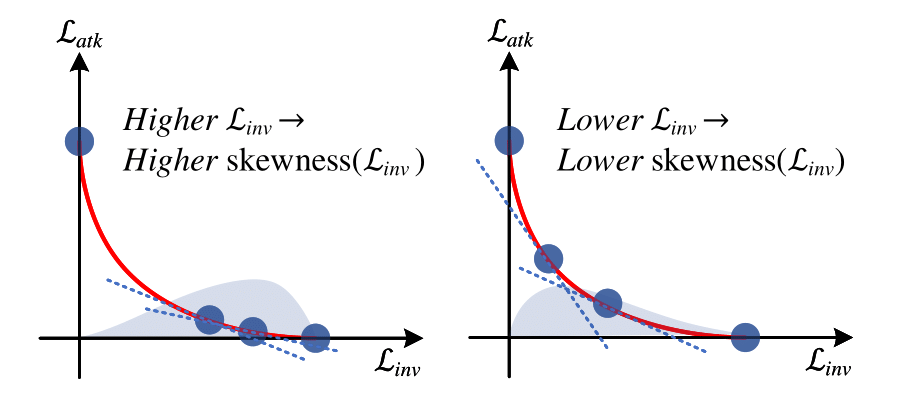}
    \label{fig:control}
    }
    \vspace{-0.3em}
    \caption{Skewness-based Hyperparameter Control}
    \vspace{-1em}
\end{figure}
\begin{figure}[h]
    \centering
    \vspace{-0.5em}
    \includegraphics[width=0.8\linewidth]{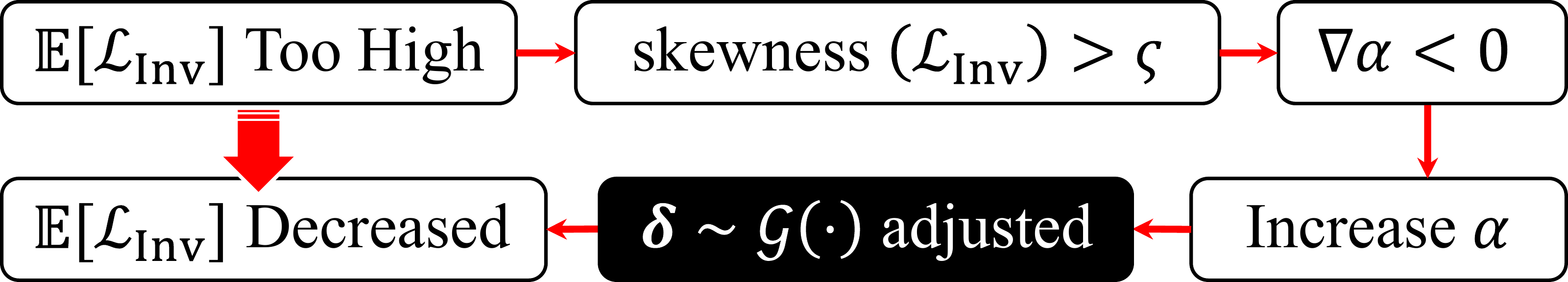}
    \caption{\textcolor{red1}{Closed-loop Control of Loss Distribution}}
    \label{fig:skewness_controller_illustation}
     \vspace{-1em}
\end{figure}

Intuitively, skewness aims to measure the lack of symmetry in the distribution of a dataset around its mean. 
On one hand, as shown in Figure~\ref{fig:skewness}, if the skewness is positive, the distribution is right-skewed or positively skewed, and vice versa.
On the other hand, as shown in Figure~\ref{fig:control}, the $\mathcal L_\mathrm{Inv}$ of the samples that are generated by middle-size $\lambda$ tends to decrease within the optimal curve as the weights $\alpha$ increase, leading to a decrease in the loss skewness.
This allows for the skewness to be utilized as an indicator variable to control the value of $\alpha$ by measuring the distribution of losses.
Thus, based on the skewness of the loss values, we define the gradient feedback of the $\alpha$-controller of loss $~\mathcal{L}$ as: 
\begin{equation}
\begin{aligned}
        \nabla \alpha &:= -\text{skewness}(\mathcal L_\mathrm{Inv})  + \varsigma,\\
        \text{skewness}(\mathcal{L_\mathrm{Inv}}) &:=  \mathbb E[\frac{(\mathcal{L}_\mathrm{Inv} - \mu(\mathcal{L}_{\mathrm{Inv}, \lambda=0,1}))^3}{\sigma(\mathcal{L}_{\mathrm{Inv}, \lambda=0,1})^3}],
\end{aligned}
\label{eq:gradalpha}
\end{equation}
where $\varsigma$ is the target skewness, $\mu$ and $\sigma$ are the mean and standard deviation of the loss when $\lambda\in\{0, 1\}$, which replaces the statistics for all samples for better stability.
The gradient $\nabla \alpha$ equals zero if and only if the loss is balanced, that is, $\text{skewness}(\mathcal L)  = \varsigma$.
\textcolor{red1}{
We apply the gradient descent with gradient defined by $\nabla \alpha$ to optimize $\alpha$, resulting in a closed-loop control of loss distribution on explored PAEs $\boldsymbol{\delta}\sim\mathcal{G}(\cdot)$, as shown in Figure~\ref{fig:skewness_controller_illustation}. The following proposition shows the convergence of the optimization. Detailed assumptions and proof are provided in the \textit{supplementary material} through an analysis of the trade-off curve.}
\begin{proposition} $\nabla^2 \alpha > 0$ if $\mathcal L_\mathrm{Atk}$ is strictly negatively correlated to $\mathcal L_\mathrm{Inv}$ for any set of measured PAEs $\boldsymbol\delta_\lambda$ that are optimal for the overall optimization objective $\mathcal L$ under a certain value of $\lambda$.
\end{proposition}
\noindent
\textcolor{red1}{For implementation, we set the target skewness of the reconstruction loss $\mathcal{L}_\mathrm{Inv}$ to zero ($\varsigma = 0$).
We use the shifted window algorithm to measure $\mu(\cdot)$ and $\sigma(\cdot)$, and then the skewness statistics, \textit{i.e}, the expectation $\mathbb E$ is estimated from the loss of recent trained batches.  Then, we use the \textit{Adam} optimizer with learning rate lr=1e-3 to optimize $\alpha$, and reparameterize $\alpha$ as $\alpha = e^\beta \in (0, +\infty)$ for the domain constraint. An equivalent loss is constructed to perform optimization along with $\mathcal{G}$ at each training iteration:
\begin{equation}
\begin{aligned}
    \textit{Adam}[{\alpha}] & \text{.zero\_grad()};\\
     \mathcal L(\alpha) &=\mathrm{float}(\nabla \alpha)  * \alpha; \\
     \mathcal L(\alpha)&\mathrm{.backward()};\\
    \textit{Adam}[{\alpha}] & \text{.step()}.\\
\end{aligned}
\label{eq:optim_alpha}
\vspace{-0.5em}
\end{equation}
}

\vspace{-1em}
\subsection{\textcolor{red}{Neural Architecture and Training Implementations}}
\textcolor{red1}{
Based on the introduced training technology, we realize the ongoing adaptation of the typical patch-based PAE generation through a lightweight construction of neural architecture and a modularized construction of the training pipeline. After training, we deploy the generator and enable the real-time generation by connecting it to a camera and display. Details are shown in the \textit{supplementary material}.}

\textcolor{red}{\emph{\textbf{$\bullet$ Neural Architecture:}}
\textcolor{red1}{
For the neural generator $\mathcal{G}_{\boldsymbol \lambda}(\mathbf{P}_\mathrm{X})$ $:= \mathrm{Dec}(\mathrm{Enc}_\mathbf{P}(\mathbf{P}_\mathrm{X},\boldsymbol \lambda))$, we constructed a lightweight physical environment encoder $\mathrm{Enc}_\mathbf{P}$ based on the multi-modal information fusion paradigm.
}
Specifically, the encoding model of $\mathbf{P}_\mathrm{X} = (\mathbf X_{local}',\mathbf X_{global}', \boldsymbol m')$ is constructed by:
\begin{equation}
\begin{aligned}
        \mathrm{Enc}_\mathbf{P} := \textit{MLP}_1 (  \textit{LN}( &\textit{MLP}_2(\textit{ResNet}_1(\mathbf X_{local}'))) \\
  +  \textit{LN}( &\textit{MLP}_3(\textit{ResNet}_2(\text{Concat}[\mathbf X_{global}', \boldsymbol m']))  \\
   + &\textit{MLP}_4(-\log ( \max \{\boldsymbol  \lambda, \exp(-10)\})))),
\end{aligned}
\end{equation}
where $\textit{Resnet}$ is for feature extraction, $\textit{MLP}$ denotes the multi-layer perception module that models the mapping, and \textit{LN} denotes the layer-norm module that regularizes the feature importance. $\mathrm{Dec}$ is a lightweight version of \textit{BigGAN }~\cite{brock2018large} decoder, with the self-attention module removed.}

\begin{figure}[h]
    \centering
    \vspace{-0.5em}
    \includegraphics[width=1.0\linewidth]{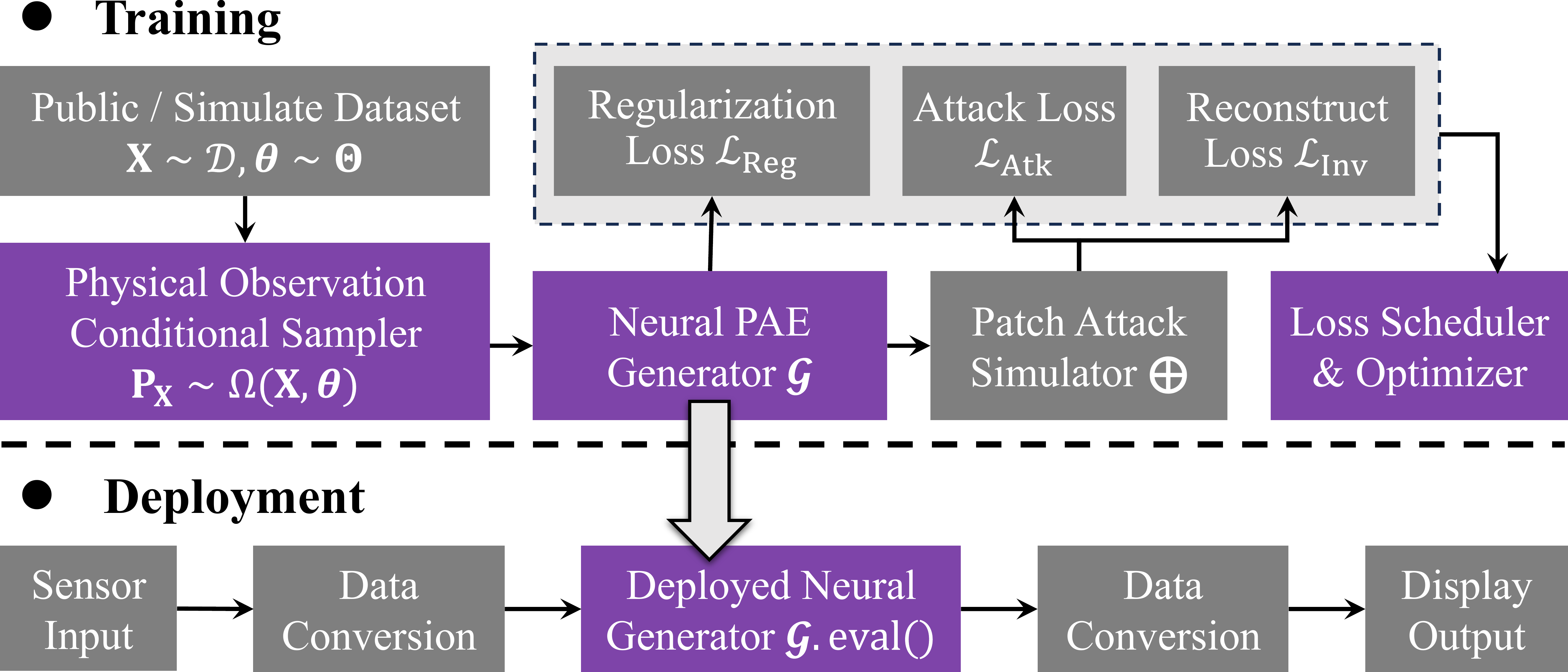}
    \caption{Training and deployment pipeline. \textcolor{red2}{The arrow in the training pipeline denotes the forward process in a training step, and the gradient optimization is performed on $\mathcal{G}$. The deployment pipeline processes the data and displays PAE as a real-time stream.}}
    \label{fig:train_pipeline}
    \vspace{-1em}
\end{figure}

\emph{\textbf{$\bullet$ Training Pipeline:}}
\textcolor{red}{Following the practice of multi-task fine-tuning, we train the generator $\mathcal{G}$ with the VAE-pretraining task and the new residual-guided adversarial attack task.}
The total loss is formulated as:
\textcolor{red1}{\begin{equation}
    \mathcal{L}_\mathrm{Total} = \mathscr  L_{ \lambda }(\mathcal L_\mathrm{Atk}, \alpha \mathcal L_\mathrm{Inv})  + \mathcal{L}_\mathrm{Reg}  +  \mathcal{L}_\mathrm{VAE},
\end{equation}
where $\mathscr  L_{ \lambda }(\cdot) :=\lambda \mathcal L_\mathrm{Atk} +(1 - {\lambda} ) \alpha \mathcal L_\mathrm{Inv}$ is the residual-guided training loss, $\mathcal{L}_\mathrm{Reg}$ is the regularization in Eq.~\ref{eq:reg}.
$\mathcal{L}_\mathrm{VAE}$ is generative training loss of $\beta$-VAE~\cite{higgins2016betavae}, and we use it to train $\mathrm{Dec}$ with images from the dataset $\mathcal X$.
It lets $\mathcal{G}_\lambda$ maintain the modeling capability of more complex natural images and improves the naturalness of the generated PAEs when the attack loss weight $\lambda$ is low.}
\textcolor{red}{
The overall training is shown in Algorithm~\ref{alg:main}.
\textcolor{red1}{
The modularized implementation of the training and deployment pipeline is shown in Figure~\ref{fig:train_pipeline}.
\textcolor{red2}{
The gray modules represent existing components in the deep learning and adversarial attack pipeline, while the purple modules are the new components to be constructed.}
The pipeline is straightforward, as no additional discriminators have been introduced.
$\mathcal{L}_\mathrm{VAE}$ is not shown since it is optional.
}
}

\begin{algorithm}[h]
\caption{\textcolor{red}{Ends-to-Ends \THEMODEL\ Training}}
\label{alg:main}

\KwIn{Dataset $\mathcal{D} \subset \textbf{Images} \times \textbf{BBoxes}$, Target model $\mathcal{F}$ with attack loss $\mathcal{L}_{\mathrm{Atk}}$, Untrained PAE generator $\mathcal{G}$}
\KwOut{Trained model $\mathcal{G}$}
$\alpha \leftarrow 1.0$ \par
\For{$epoch \in \{1, 2...n\}$, $step \in \{1, 2...m\}$}{
    Sample $X$ from the dataset $\mathcal{D}$. \par
     Sample the patch location ${\boldsymbol\theta}$. \par
    Sample task ratios $\boldsymbol{\lambda}$ by Equation~\ref{eq:sample_lambda}. \par 
    Pre-process $\mathbf{P}_\mathrm{X}$ by Equation~\ref{eq:Y_augment}. \par
    Generate PAEs $\boldsymbol{\delta}_{\boldsymbol\lambda} \leftarrow\mathrm{Dec} \circ \mathrm{Enc}_\mathbf{P}(\mathbf{P}_\mathrm{X},\boldsymbol{\lambda})$. \par
    \textcolor{red1}{$\mathbf X' \leftarrow  \mathbf X \oplus (\boldsymbol \delta_{\boldsymbol\lambda}, \boldsymbol{\theta})$. (Equation~\ref{equ:gaussian_blur})}\par
    \textcolor{red1}{$\mathcal L_\mathrm{Inv}\leftarrow\mathrm{MSE}(\mathbf X, \mathbf X') +  \mathrm{LPIPS}(\mathbf X, \mathbf X')$.}\par
    \textcolor{red1}{$\mathscr  L_{ \boldsymbol\lambda } \leftarrow \boldsymbol{\lambda} \mathcal L_\mathrm{Atk}(\cdot) +(1 - \boldsymbol{ \lambda} ) \alpha \mathcal L_\mathrm{Inv}$.} \par
    \textcolor{red1}{$\mathcal{L}_\mathrm{Reg} \leftarrow \gamma \cdot ||\mathbf Z_{\boldsymbol\lambda}||_2^2 + \mathcal{L}_\mathrm{TV} $.  (Equation~\ref{eq:reg})}\par
    \textcolor{red1}{$ \mathcal{L}_\mathrm{VAE} \leftarrow \beta$-VAE loss~\cite{higgins2016betavae} on $\mathrm{Dec}$ with data $\mathbf{X}$.} \par 
    \textcolor{red1}{$\mathcal{L}_\mathrm{Total}\leftarrow \mathscr  L_{ \boldsymbol\lambda }  + \mathcal{L}_\mathrm{Reg}  +  \mathcal{L}_\mathrm{VAE}$.} \par
    \textcolor{red1}{Optimize $\mathcal{G}$ with $ \mathcal{L}_\mathrm{Total}$ using \textbf{Adam} optimizer.} \par
    \textcolor{red1}{Optimize $\alpha$ with $\mathcal{L}_\mathrm{Inv}$ by Equation~\ref{eq:gradalpha},}~\ref{eq:optim_alpha}. 
}
\Return $\mathcal{G}$.

\end{algorithm}

\section{Experiments}

We evaluate our framework in both digital and physical environments on the object detection task since it is classical for both physical and digital attacks.
The experiments are conducted in the following aspects:
\ding{182} The performance of our model in the benchmark.
\ding{183} \textcolor{red}{The analysis corresponds with the methodologies that tackle the challenge of noisy gradient feedback in training and the alignment issues between trained generators and attack scenarios.}
\ding{184} The ablation of the key components.



\subsection{Experiment Settings}

In this subsection, we describe the key settings of the experiments.
We use the COCO~\cite{lin2015microsoft} and the Inria~\cite{dalal2005histograms} as the dataset of the person detection task, while data collected from \textit{CARLA} is used for the simulation experiment.
\textcolor{red}{Detailed \textit{method implementation}, \textit{dataset description}, and \textit{victim model description} are shown in the \textit{supplementary material}.}

\noindent
\textbf{Metrics:}
\textcolor{red}{The average precision (AP), formulated as follows by the precision $Pre(c)$ and recall $Rec(c)$ under the confidence threshold $c$, and average success rate (ASR) are applied as the metric of attack performance.}
\begin{equation}
\begin{aligned}
    AP &=  \int_{r=0}^1 \max_{c} Pre(c) \cdot [Rec(c) \le r]\ \mathrm d{r} \in [0,1].
\end{aligned}
\end{equation}
We adopt two confidence thresholds: the first is 
$c_{min} = 50\%$, with the corresponding metric denoted as $AP_{50}$, which serves as the evaluation metric identical to the practice of previous object detection attack evaluation. The second threshold is $c_{min}=1\%$, with the metric denoted as $AP_{01}$, as a more stringent evaluation metric.
All $\boldsymbol{iou}$ thresholds are set as 0.5.
The low-confidence and unattacked bounding boxes are filtered out during post-processing.
\textcolor{red}{For ASR, the attack is regarded as successful if the target's detection confidence is below 0.5.}

We use the Structural Similarity Index (SSIM) and the Learned Perceptual Image Patch Similarity (LPIPS) as metrics to measure the stealthiness of the patches.
Both SSIM and LPIPS have a range of [0, 1]. \textit{SSIM is not included as a loss during the training of our model}.

\begin{table}[h]
  \caption{\textcolor{red2}{Comparison of different baselines.}}
\resizebox{\linewidth}{!}{
    \color{red2}
    \centering
    \begin{tabular}{c|ccccc}
         \toprule
         Baselines &  PGD &  \textit{AdvGAN} &  \textit{GAN-NAP}& \textit{T-SEA}&  \textit{CAP}\\
         \midrule
         Optimization Space&  Image
& CNN&  Latents & Image & Image*\\
 Adaptive& \ding{51}& \ding{51}& \ding{55}&\ding{55} &\ding{55}\\
 Real-Time Adapted & \ding{55}& \ding{51}& \ding{55}&\ding{55} &\ding{55}\\
 Physical Capability& \ding{51}*& \ding{55}& \ding{51}&\ding{51} &\ding{51}\\
         Patch Attack&  \ding{51} &  \ding{55}&  \ding{51}& \ding{51}&\ding{51}\\
 Perturbation& \ding{51}& \ding{51}& \ding{55}&\ding{55}&\ding{55}\\
 \bottomrule
    \end{tabular}
}
\vspace{-1em}
\label{tab:cmp_methods}
\end{table}

\textcolor{red2}{
\noindent
\textbf{Baseline Attack Methods:} Our baselines include  \textit{PGD}~\cite{madry2017towards}, \textit{AdvGAN}~\cite{xiao2018generating}, \textit{GAN-NAP}~\cite{Hu2021NaturalisticPA}, also known as \textit{NAP}, \textit{T-SEA}~\cite{TSEA} and \textit{CAP}~\cite{wei2024revisiting}.
Each of them represents a distinct optimization method.
The comparison is shown in Table~\ref{tab:cmp_methods}, where \textit{CAP} applies a min-max optimization together with the ISP surrogate network.
For the strict physical patch attack scenario that matches our design, \textit{GAN-NAP}, \textit{T-SEA}, and \textit{CAP} are included.
Although not physically applicable due to computational issues, we also include PGD (with single-image EoT~\cite{EOT}), as there is no prior adaptive physical adversarial patch.
To further validate our technical contribution, we supplement additional comparison with \textit{GDPA}~\cite{Li2021GenerativeDP} on real-time digital adversarial patch generation for face recognition in Section~\ref{app:mode_vollapse}, and with \textit{AdvGAN} on real-time perturbation generation in Section~\ref{sec:application}.
}

\noindent
\textbf{Attack Method Implementations:}
\textcolor{red}{
For loss construction of \THEMODEL\, the maximum confidence value is employed following \textit{GAN-NAP} with the addition of a selection mask. We select the bounding box by the $\boldsymbol{iou}$ between the detector's generated bbox and the target person's bbox:}
\begin{equation}
    \mathcal{L}_\mathrm{Atk} =  || \boldsymbol{conf} \cdot [\boldsymbol{iou} > 0.3] ||_\infty,
\end{equation}
\textcolor{red}{Detailed implementations of \THEMODEL\ and baselines are in the \textit{supplementary material}.}

\begin{table}[h]
    \caption{\textcolor{red}{Spatial Transformation Settings. Position and size are related to the annotated bounding-box.}}
    \centering
    \begin{tabular}{c|cccc}
    \toprule
         Name &  Position &  Size &  Rotation & 3D Rotation \\
    \midrule
         Base&  [0.4, 0.6]&  [0.04, 0.06]&  [-9°, +9°]& 0\\
         S+&  [0.4, 0.6]&  [0.03, 0.07]&  [-9°, +9°]& 0\\
         P+&  [0.3, 0.7]&  [0.04, 0.06]&  [-9°, +9°]& 0\\
         AF&  [0.4, 0.6]&  [0.04, 0.06]&  [-9°, +9°]& [-30°, +30°]\\
         Zero&  0.5&  0.05&  0& 0\\
    \bottomrule
    \end{tabular}

    \label{tab:transformation}
\end{table}

\begin{table}[h]
    \caption{\textcolor{red}{Color Transformation Settings}}
    \centering
    \begin{tabular}{c|cccc}
    \toprule
          &  Brightness &  Saturation &  Contrast & Gaussian Noise \\
    \midrule
         Param. &  ±0.05&  ±0.1 & ±0.1 &  [0, 0.02]\\
    \bottomrule
    \end{tabular}

    \label{tab:colortransformation}
\end{table}

\noindent
\textcolor{red}{\textbf{Attack Simulation:} To ensure fairness, we use \textit{consistent} patch placement parameters in the evaluation.
The patch placement model for evaluation is based on affine transformation $A$, which can be decomposed into rotation matrices, scaling matrices, and translation matrices:
\begin{equation}
    A = A_{rotate}\cdot A_{scale}\cdot A_{trans}.
\end{equation}
Thus, we define separate parameter distribution models for rotation, scaling, and translation in the transformation.
Specifically, we sample the relative position, the relative size, the rotation in the 2D image, and the rotation in the 3D space (the rotation axis is parallel to the camera's imaging plane), from a predefined uniform distribution shown in Table~\ref{tab:transformation}.
Without specification, \textit{Base} transformation is applied in the following experiments.
In comparison to previous evaluations, a smaller patch size was implemented across the evaluated attacks, which increases the attack difficulty.
For color distortion, we apply brightness, saturation, and contrast transformation with uniform sampled parameters, and then we add Gaussian noise with standard variance sampled from the uniform distribution, as shown in Table~\ref{tab:colortransformation}.}

\begin{table*}[ht]

\caption{Performance Evaluation. \color{red}{↓} \color{black}indicates smaller values are better, while \color{red}{↑} \color{black} signifies larger values are preferred. Since the patch size is already constrained, stealthiness is only for better fairness.}
  \centering 
  \resizebox{\linewidth}{!}{
    \begin{tabular}{c c|c c c c c|c c c c c}
   \toprule
    \multicolumn{2}{c|}{Datasets} &  \multicolumn{5}{c|}{MSCOCO} & \multicolumn{5}{c}{Inria} \\
    Models& Method & AP$_{50}$\color{red}{↓} & AP$_{01}$\color{red}{↓}   & SSIM\color{red}{↑}& LPIPS\color{red}{↓}  & \color{red}{ASR↑}& AP$_{50}$\color{red}{↓}&  AP$_{01}$\color{red}{↓}& SSIM\color{red}{↑}& LPIPS\color{red}{↓}  & \color{red}{ASR↑}\\ 
    \midrule

    \multirow{7}{*}{\makecell[c]{Yolov8-n}}& Clean & $0.864$& $0.921$& --& -- &--& $0.933$& $0.941$& --& -- &--\\
    & Noise & $0.854_{\downarrow 1.0\%}$& $0.913_{\downarrow 0.8\%}$& 0.9892& 0.0137
 &0.7\%& $0.905_{\downarrow 2.7\%}$& $0.918_{\downarrow 2.3\%}$& 0.9950& 0.0070
 &2.6\%\\
 & PGD& $\underline{0.577}_{\downarrow 28.7\%}$& $\underline{0.715}_{\downarrow 20.6\%}$& 0.9883& 0.0218
 &23.5\%& $\underline{0.577}_{\downarrow 35.6\%}$& $\underline{0.644}_{\downarrow 29.7\%}$& 0.9947&0.0133
 &25.0\%\\
    & \textit{T-sea} & $0.604_{\downarrow 25.9\%}$&$0.750_{\downarrow 17.1\%}$&0.9893&0.0158
 &22.7\%& $0.389_{\downarrow 54.3\%}$&$0.430_{\downarrow 51.1\%}$&0.9953&0.0075
 &\underline{36.2\%}\\
    & \textit{GAN-NAP} &  $0.710_{\downarrow 15.4\%}$& $0.810_{\downarrow 11.0\%}$& \underline{0.9897}& \underline{0.0127} &11.9\%& $0.594_{\downarrow 33.9\%}$& $0.642_{\downarrow 30.0\%}$& 0.9954& 0.0084
 &25.0\%\\
 & \textcolor{red2}{\textit{CAP}} & \color{red2} $0.581_{\downarrow 28.2\%}$ & \color{red2} $0.724_{\downarrow 19.7\%}$ & \color{red2} \underline{0.9897} & \color{red2} 0.0148 & \color{red2} \underline{25.6\%} & \color{red2} $0.700_{\downarrow 23.3\%}$ & \color{red2} $0.757_{\downarrow 18.5\%}$ & \color{red2} \textbf{0.9956} & \color{red2} \textbf{0.0068} & \color{red2} 15.8\%\\
\rowcolor{gray!25}
    & \textit{\THEMODEL${_{\lambda=.8}}$}&  $\mathbf{0.331}_{\downarrow 53.3\%}$& $\mathbf{0.438}_{\downarrow 48.3\%}$& \textbf{0.9912}&\textbf{ 0.0091}
 &\textbf{42.0\%}& $\mathbf{0.307}_{\downarrow 62.6\%}$& $\mathbf{0.354}_{\downarrow 58.7\%}$& \textbf{0.9956}& \underline{0.0069} &\textbf{42.9\%}\\

    \midrule

    \multirow{7}{*}{\makecell[c]{Yolov5-s}}& Clean & $0.790$& $0.876$& --& -- &--& $0.917$& $0.921$& --& -- &--\\
    & Noise & $0.757_{\downarrow 3.4\%}$& $0.848_{\downarrow 2.8\%}$& 0.9892& 0.0137 &2.6\%
& $0.851_{\downarrow 6.6\%}$& $0.865_{\downarrow 5.7\%}$& 0.9951& \textbf{0.0068}
 &3.9\%
\\
 & PGD& $\underline{0.270}_{\downarrow 52.1\%}$& $\underline{0.423}_{\downarrow 45.3\%}$& 0.9884& 0.0223
 &\underline{44.0\%}& $\underline{0.283}_{\downarrow 63.4\%}$& $\underline{0.375}_{\downarrow 54.6\%}$& 0.9948&0.0137
 &\underline{41.3\%}\\f
    & \textit{T-sea} & $0.482_{\downarrow 30.8\%}$&$0.631_{\downarrow 24.5\%}$&0.9898&0.0141 &22.7\%& $0.365_{\downarrow 55.2\%}$&$0.410_{\downarrow 51.1\%}$&0.9950&0.0091
 &30.3\%\\
    & \textit{GAN-NAP} &  $0.401_{\downarrow 38.9\%}$& $0.564_{\downarrow 31.2\%}$& 0.9893& 0.0148
 &31.7\%& $0.352_{\downarrow 56.5\%}$& $0.433_{\downarrow 48.8\%}$& 0.9952& 0.0086
 &29.5\%\\
 & \textcolor{red2}{\textit{CAP}} & \color{red2} $0.393_{\downarrow 39.8\%}$& \color{red2}$0.534_{\downarrow 34.2\%}$& \color{red2}\underline{0.9899}& \color{red2}\underline{0.0132}& \color{red2}34.2\% & \color{red2} $0.653_{\downarrow 26.4\%}$ & \color{red2} $0.695_{\downarrow 22.6\%}$ & \color{red2} \textbf{0.9955} & \color{red2} 0.0072 & \color{red2} 13.6\%\\
\rowcolor{gray!25}
    & \textit{\THEMODEL${_{\lambda=.8}}$}&  $\mathbf{0.225}_{\downarrow 56.6\%}$& $\mathbf{0.420}_{\downarrow 45.6\%}$& \textbf{0.9917}& \textbf{0.0075}
 &\textbf{49.1\%}& $\mathbf{0.139}_{\downarrow 77.8\%}$& $\mathbf{0.256}_{\downarrow 66.5\%}$& $\textbf{0.9955}$ & \textbf{0.0068}
 &\textbf{55.4\%}\\

 \midrule
  \multirow{7}{*}{Yolov5-m} & Clean & $0.864$& $0.907$&  --& -- &--& $0.920$& 	$0.924$& --& -- &--\\
    & Noise & $0.840_{\downarrow 2.4\%}$&	$0.892_{\downarrow 1.5\%}$& 0.9892& \underline{0.0137} &2.1\%
& $0.837_{\downarrow 8.3\%}$& $0.848_{\downarrow 7.6\%}$& 0.9951& 0.0068
 &3.6\%\\
 & PGD& $\underline{0.372}_{\downarrow 49.2\%}$& $\underline{0.488}_{\downarrow 41.8\%}$& 0.9884& 0.0216
 &\textbf{37.6\%}& $0.364_{\downarrow 55.7\%}$& $0.420_{\downarrow 50.4\%}$& 0.9948&0.0130
 &25.3\%\\
    & \textit{T-sea} &  $0.565_{\downarrow 29.9\%}$& $0.610_{\downarrow 29.7\%}$& \underline{0.9900}& 0.0139
 &17.9\%& $\underline{0.350}_{\downarrow 57.0\%}$& $\mathbf{0.372}_{\downarrow 55.1\%}$&  0.9953& 0.0080
 &\underline{26.6\%}\\
    & \textit{GAN-NAP} & $0.687_{\downarrow 17.7\%}$& $0.772_{\downarrow 13.5\%}$& 0.9896& 0.0138
 &12.3\%
&  $0.517_{\downarrow 40.3\%}$& $0.558_{\downarrow 36.6\%}$& 0.9954 & 0.0086
 &19.2\%\\
 & \textcolor{red2}{\textit{CAP}} & \color{red2} $0.521_{\downarrow 34.4\%}$& \color{red2}$0.659_{\downarrow 24.7\%}$& \color{red2}0.9899& \color{red2}\underline{0.0137}& \color{red2}26.4\% & \color{red2} $0.545_{\downarrow 37.5\%}$ & \color{red2} $0.583_{\downarrow 34.1\%}$ & \color{red2} \underline{0.9957} & \color{red2} \underline{0.0060} & \color{red2} 13.8\%\\
\rowcolor{gray!25}
    & \textit{\THEMODEL${_{\lambda=.8}}$} & $\mathbf{0.306}_{\downarrow 55.8\%}$& $\mathbf{0.427}_{\downarrow 47.9\%}$&\textbf{ 0.9918}& \textbf{0.0073}
 &\underline{36.7\%}& $\mathbf{0.323}_{\downarrow 59.7\%}$& $\underline{0.388}_{\downarrow 53.6\%}$& \textbf{ 0.9960}& \textbf{0.0049}
 &\textbf{28.1\%}\\
 
    \midrule
 
   \multirow{7}{*}{Yolov3-m}& Clean &   $0.919$& $0.932$&--& -- &--& $0.937$& $0.938$& --& -- &--\\
    & Noise & $0.898_{\downarrow 2.1\%}$& $0.914_{\downarrow 1.8\%}$& 0.9891& 0.0142
 &1.5\%& $0.895_{\downarrow 4.3\%}$& $0.898_{\downarrow 4.0\%}$& 0.9950&	\textbf{0.0072}
 &1.7\%\\
 & PGD& $\underline{0.323}_{\downarrow 59.6\%}$& $\underline{0.391}_{\downarrow 54.0\%}$& 0.9882& 0.0219
 &39.6\%& $\underline{0.306}_{\downarrow 63.1\%}$& $\underline{0.338}_{\downarrow 60.1\%}$& 0.9946 & 0.0140
 &\underline{27.8\%}\\
    & \textit{T-sea} &  $0.679_{\downarrow 24.0\%}$& $0.728_{\downarrow 20.3\%}$& \underline{0.9896}& \underline{0.0153}
 &14.7\%& $0.774_{\downarrow 16.3\%}$& $0.784_{\downarrow 15.4\%}$& \underline{0.9953} & \underline{0.0083}
 &5.2\%\\
    & \textit{GAN-NAP} & $0.659_{\downarrow 26.0\%}$&  $0.692_{\downarrow 23.9\%}$& 0.9893& 0.0181
 &13.3\%& $0.640_{\downarrow 29.8\%}$& $0.663_{\downarrow 27.5\%}$& \textbf{0.9954}& 0.0108
 &16.0\%\\
 & \textcolor{red2}{\textit{CAP}} & \color{red2} $0.375_{\downarrow 54.4\%}$& \color{red2}$0.457_{\downarrow 47.4\%}$& \color{red2}0.9889& \color{red2}0.0163& \color{red2}\underline{42.6\%} & \color{red2} $0.464_{\downarrow 47.3\%}$ & \color{red2} $0.500_{\downarrow 43.9\%}$ & \color{red2} 0.9952 & \color{red2} 0.0085 & \color{red2} 23.1\%\\
\rowcolor{gray!25}
    & \textit{\THEMODEL$_{\lambda=.8}$}& $\mathbf{0.116}_{\downarrow 80.3\%}$& $\mathbf{0.159}_{\downarrow 77.2\%}$& \textbf{0.9903}&\textbf{ 0.0118}
 &\textbf{47.1\%}& $\mathbf{0.089}_{\downarrow 84.8\%}$& $\mathbf{0.124}_{\downarrow 81.4\%}$& 0.9950& 0.0096
 &\textbf{45.3\%}\\

    \midrule
    \multirowcell{7}{Faster-\\ RCNN}& Clean &  $0.920$& $0.927$& --& -- &--&$0.944$& $0.944$& --& -- &--\\
    & Noise & $0.909_{\downarrow 1.1\%}$& $0.916_{\downarrow 1.0\%}$& 0.9891& 0.0138
 &0.5\%& $0.914_{\downarrow 3.0\%}$& $0.917_{\downarrow 2.7\%}$& 0.9951&  0.0068
 &2.0\%\\
 & PGD& $\underline{0.478}_{\downarrow 44.2\%}$& $\underline{0.493}_{\downarrow 43.4\%}$& 0.9884& 0.0201
 &\underline{13.0\%}& $\underline{0.369}_{\downarrow 57.5\%}$& $\underline{0.378}_{\downarrow 56.6\%}$& 0.9948&0.0124
 &\underline{17.3\%}\\
    & \textit{T-sea} &  $0.831_{\downarrow 8.9\%}$& $0.844_{\downarrow 8.3\%}$& \underline{0.9901}& \underline{0.0129}
 &4.0\%&  $0.601_{\downarrow 34.3\%}$& $0.614_{\downarrow 33.0\%}$& 0.9952& 0.0097
 &15.5\%\\
    & \textit{GAN-NAP} & $0.753_{\downarrow 16.8\%}$& $0.768_{\downarrow 15.8\%}$& 0.9900& 0.0144
 &8.1\%& $0.834_{\downarrow 11.0\%}$& $0.839_{\downarrow 10.5\%}$& \textbf{0.9958}& \textbf{0.0057}
 &7.9\%\\
 & \textcolor{red2}{\textit{CAP}} & \color{red2} $0.802_{\downarrow 11.9\%}$& \color{red2}$0.816_{\downarrow 11.1\%}$& \color{red2}0.9900& \color{red2}0.0137& \color{red2}5.6\%& \color{red2} $0.861_{\downarrow 8.3\%}$ & \color{red2} $0.865_{\downarrow 7.9\%}$ & \color{red2} \underline{0.9957} & \color{red2} \underline{0.0061} & \color{red2} 3.6\%\\
\rowcolor{gray!25}
    & \textit{\THEMODEL${_{\lambda=.8}}$}&    $\mathbf{0.383}_{\downarrow 53.7\%}$& $\mathbf{0.398}_{\downarrow 52.9\%}$& \textbf{0.9911}&  \textbf{0.0100}
 &\textbf{18.5\%}& $\mathbf{0.235}_{\downarrow 70.9\%}$& $\mathbf{0.244}_{\downarrow 70.0\%}$& 0.9953&  $0.0082$ &\textbf{30.1\%}\\

    \midrule
    
    \multirowcell{8}{\color{red}RT-\\DETR\\ -L} & 
Clean &  $0.942$& $0.958$& --& -- & -- &$0.932$& $0.932$& --& -- &--\\ 
    & Noise & ${0.934}_{\downarrow 0.7\%}$& ${0.951}_{\downarrow 0.6\%}$& 0.9892& \underline{0.0137}&0.6\%& ${0.905}_{\downarrow 2.7\%}$& ${0.912}_{\downarrow 2.0\%}$& 0.9950&  \textbf{0.0070}&2.9\%\\ 
 & PGD& ${0.814}_{\downarrow 12.8\%}$& ${0.858}_{\downarrow 10.0\%}$& 0.9884& 0.0223&9.4\%& ${0.750}_{\downarrow 18.2\%}$& ${0.763}_{\downarrow 16.9\%}$& 0.9948&0.0131&11.5\%\\
    & \textit{T-sea} &  ${0.933}_{\downarrow 0.9\%}$& ${0.951}_{\downarrow 0.7\%}$& 0.9888& 0.0161&0.7\%&  ${0.901}_{\downarrow 3.0\%}$& ${0.908}_{\downarrow 2.4\%}$& 0.9949& 0.0092&3.5\%\\
    & \textit{GAN-NAP} & ${0.933}_{\downarrow 0.9\%}$& ${0.950}_{\downarrow 0.8\%}$& \underline{0.9897}& 0.0141&0.6\%& ${0.894}_{\downarrow 3.8\%}$& ${0.901}_{\downarrow 3.1\%}$& \textbf{0.9954}& 0.0090&3.6\%\\
 & \textcolor{red2}{\textit{CAP}} & \color{red2} $0.932_{\downarrow 1.0\%}$& \color{red2}$0.950_{\downarrow 0.8\%}$& \color{red2}0.9894& \color{red2}0.0145& \color{red2}0.9\% & \color{red2}$0.908_{\downarrow 2.4\%}$ & \color{red2} 	$0.914_{\downarrow 1.8\%}$ & \color{red2} 0.9952 & \color{red2} \underline{0.0076} & \color{red2} 2.8\%\\
\rowcolor{gray!25}
    & \textit{\THEMODEL${_{\lambda=.8}}$}&    $\underline{0.756}_{\downarrow 18.5\%}$& $\underline{0.815}_{\downarrow 14.3\%}$& \textbf{0.9915}&  \textbf{0.0101}&\underline{14.4\%}& $\underline{0.617}_{\downarrow 31.5\%}$& $\underline{0.645}_{\downarrow 28.7\%}$& \underline{0.9953}&  0.0090&\underline{19.8\%}\\
\rowcolor{gray!25}
    & \textit{\THEMODEL${_{\lambda=1.}}$}&    $\mathbf{0.573}_{\downarrow 36.9\%}$& $\mathbf{0.659}_{\downarrow 29.9\%}$& 0.9891&  0.0164&\textbf{31.2\%}& $\mathbf{0.585}_{\downarrow 34.7\%}$& $\mathbf{0.615}_{\downarrow 31.7\%}$ & 0.9951&  0.0101&\textbf{21.6\%}\\

    \bottomrule
    \multicolumn{12}{l}{\textcolor{red2}{The values bolded and underlined indicate the best and the second-best results, respectively. Subscripts show change relative to ``Clean''.}} 
    \end{tabular}
  }
  \label{tab:main_results}
\end{table*}

\begin{figure*}[t]
    \centering
    \includegraphics[width=\linewidth]{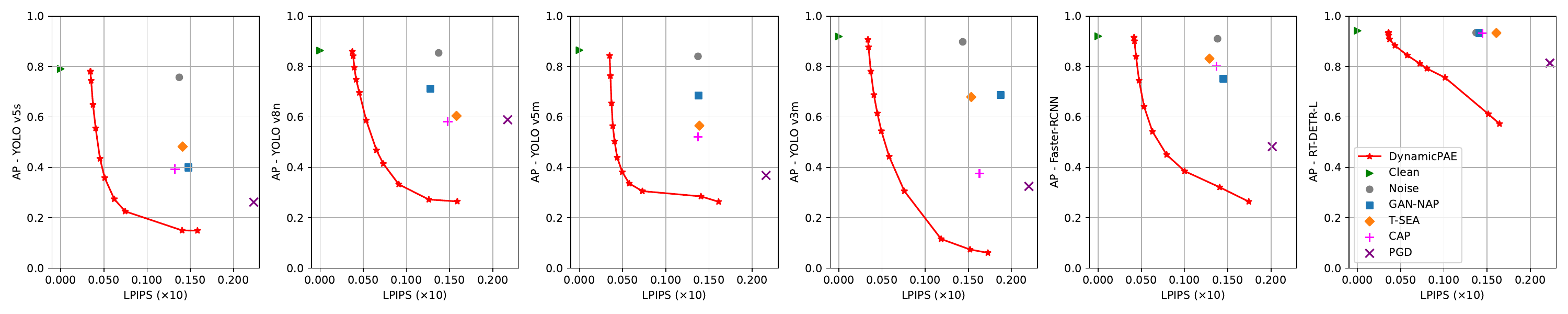}
    \vspace{-2em}
    \caption{\textcolor{red2}{Attack aggressiveness (measured by AP$_{50}$, lower is better) \& visual distortion (measured by LPIPS, higher is better) evaluated on COCO dataset, corresponding to Table~\ref{tab:main_results}. Our framework \THEMODEL, with test-time settings $\lambda\in\{0.0, 0.1, ..., 1.0\}$, significantly pushes the limit of aggressiveness-distortion trade-off.}}
    \vspace{-0.5em}
    \label{fig:ad_curve}
\end{figure*}

\subsection{Performances Evaluation}



\textcolor{red2}{$\bullet$ \textbf{Attack Performance:}  We evaluate performance based on settings similar to classical PAE studies~\cite{advpatch}, and the stealthiness metrics are included to enhance the fairness for biased optimization. We trained our model on the \textit{COCO} dataset only and others on data identical to the evaluation. $\lambda$ is the inference-time parameter controlling the stealthiness.
The results are shown in Table~\ref{tab:main_results}, demonstrating the \textbf{significant superiority} of \THEMODEL\ on the attack performance.
Specifically, \textit{\THEMODEL${_{\lambda=.8}}$} boost the average drop of $AP_{50}$ to \textcolor{red}{$-58.8\%$} (\textcolor{red}{$2.07\times$} on average) while preserving comparable stealthiness measured by SSIM, and only under one configuration (Yolov5-m, Inria) can \textit{T-sea} achieve comparable aggressiveness. Furthermore, our model achieves $1.3\times$ of $AP_{50}$ average drop compared to performing PGD on each image.
In particular, for models with larger parameter sizes (\textit{Yolov3-m}, \textit{Faster-RCNN} and \textit{RT-DETR-L}), our approach exhibits more significant attack performance superiority, suggesting the scalability.
Furthermore, our method has not been trained on the Inria dataset, indicating its adaptation capability.
The \textit{trade-off} between the attack performance and the distortion of the original image is shown in Figure~\ref{fig:ad_curve}.
Since our model can perform generation conditioned on ${\lambda}$ and automatically balance the attack and invisibility by \textit{Skewness-Aligned Objective Re-weighting}, the curve in each sub-graph is evaluated in just one training session with the test-time settings of $ \lambda = \{0.0, 0.1, 0.2, ..., 1.0\}$.
However, the static PAE methods are only able to produce a single result, and thus, our model achieves better practicality in PAE-driven model evaluation.
We evaluate more transformation settings in the \textit{supplementary material} and have the same result, except \textit{PGD} performs better in zero spatial transformation.
}
\textcolor{red}{
We acknowledge that with a sufficiently large number of iterations, under the same distribution of $\boldsymbol{\oplus}$, and the same image, \textit{PGD} may be able to achieve better attack results, but at an unacceptable time overhead.}

\textcolor{red2}{
Note that the experiment is conducted on the white-box scenario to focus on comparing physical robustness and attack capability.
The evaluation of \textit{black-box transfer attack}, although \THEMODEL\ has not been designed for, is shown in the \textit{supplementary material}.
The results show that (1) when using the robust surrogate model, \textit{e.g.}, \textit{RT-DETR-L}, the superiority of \THEMODEL\ is still maintained in the transfer attack scenario. (2) When using the non-robust light-weighted surrogate model, \textit{e.g.}, \textit{Yolo-v5s}, \textit{T-sea} may achieve slightly better transferability. The lower transferability of \THEMODEL\ in this case may be attributed to the higher \textit{model complexity of PAE}.
Given the greater superiority of (1) (ASR +$18.4\%$) than (2) (ASR +$6.2\%$), we believe \THEMODEL\ is also applicable to transfer attacks, and regularization may further improve the performance.
}


\noindent
\textcolor{red}{$\bullet$ \textbf{Speed Evaluation}:
The total Floating Point Operations (\textit{FLOPs}) and Parameters of the $\mathcal{G}$ that \THEMODEL\ framework successfully trained for attacking object detectors is only 5.45G and 29.29M, respectively, which ensures efficient inference on resource-limited edge devices. (The \textit{FLOPs} of \textit{ResNet-50} is 4.09G).
We present a comparison of time consumption in real devices in Table~\ref{tab:time_bench}.}
All of the training and validation except \textit{\THEMODEL}$_{Inf.}$ is done on one Nvidia A40 with bf16 enabled.
\textit{\THEMODEL}$_{Val.}$, \textit{\THEMODEL}$_{Inf.}$ and \textit{PGD} are timed through \textit{torch.cuda.Event} after \textit{CUDA} warmup, and they are tested on the Inria dataset with multiple runs.
\textit{\THEMODEL}$_{Val.}$ achieves a speedup of more than \textbf{2000} times compared to generating the PAE for each image with \textit{PGD} and maintains higher attack performance.
\textit{\THEMODEL}$_{Inf.}$ shows that our model can provide real-time patch generation even on personal graphics cards. 
Evaluation of more \textit{PGD} parameters is shown in section~\ref{app:pdg}.

\begin{table}[t]
    \caption{Evaluation of GPU processing time. PGD and \textit{\THEMODEL}$_{Val.}$ denote the average time of PAE generation on a batch of images with \textit{batchsize = 32}, and \textit{\THEMODEL}$_{Inf.}$ shows the single-image inference latency. Static PAEs ($\textit{T-sea}$ and $\textit{GAN-NAP}$) do not require inference computation, and thus they are not included.}
    \centering
    \begin{tabular}{c|cccc}
    \toprule
         &  Yolov5-s&  Yolov8-n&  Yolov3-m& Yolov5-m\\
         \midrule
         PGD & 288.3s& 175.1s& 702.5s&451.0s\\
         \textit{\THEMODEL}$_{Val.}$&  \multicolumn{4}{c}{87.36ms}\\
         \textit{\THEMODEL}$_{Inf.}$ & \multicolumn{4}{c}{\color{red} 10.95 ms (Laptop RTX4050, 50W power limit)}\\
 \bottomrule
    \end{tabular}
    \label{tab:time_bench}
    \vspace{-1em}
\end{table}

\begin{table}[t]
    \caption{\textcolor{red}{Comparison between training time and performance indicates the superiority of \THEMODEL\ in the overall adversarial evaluation efficiency}.}
    \centering
    \begin{tabular}{cc|cc|cc}
    \toprule
 \multicolumn{2}{c|}{\textit{T-sea}}& \multicolumn{2}{c|}{\textit{GAN-NAP}} &\multicolumn{2}{c}{\THEMODEL} \\
  $AP_{50}$\color{red}{↓}& Time\color{red}{↓}& $AP_{50}$\color{red}{↓}& Time\color{red}{↓}& $AP_{50}$\color{red}{↓}&Time\color{red}{↓}\\
 \midrule
 58.9\%&0.7h&44.8\%&0.2h&28.8\%& 0.7h\\
 57.0\%& 1.3h& 40.6\%& 0.5h& 21.5\%&1.4h\\
 53.0\%& 2.7h& 39.8\%& 1.0h& 19.1\%&2.8h\\
 48.9\%&5.4h&40.2\%&2.0h&17.8\%& 5.7h\\
 48.3\%&10.8h&39.6\%&3.9h&17.3\%& 11.3h\\
 47.9\%&21.6h&39.6\%&7.8h&14.7\%& 22.6h\\
 \bottomrule
 
 \end{tabular}
    \label{tab:training_time_bench}
    \vspace{-1em}
\end{table}

\textcolor{red}{Considering the overall time consumption  when evaluating the target model on the scenario defined by the task and the dataset, \THEMODEL\ still achieves superiority.
We evaluate the training time when targeting Yolo-v5s on the \textit{COCO} dataset, and show the performance of different frameworks with different training iterations compared to full training in Table~\ref{tab:training_time_bench}.
All training is done on a single A800 card, and the adaptive learning rate is performed by Adam.
Although \THEMODEL\ needs to train the network, the time consumption is comparable, and its attack performance under the same training time is significantly higher.
This is because both static attacks and the generative attacks in our framework require running backpropagation algorithms on the computational graph of the target model, whose overhead is comparable in magnitude to that of optimizing $\mathcal{G}$, as the \textit{FLOPs} of evaluated detectors ranges from 8.7G (Yolo-v8n) to 267.37G (Faster-RCNN), which is much higher due to the high-resolution input settings of detectors. Also, the efficiency of our training framework benefits from the single forward-backward design shown in Algorithm~\ref{alg:main}.}

 
    

\subsection{On Overcoming the Noisy \textcolor{red}{Feedback in Training}} 

Existing works have shown the special characteristics of AEs and the adversarial attack problems, including \textcolor{red}{AE's} physical robustness~\cite{DBLP:conf/iclr/KurakinGB17a}, distributional properties~\cite{Song2017PixelDefendLG,ma2018characterizing, Wu24NAPGuard}, and gradient properties of the optimization process~\cite{athalye2018obfuscated}.
\textcolor{red}{For the dynamic PAE training, we identify the key challenge as \textit{noisy feedback}, which causes the problem of \textit{degenerated solution}. We model the problem with the \textit{limited feedback information restriction} and construct the technical solution based on it. This section explores the impact of noisy gradient feedback and the effectiveness of our model}. 


\noindent
\textcolor{red}{$\bullet$ \textbf{Noisy Feedback Causes PGD Attacks Time-Consuming:}}
\label{app:pdg}
\textcolor{red}{We first show that the simulated physical adversarial patch attacks, as a typical physical attack, can be regarded as facing the {obfuscated gradient}~\cite{athalye2018obfuscated}, or noisy gradient feedback problem indicated by our experimental results, as the spatial transformations significantly increase the difficulty of PGD attacks.
{Obfuscated gradient} has been recognized as a typical adversarial defense mechanism that increases the difficulty of performing optimization by adding a random transformation.}
\textcolor{red2}{
The randomness in the physical simulation, introduced as the EoT~\cite{EOT} transformation in PAE methods, has the same effect.
Specifically, when randomness is not injected, Table~\ref{tab:apppgdzero} shows that PGD achieves high attack performance when the step number reaches 32. However, when randomness is injected with \textit{Base} transformation in Table~\ref{tab:apppgdv5}, an extensive number of total steps, whether it’s EoT iteration steps or the optimization steps, is needed to achieve satisfactory attack performance.
This illustrates the challenges of PGD optimization in the presence of noisy gradient feedback resulting from random simulation.
}
\begin{table}[h]
    \centering
    \caption{Comparison with PGD in the patch attack on Yolo-v5s, evaluated with \textit{zero} transformation in Table~\ref{tab:transformation} and \textit{Inria} dataset.}
    \resizebox{\linewidth}{!}{
    
    \begin{tabular}{c|cc|ccc}
    \toprule
         Method &  EoT steps&  Optim. steps&  $AP_{50} $\color{red}{↓} & $AP_{01}$\color{red}{↓} &time\color{red}{↓} \\
         \midrule
         Clean&  --&  --&  79.0\%& 87.6\%&--\\
         \hline
         \multirow{4}{*}{PGD}&  1&  8&  48.0\%& 53.9\%&1.20s
\\
         &  1&  32&  11.5\%& 24.6\%&4.55s
\\
 & 1& 128& 3.4\% & 15.2\% &17.97s
\\
 & 1& 512& 1.7\% & 11.0\% &71.50s
\\
         \hline
 \textit{\THEMODEL}& --& --& 13.7\%& 24.8\%&0.09s\\
 \bottomrule
    \end{tabular}
    }
    
    \label{tab:apppgdzero}
\end{table}

\begin{table}[h]
    \vspace{-0.5em}
    \centering
    \caption{\textcolor{red2}{Comparison with PGD in the patch attack on Yolo-v5s and Yolo-v8n, evaluated with \textit{base} transformation in Table~\ref{tab:transformation} and more complex \textit{COCO} dataset.}}
    \resizebox{\linewidth}{!}{
    \color{red2}
    \begin{tabular}{c|cc|ccc|ccc}
    \toprule
         & \multirowcell{2}{EoT\\Steps} &  \multirowcell{2}{Optim.\\Steps}&  \multicolumn{3}{c}{Yolo-v5s} & \multicolumn{3}{c}{Yolo-v8n}\\
        Method &&&  $AP_{50}  $\color{red}{↓}& $AP_{01} $\color{red}{↓} &time\color{red}{↓} & $AP_{50}  $\color{red}{↓}& $AP_{01} $\color{red}{↓} &time\color{red}{↓}\\
         \midrule
         Clean&  --&  --&  79.0\%& 87.6\%&-- & 86.4\%& 92.1\%&--\\    
         \hline
         \multirow{6}{*}{PGD}&  1&  128&  50.4\%& 65.7\%&18.6s
 & 73.8\%& 83.5\%&11.4s \\
         &  4&  128&  38.1\%& 55.2\%&73.8s
 &   65.9\%& 77.6\%&44.9s \\
 & 1& 512& 38.0\%& 54.5\%&74.1s
 &  65.1\%& 77.1\%&45.5s \\
 & 16& 128& 29.0\%& 45.2\%&230.3s
 & 60.2\%& 73.9\%&140.1s\\
 & 4& 512& 27.5\%& 43.3\%&295.2s
 &  58.4\%& 71.7\%&180.0s\\
 & 1& 2048& 27.0\%& 42.3\%&296.5s
 & 57.7\%& 71.5\%&181.5s\\
         \hline
 \textit{T-sea}& --& --& 48.2\%& 63.1\%&-- & 60.4\%& 75.0\%&--\\
 \textit{\THEMODEL}& --& --& 16.3\%& 28.8\%&0.1s &27.9\%& 37.8\%&0.1s\\
 \bottomrule
    \end{tabular}
    }
    \label{tab:apppgdv5}
\end{table}

    
         

\textcolor{red}{
We present the performance of static PAE (\textit{T-sea}) and \THEMODEL\ framework for comparison.
To ensure a fair comparison, both our model and PGD are evaluated by resampling the patch locations (same setting as Table~\ref{tab:location_roubustness}).}
\textcolor{red2}{
The result shows that PGD shall take hundreds of steps to surpass the performance of static PAE under the \textbf{base} transformation, indicating its incapability of real-time attacks. }

\begin{figure}[h]
    \vspace{-1em}
\subfloat[w/o residual task.]{
                \includegraphics[width=0.32\linewidth]{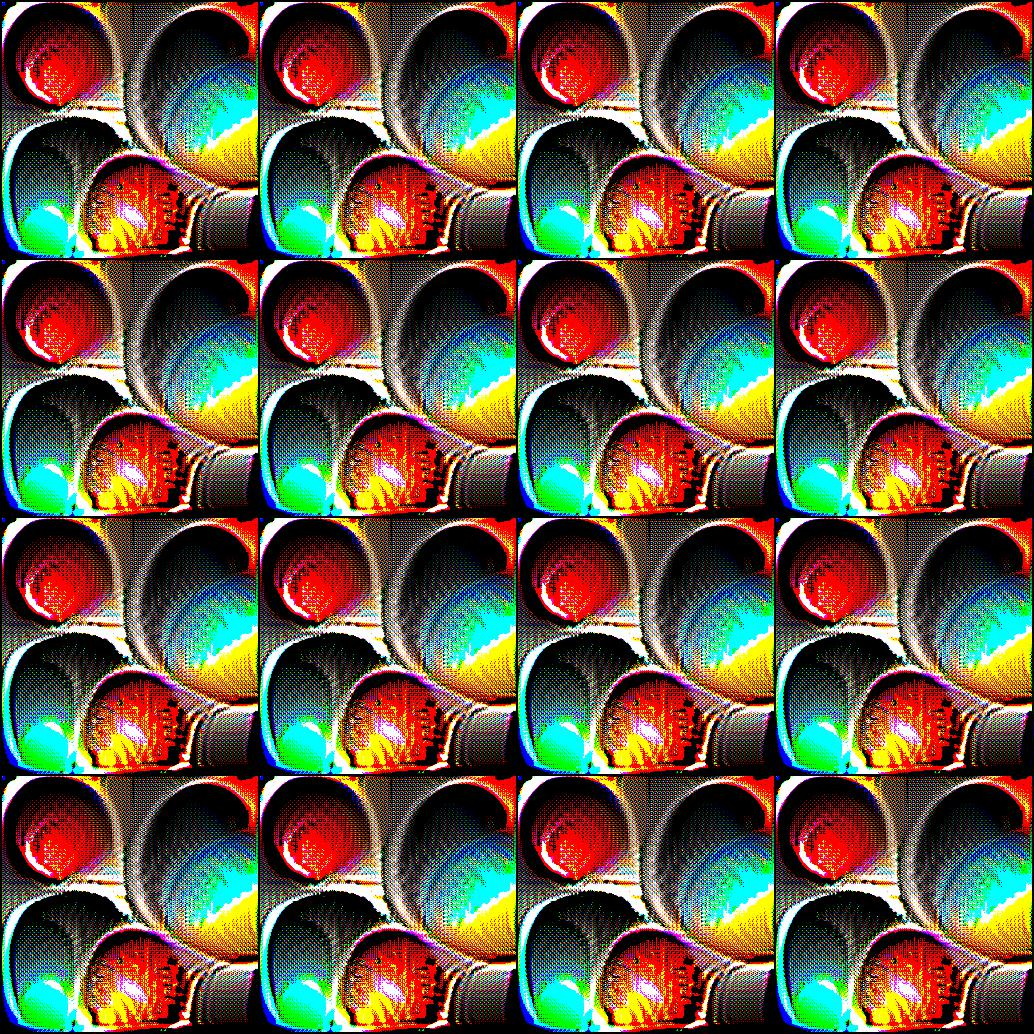}
                \label{fig:patches_before_aux_task}
    \
    }
\subfloat[with residual task.]{
                \includegraphics[width=0.32\linewidth]{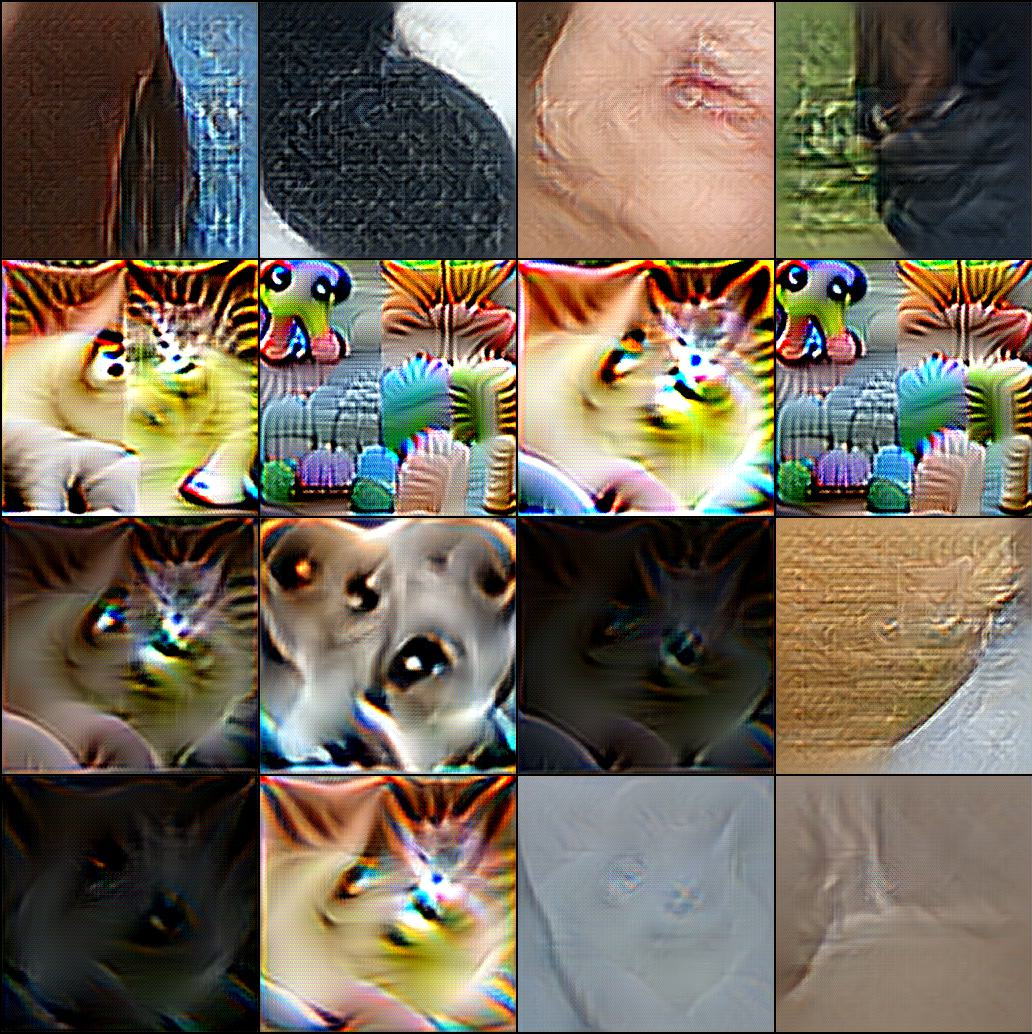}
                \label{fig:patches_after_aux_task}
    \
    }
    \centering
    \vspace{-0.5em}
    \caption{Comparison of Patch Generation. \textcolor{red}{After applying the residual task, the mode collapse problem in training the dynamic PAE generator is solved.}}
    \vspace{-0.5em}
\end{figure}

\noindent $\bullet$ \textcolor{red}{\textbf{Noisy Feedback Causes Generative Training Degenerate:}}
\label{app:mode_vollapse}
\textcolor{red}{When training PAE generators, we find that the new challenge caused by the noisy gradient feedback is the degenerated solution presented as mode collapse, and the theoretical modeling is constructed in section~\ref{sec:aux_task}.}
We visualized the training result of the generated PAEs on the setting of \textit{Yolov5}, \textit{base} transformation, and COCO dataset. The results of training without the auxiliary residual task are shown in Figure~\ref{fig:patches_before_aux_task}.
Although these PAEs achieve attack capabilities, they exhibit significant mode collapse with only pixel-level weak differences between different PAEs.
Figure~\ref{fig:patches_after_aux_task} illustrates the results under the same configuration after introducing the residual task.
The first row, the second row, and the last two rows represent the case with attack weight $\lambda=0$, $1$, and sampled between $[0, 1]$, respectively. \textcolor{red}{It can be observed that the diversity of the generations is significantly enhanced, especially when $\lambda=1$ (row 2 of Figure~\ref{fig:patches_before_aux_task} and~\ref{fig:patches_after_aux_task}), where the training objectives are equivalent.}

\begin{figure}[h]
    \centering
    \vspace{-0.5em}
    \includegraphics[width=\linewidth]{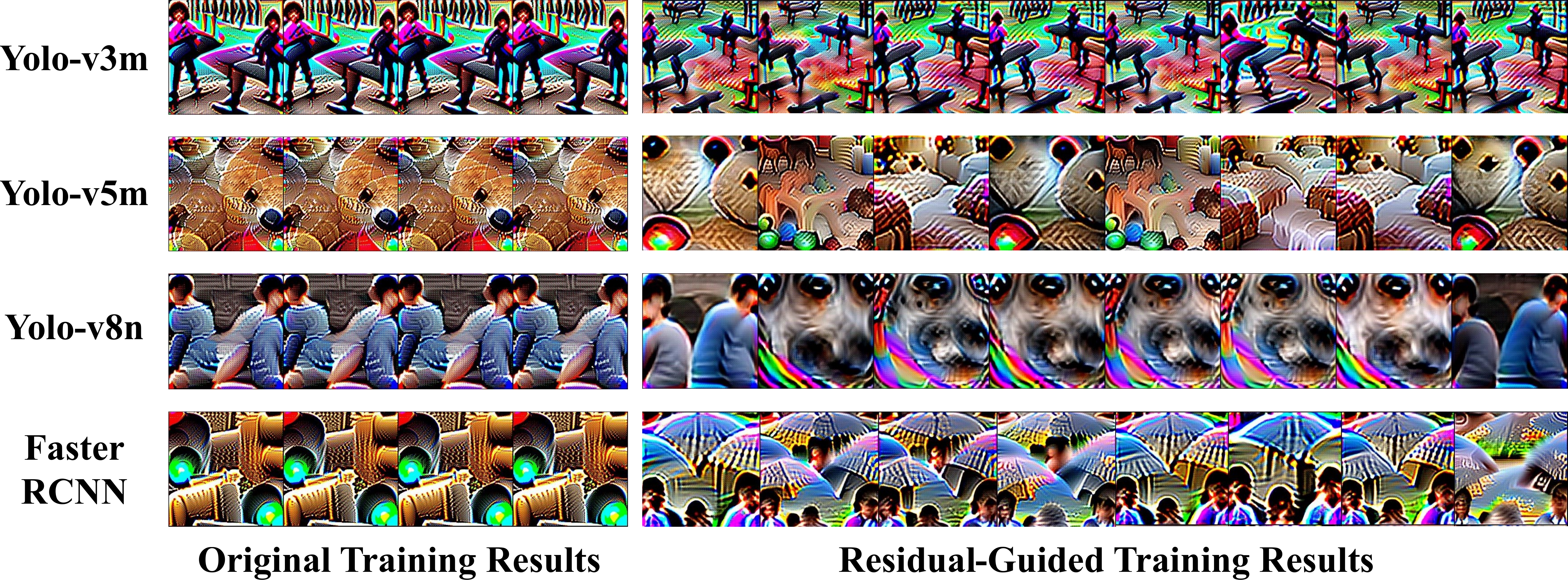}
    \vspace{-1em}
    \caption{\textcolor{red}{Patch Visualization for Different Targets. The original training results in identical patches, while the residual-guided training results in different patches.}}
     \vspace{-0.5em}
    \label{fig:diff_model_patches}
\end{figure}

\textcolor{red}{
We visualize the patch generation results for other target models in Figure~\ref{fig:diff_model_patches}.
Without the residual-guided training, the generated patches are identical in all four target models.
Residual-guided training successfully finds diverse solutions when attacking each model.
Although the explored PAE patterns are not diverse, the escape behavior of the degenerated and identical solution is consistent.
Exploring more diverse aggressive solutions might be a difficult task within limited training computation, as finding each pattern requires a large number of queries under the assumption of limited feedback information.
}

In addition, this mode collapse is different from the posterior collapse in variational autoencoders (\textit{VAE}), in which the noise is injected in latent $\mathbf{Z}$ for prior-based regularization instead of stemming from the task.
\textcolor{red}{
Moreover, although diversity is not the primary goal of the attack tasks, this mode collapse is weakening the attack performance of dynamic PAEs to be close to static PAEs, thus we recognize it as the degeneracy problem.
}

\begin{figure}[h]
    \centering
    \subfloat[PAE Representation Space]{
                \includegraphics[width=0.47\linewidth]{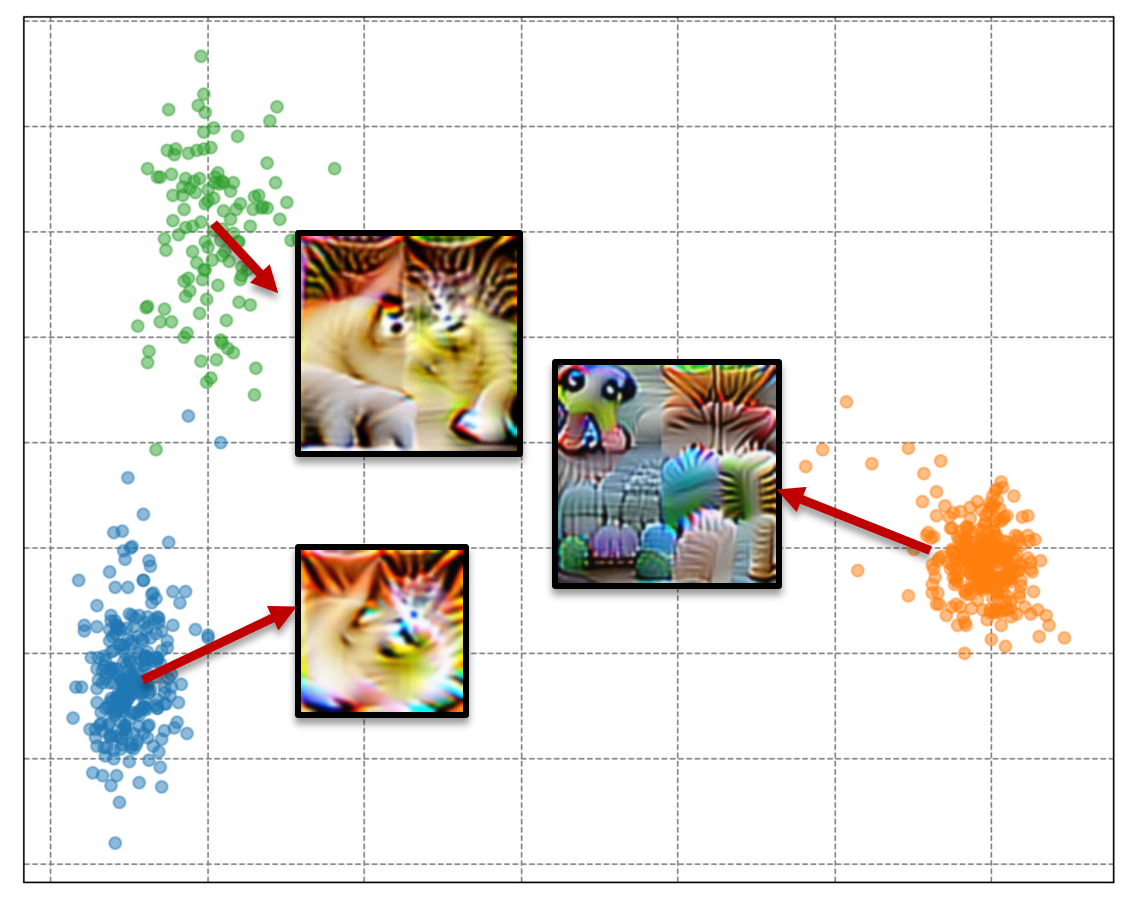}
                \label{fig:clust}
    \
    }
    \subfloat[Physical Contexts KNN]{
                \includegraphics[width=0.47\linewidth]{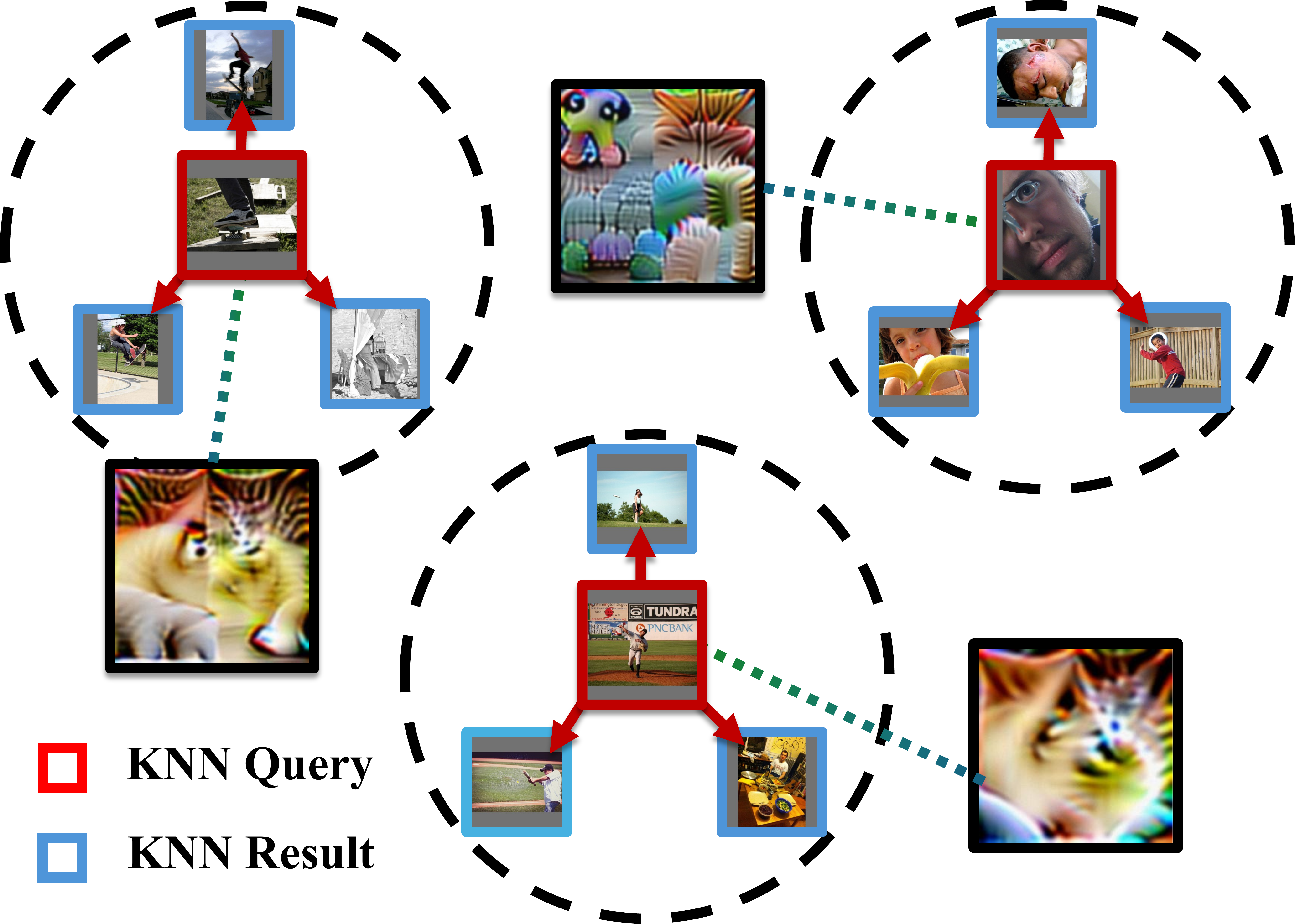}
                \label{fig:clust_context}
    \
    }
    
    \vspace{-0.25em}
    \caption{\textcolor{red2}{Visualization of the learned correlation between PAEs and the related physical environments (collected by latent-space K-nearest neighbor (KNN)).}}  
    \vspace{-1em}
\end{figure}

\noindent
\textcolor{red}{
$\bullet$ \textbf{Residual-Guided Training Overcomes the Degeneracy Problem:}
To show the different PAEs in Figure~\ref{fig:patches_after_aux_task} and~\ref{fig:diff_model_patches} is generated based on the \textit{attacker's observation} rather than randomly generated, we further analyze the latent representation $\mathbf Z$ of the trained generator.
We first annotate the latent representations based on the style of PAEs, and apply \textit{LDA} for dimension reduction.
The result is shown in Figure~\ref{fig:clust}, indicating that the model successfully learns a linearly decomposable representation of PAEs.
We conduct the K-nearest neighbor (KNN) search (k=3) in the dimension-reduced space and visualize the corresponding physical context in Figure~\ref{fig:clust_context}.
The outcomes demonstrate that the KNN result has significant similarity with the KNN query in terms of certain characteristics, including human actions, the chromatic attributes of clothing, and the exposed body parts~\emph{et al.}, indicating that our model successfully learns a distinction between $\mathbf P_\mathrm{X}$ based on the vulnerability of the target model by learning PAE generation.
}


\begin{figure}[h]
  \begin{minipage}{0.52\linewidth}
    \centering
    \includegraphics[width=\linewidth]{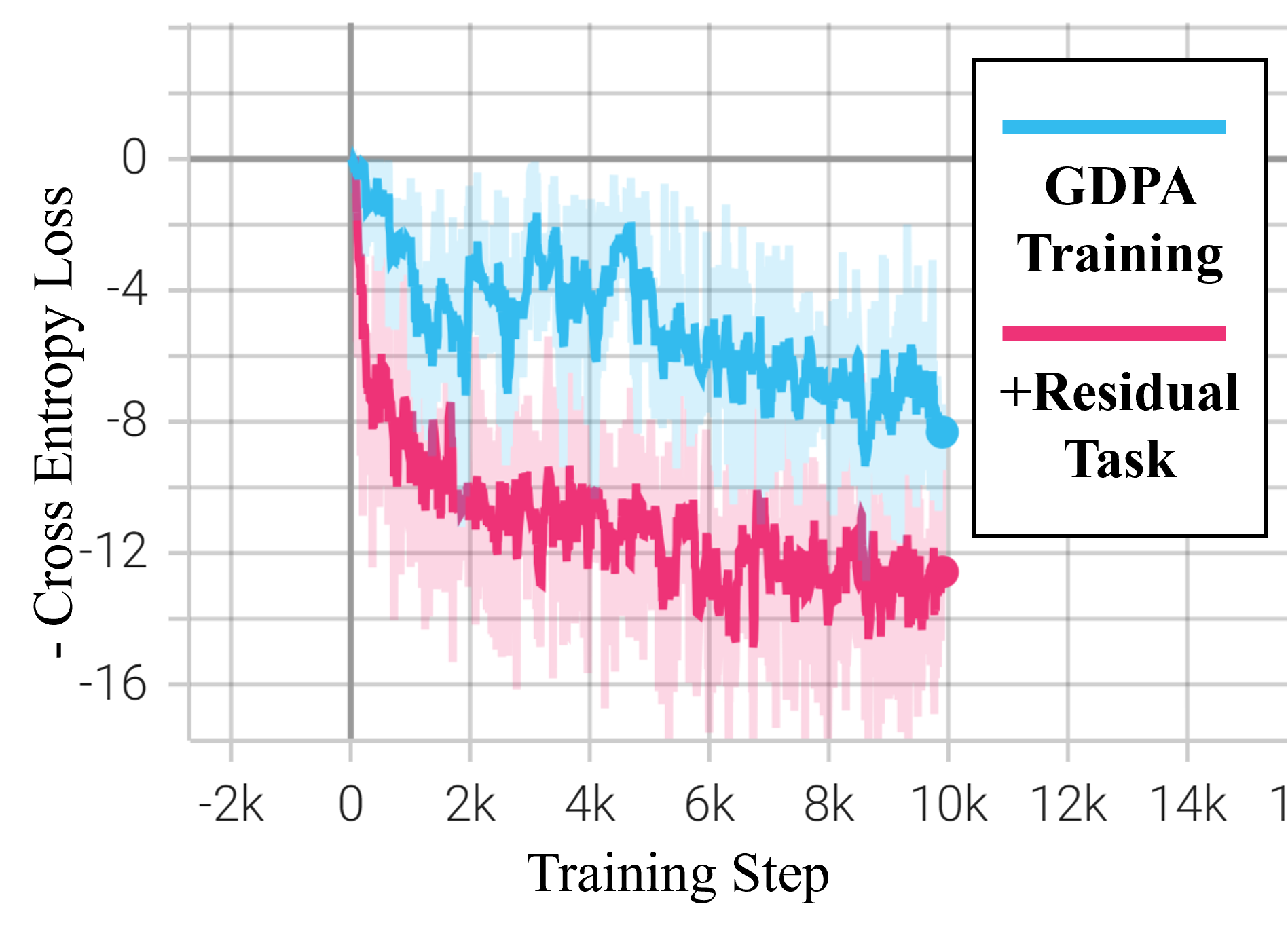}
    \vspace{-1.5em}
    \caption{\textcolor{red2}{Loss curves under different training strategies.}}
    \label{fig:gdpa_comparison}
  \end{minipage}
  \hfill
  \begin{minipage}{.45\linewidth}
    \makeatletter\def\@captype{table}\makeatother
    \caption{\textcolor{red2}{ASR on the VGGFace validation set. The residual task improves the training convergence.}}
    \label{tab:gdpa_comparision}
      \vspace{-0.5em}
    \resizebox{\linewidth}{!}{
      \color{red2}
      \begin{tabular}{ccc}
        \toprule Epoch & \textit{GDPA} & + Res. Task  \\
        \midrule
        1 & 5.1\% & 58.9\% \\ 
        2 & 6.2\% & 70.8\% \\
        4 & 20.9\% & 77.8\% \\
        25 & 57.1\%& 88.2\% \\
        50 &  42.3\%& 88.0\% \\
        100 & 78.9\% & 88.6\% \\
        \bottomrule
      \end{tabular}
      }
  \end{minipage}
     \vspace{-1em}
\end{figure}

\textcolor{red2}{The comparison of the attack performance is shown in ablation studies.
In addition to object detection tasks, \cite{Li2021GenerativeDP} attempted the generative approach for real-time digital adversarial patch attacks against image classification and face recognition. We observe that, even without incorporating randomized physical simulation and with patch sizes limited to 32$\times$32, it still suffers from certain degeneracy issues. We introduce the proposed training strategy into this model (detailed implementation is in \textit{supplementary material}) and test it with the 2\%-ratio patch attack setting and the VGGFace classification task. As shown in Figure~\ref{fig:gdpa_comparison} and Table~\ref{tab:gdpa_comparision}, the final ASR and training convergence are significantly improved ($+10\%$ ASR, $4\times$ training speed). This highlights the criticality of the proposed training technique.}

\textcolor{red}{
Overall, the experimental analysis shows that: \ding{182} The obfuscated gradient problem exists in the PAE training environment. \ding{183} \textit{limited feedback information restriction}, which models the noisy gradients and predicts the degeneracy problem of the conditional PAEs, matches the experiment results. \ding{184} By breaking the restriction of the \textit{limited feedback information}, the residual-task guided training overcomes the degeneration problem.
This result simultaneously demonstrates the effectiveness of the \textit{limited feedback information restriction} model and the residual-task construction. 
}

\subsection{On the Practicality in \textcolor{red}{Real} Attack Scenarios}
\textcolor{red}{To evaluate the real-world attack capability of \THEMODEL, we conduct experiments on different environments and parameters to show the effectiveness of the proposed alignment method.}
The experiments demonstrate the ability of \THEMODEL\ to (1) reveal potential security issues and \textcolor{red1}{(2) generate PAEs that are robust in the more complex deployment environments.}

\begin{figure}[h]
    \vspace{-0.5em}
    \centering
    \subfloat[\textit{GAN-NAP}]{
        \includegraphics[width=0.31\linewidth]{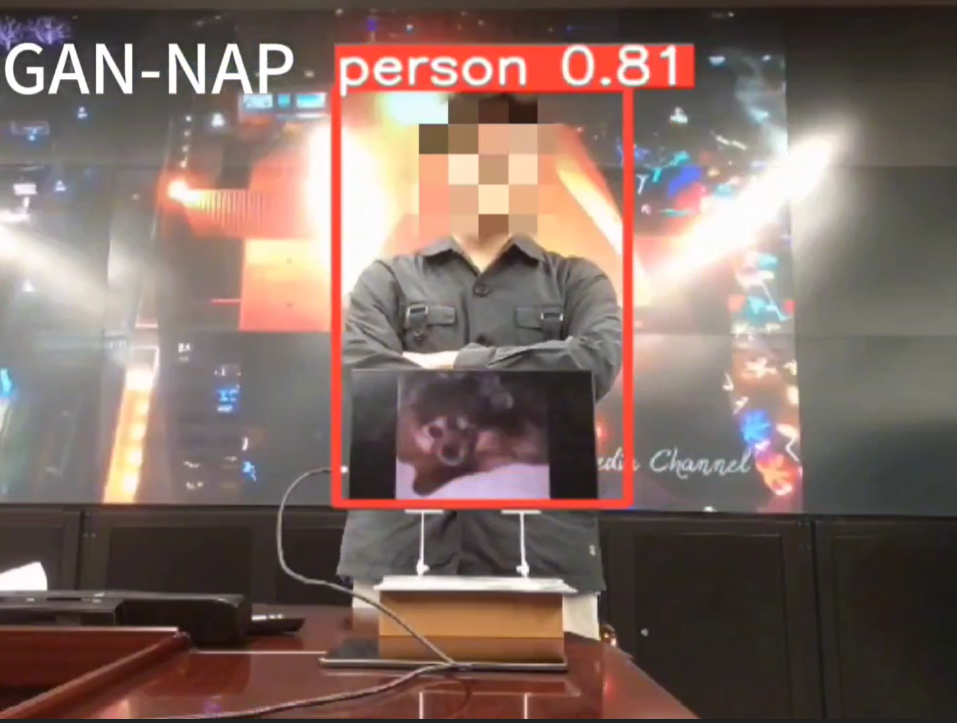}
        \label{fig:gan_nap_phy}}
    \subfloat[\textit{T-sea}]{
        \includegraphics[width=0.31\linewidth]{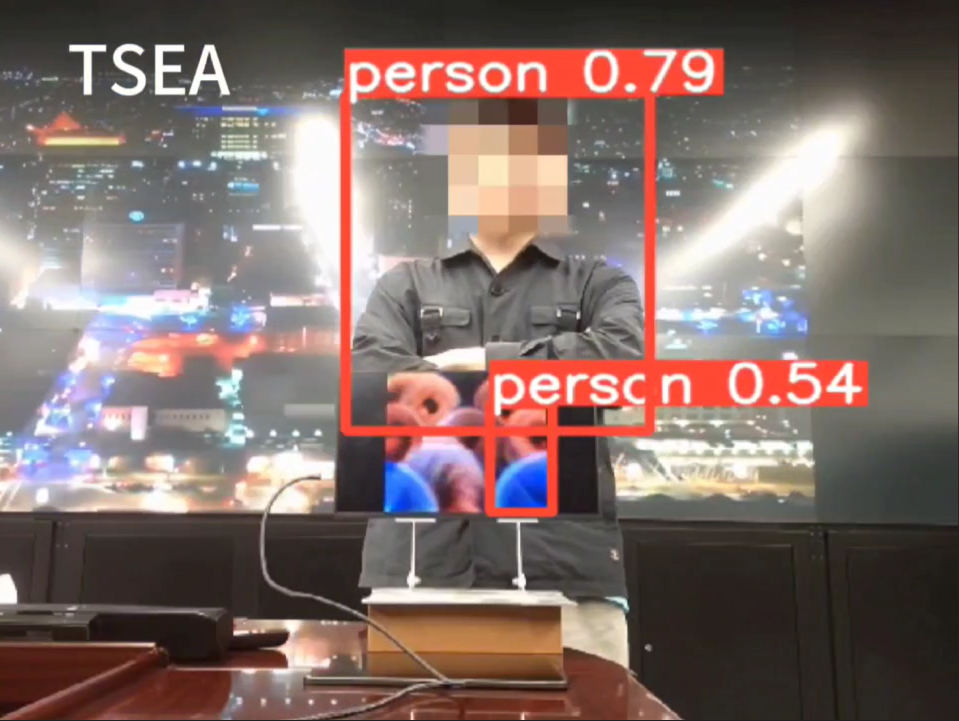}
        \label{fig:tsea_phy}}
    \subfloat[\textit{\THEMODEL}]{
        \includegraphics[width=0.31\linewidth]{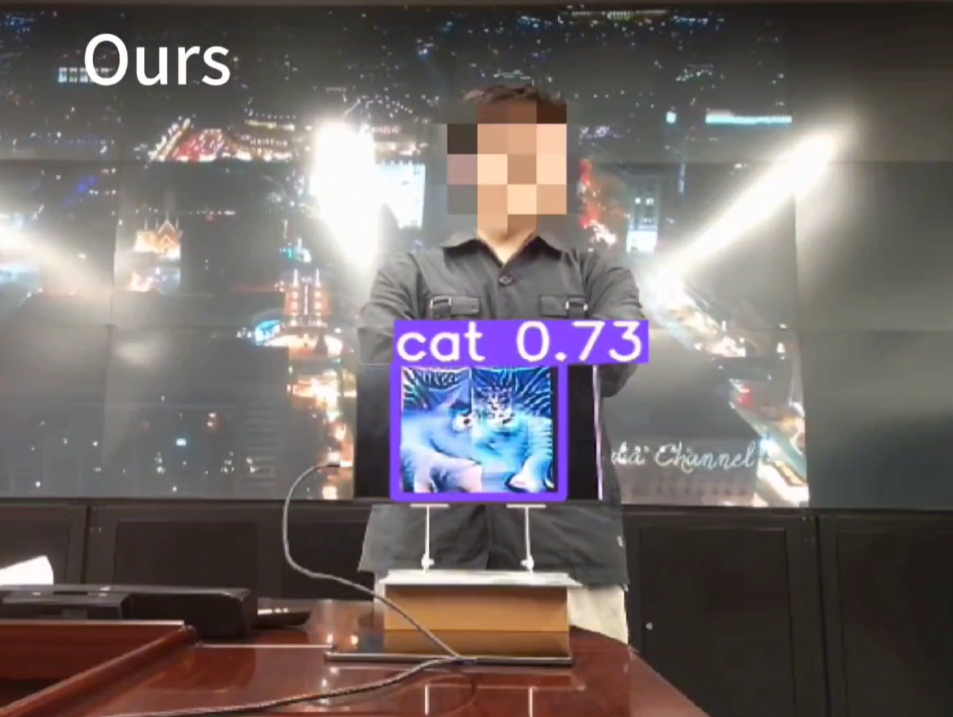}
        \label{fig:ours_phy}}
    \vspace{-0.5em}
    \caption{Our method achieves better attack performance in the scenario of the ever-changing background.}
    \vspace{-0.25em}
    \label{fig:phy_vis}
\end{figure}

\noindent
$\bullet$ \textbf{Results in Physical Environments:} We implement our model and deploy it \textcolor{red}{on laptop}, and evaluate the attack performance of the framework within the physical environment.
\textcolor{red1}{As shown in \textcolor{red}{Figure}~\ref{fig:phy_vis}, our method achieves higher attack performance since neither of the two static PAEs caused the target detector to miss the detection. Additional physical experiments, accompanied by visualized figures and videos that demonstrate real-time adaptability to lightning changes and robustness against physical disturbances, are provided in the \textit{supplementary material}.}

\noindent
$\bullet$ \textbf{Confirmation of Physical Robustness:}
~\label{sec:robustness}
Since an accurate location, as the input of $\mathcal G$, may be difficult to obtain in applications, we perform tests with inaccurate inputs.
\textcolor{red}{During the training, the patch location is encoded into the generator's input $\mathbf{P}_X$, and patches' positioning parameters $\boldsymbol \theta$ are drawn from the distribution $p$ defined by Table~\ref{tab:transformation}. In Table~\ref{tab:location_roubustness}, \textit{Randomized} denotes the patch locations corresponding to $\boldsymbol \theta$ and $\mathbf P_X$ are  \emph{i.i.d.} parameters drawn from $p$, or the patch position and the generator's positioning input are unrelated,  and \textit{Accurate Input} denotes they share identical positioning parameters for each person attack test case.}
As the table shows, the attack performance did not drop much as we resampled the patch locations, which is attributed to the implicit regularization of DNN and the augmentation pipeline.
Moreover, the attack performance is still significantly higher than the baselines in the main performance evaluation.
This indicates that the model can leverage precise location information when available while still maintaining relatively high performance even with inaccurate location inputs.

\begin{table}[h]
    \centering
    \caption{Evaluation of the robustness against imprecise location inputs targeting Yolo-v5s on the COCO dataset. Randomized refers to re-sampling the attack locations.}
    \begin{tabular}{c|cc|cc}
    \toprule
 & \multicolumn{2}{c|}{Accurate Input}& \multicolumn{2}{c}{Randomized}\\
         Transformation&  $AP_{50}$\color{red}{↓}&  $AP_{01}$\color{red}{↓}&  $AP_{50}$\color{red}{↓}& $AP_{01}$\color{red}{↓}\\
 \midrule
         Base&  14.9\%&  26.9\%
&  16.3\%& 28.8\%
\\
         P+&  28.5\%&  44.7\%
&  30.5\%& 46.3\%
\\
         S+&  17.2\%&  31.8\%
&  20.7\%& 36.4\%
\\
         AF&  19.9\%&  36.4\%&  24.0\%& 41.0\%\\
 \bottomrule
    \end{tabular}
    \vspace{-0.25em}
    \label{tab:location_roubustness}
\end{table}

\noindent
\textcolor{red1}{$\bullet$ \textbf{Generalization in PAE Deployments:}}
\textcolor{red1}{
Unlike the generalization topic centered on the static data in classical tasks, the generalization of PAEs shall consider the manufacturing and resampling distortions~\cite{wang2024adversarialexamplesphysicalworld}.
To evaluate the generalizability, we test our trained generator in the 3D simulated physical deployment scenarios  ~\cite{hu2023physically,huang2025advreal}.
As shown in Figure~\ref{fig:diff_model_patches}, compared to our 2D training environment (e.g., AF), the patch transformations in the 3D simulation environment exhibit larger variations and are also more closely aligned with the physical world.
We evaluate the trained \THEMODEL\ with \textit{T-sea}, \textit{GAN-NAP}, and \textit{AdvReal} in both environments.
The background dataset for training is \textit{COCO, Inria, Inria, and nuScences}, respectively.
\textit{nuScences}~\cite{nuscenes2019} is a more realistic dataset for autonomous driving.
For the physical observation, only a coarse patch positioning in the 3D evaluation is provided as the input of \THEMODEL.}

\textcolor{red1}{
Table~\ref{tab:diff_pae_deployment} shows the key settings and results.
Besides achieving better attack performance in the 2D scenario, our \THEMODEL\ maintains the superiority over \textit{T-sea} and \textit{GAN-NAP} in the 3D scenario.
Therefore, by training with the \textit{Conditional-Uncertainty-Aligned Data Model} defined in section~\ref{sec:learning framework}, \THEMODEL\ can generalize to unseen physical deployment processes.
Although AdvReal achieved better results in the 3D scenario by training the static PAE with consistent 3D simulation, its performance on our small 2D-patch attack scenario does not improve.
Furthermore, the improved attack performance with better simulation (Base $\to$ AF) indicates the direction of scaling \THEMODEL\ with higher-fidelity physical world simulation systems.
}
\begin{table}[h]
    \centering
    \caption{\textcolor{red1}{Performance Evaluation on Different PAE Deployment Simulations. 2D-Base/AF denotes the settings in Table~\ref{tab:transformation}, and 3D-AdvReal denotes the AdvReal 3D simulation setting illustrated in Figure~\ref{fig:patch_affine}.}}
    \label{tab:diff_pae_deployment}
         \resizebox{\linewidth}{!}{
    \color{red1}
    \begin{tabular}{cc|cc}
    \toprule
         \multirowcell{2}{Attack\\ Method}&  \multirowcell{2}{Training\\ Settings}&  \multirowcell{2}{ASR-Inria\color{red}{↑}\\ 2D-Base}&\multirowcell{2}{ASR-nuScenes\color{red}{↑}\\ 3D-AdvReal}\\
         & & &\\
     \midrule
         \textit{GAN-NAP} &  2D-Base& 29.51\%& 14.94\%\\
         \textit{T-sea} &    2D (T-sea)& 30.34\%&  64.07\%\\
         \textit{AdvReal} & 2D+3D (AdvReal)& 28.73\%& \textbf{98.27}\%\\
 \THEMODEL& 2D-Base& \underline{55.30}\%&69.26\%\\
         \THEMODEL&  2D-AF&  \textbf{67.35}\%& \underline{93.29}\%\\
    \bottomrule
    \end{tabular}
    }
\end{table}
\begin{figure}[h]
    \centering
    \vspace{-1em}
    \includegraphics[width=0.6\linewidth]{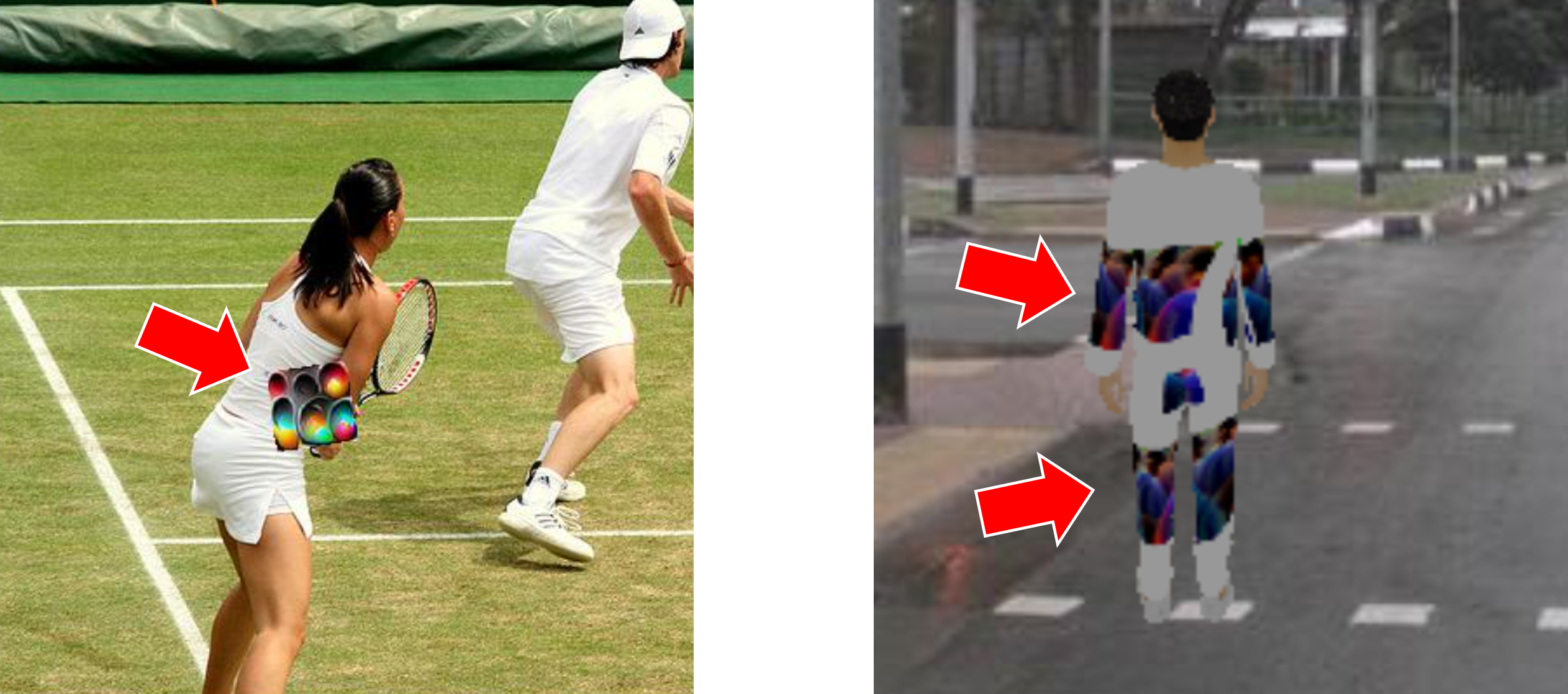}
    \caption{\textcolor{red1}{PAE deployment simulations. (Left: \textit{AF} defined in Table~\ref{tab:transformation}. Right: 3D simulation in AdvReal~\cite{huang2025advreal}).}}
    \label{fig:patch_affine}
    \vspace{-0.25em}
\end{figure}
\noindent

\noindent
$\bullet$ \textbf{Hyper-parameter Sensitivity:}
\textcolor{red}{The training of the generator shall be aligned with the stability requirement of the adversarial evaluation scenario.}
However, many generative models are sensitive to hyperparameters, especially GAN models, which harms their practicality. To make it clear, we present an investigation of the hyperparameters in our proposed model. The analysis includes the parameter $\alpha_0$ as the initial weight of invisibility and $\gamma$ in the regularization loss in Eq.~\ref{eq:reg}.
The results are shown in Table~\ref{tab:param_analysis}.
For regularization loss, we evaluate its weight with settings $\gamma \in \{10^{-4}, 10^{-2}, 1.0\}$, and the attack performance remains steady, indicating that it does not require excessive parameter tuning.
For the task scheduler, we evaluate it with different initial weights of invisibility $\alpha_0$, representing the variability in loss intensity between attack tasks, and both the auto-adjusted weight $\alpha_{final}$ and evaluation results remain similar, indicating the functionality of the skewness-based objective re-weighting mechanism that eliminates the free parameter and strike a consistent balance between attack and stealthiness.
Moreover, on the aggressiveness-distortion curves of the main performance experiments, the uniform and consistent data point distribution of our method across the different models also indicates it.
    
\begin{table}[h]
    \centering
       \caption{Analysis of parameter sensitivity. The weight of regularization $\gamma$ is insensitive. The sensitivity of the weight of the residual task is reduced by the \textit{skewness-aligned objective re-weighting} method.}
    \begin{tabular}{cc|c|cc|cc}
     
    \toprule
 \multicolumn{2}{c|}{Config}&  Adjusted&\multicolumn{2}{c|}{$\lambda = 1.0$} & \multicolumn{2}{c}{$\lambda = 0.5$}\\
         $\gamma$&$\alpha_0$&   $\alpha_{final}$&$AP_{50}$\color{red}{↓}&  SSIM\color{red}{↑}& $AP_{50}$\color{red}{↓}&SSIM\color{red}{↑}\\
         \midrule
         1.0  &100.0&   33.05&14.31\%&  0.9891
&  38.48\%&  0.9929\\
         0.01&100.0&   33.42&14.07\%
&  0.9891
&  
35.63\%&0.9928\\
         1e-4&100.0&   30.31&14.02\%
&  0.9890&  34.29\%&0.9928\\
         1.0&1.0&  35.77&14.90\%
& 0.9891
& 
31.99\%&0.9927\\
         1.0& 10.0 &  36.85 & 15.50\%
& 0.9892
& 32.86\%& 0.9925\\
         \bottomrule
    \end{tabular}

    \label{tab:param_analysis}
\end{table}

\subsection{Ablation Studies}

\begin{table}[t]
    \centering
      \caption{Ablation study of the auxiliary residual task. By collaborating with the attack training, the residual task consistently improves the attack performance compared to vanilla training (in columns w/o).}
      
    \begin{tabular}{cc|cc|cc}
    \toprule
 \multicolumn{2}{c|}{Configs} & \multicolumn{2}{c|}{+ Residual Task} & \multicolumn{2}{c}{w/o}\\
         Model & Trans. &  $AP_{50}$\color{red}{↓}&  $AP_{01}$\color{red}{↓}& $AP_{50}$\color{red}{↓}&$AP_{01}$\color{red}{↓}\\
         \midrule
         Yolov5-s& Base & 14.87\%&  26.94\%& 22.80\%& 41.01\%\\
         Yolov5-m& Base&   26.43\%&  36.00\%
&  42.02\%&  55.89\%
\\
         Yolov3-m & Base&  6.24\%&  10.18\%
&  8.23\%&11.57\%
\\
         FSCNN& Base&  26.33\%&  27.89\%
&  65.63\%&67.93\%\\
         Yolov8-n& Base& 26.04\%& 35.85\%
& 40.82\%&46.12\%
\\
         Yolov5-s & P+& 28.46\%& 44.68\%
& 32.96\%&49.40\%\\
         Yolov5-s & S+&  17.15\%&  31.76\%
&  22.39\%& 39.34\%\\
         Yolov5-s & AF&  19.87\%&  36.44\%
&  31.73\%&49.51\%
\\
         \bottomrule
    \end{tabular}
    \vspace{-0.5em}
    \label{tab:gradrescale}
\end{table}

\begin{table}[t]
    \centering
      \caption{
\textcolor{red}{Ablation study of the residual training. The improvement attributed to the proposed residual task is more significant than the residual neural architecture.}}
    \begin{tabular}{ccc|cc}
          \toprule
            Neural Arch.&  Res. Task& Model&  $AP_{50}$ \color{red}{↓} &  $AP_{01}$ \color{red}{↓}  \\
          \midrule
  \multirowcell{4}{w/o  All Skip\\Connections\\in $\mathrm{Dec}$}& \ding{55}
&v5& 18.57\%& 35.14\%\\
 & \ding{55} & v8& 49.36\%& 54.63\%\\ 
  & \ding{51}
&v5& 17.11\%& 32.96\%\\
 & \ding{51}& v8& 31.32\%&44.26\%\\

\midrule

  \multirowcell{4}{w/o  Skip-$\mathbf{Z}$ \\ Connections \\ in $\mathrm{Dec}$}& 

\ding{55}
& v5& 22.00\%& 40.50\%\\ 
  & \ding{55}
&v8& 44.04\%& 49.44\%\\ 
            &  \ding{51}
&v5&  17.66\%&34.22\%\\ 
           &  \ding{51}
&v8&  27.11\%& 36.19\%\\
\midrule
 \multirowcell{4}{All Skip\\Connections \\Added}& 
\ding{55}
& v5& 22.80\%& 41.01\%\\  
            &  \ding{55}
&v8&  40.82\%&  46.12\%\\
 & \ding{51}
& v5& \textbf{14.87\%} & \textbf{26.72\%} \\ 
  &  \ding{51}
&v8& \textbf{26.04}\%& \textbf{35.85\%} \\
 
 \bottomrule
     \end{tabular}
    \label{tab:abl_resnet}
    \vspace{-1em}
\end{table}

\noindent
$\bullet$ \textbf{Auxiliary Residual Task:}
Ablation experiments across different target models and different patch transformations are conducted to evaluate the residual task.
$\lambda$ is set to 1 during tests.
As shown in Figure~\ref{tab:gradrescale}, the constructed auxiliary residual task consistently improves the attack performance of dynamic PAE, demonstrating the necessity of such construction in the dynamic PAE optimization process.
We recognize that the improvement is more significant in large victim models that are hard to attack, including \textit{Yolo-v5m} and \textit{Faster-RCNN}, which may be because larger models have more diverse vulnerabilities, allowing for the discovery of more conditional solutions through the improved optimization process.
We also recognize that in the setting of \textit{P+}, where the spatial transformation in attack injection $\oplus$ is more diverse, the drop ratio of $AP$ is relatively low.
This may be because the higher uncertainty makes the information available for adaptation limited, with less room for performance improvement.
It is worth exploring how to construct more effective residual tasks or alter attack tasks to enhance attack capabilities further.

\textcolor{red}{
Introducing residual connections in neural networks is a classic method in deep learning to address optimization issues. 
We further conduct experiments to compare residual connections and the residual task.
The \THEMODEL\ framework adopts the image decoder architecture of \textit{BigGAN}~\cite{brock2018large}, which combines the additional \textit{Z-skip} residual connections with the \textit{ResNet} architecture. We perform the ablation study on \textit{Z-skip} and other skip connections in \textit{ResNet}, along with the proposed residual task, in dynamic PAE generation. As shown in Table~\ref{tab:abl_resnet}, the residual task significantly improves performance across different configurations. The improvement from residual connections is not significant, and negative effects exist in several settings. Therefore, \textbf{the residual task technique is more important than the residual connection in training the dynamic PAE generator.}
}

 \begin{table}[t]
     \centering
          \caption{Ablation of latent regularization. Encouraging the codewords of PAEs to be located in the center of the representation space achieves better attack performance and training stability.}
     \begin{tabular}{ccc|cc}
          \toprule
            $\boldsymbol{z}_{adv}$&  Aux. Task& Model&  $AP_{50}$ \color{red}{↓} &  $AP_{01}$ \color{red}{↓}  \\
          \midrule
  \multirowcell{3}{w/o\\ regularization}
& \ding{55}
&v5& 22.39\%& 40.66\% \\
 & \ding{55} & v8& 39.96\% & 45.33\% \\ 
  & \ding{51}
&v5/8& \multicolumn{2}{c}{--- (Gradient Exploded)}\\

\midrule

  \multirowcell{4}{$\sim \mathcal N (0, I)$ \\ (VAE prior)}
& 

\ding{55}
& v5& 18.97\%& 35.16\%
\\ 
  & \ding{55}
&v8& 44.95\%& 50.53\%\\ 
            &  \ding{51}
&v5&  17.66\%&34.00\%\\ 
           &  \ding{51}
&v8&  \multicolumn{2}{c}{--- (Gradient Exploded)}\\
\midrule
 \multirowcell{4}{$l_2$-norm \\ (\textit{\THEMODEL}  \\ Applied)} & 
\ding{55}
& v5& 22.80\%& 41.01\%\\  
            &  \ding{55}
&v8&  40.82\%&  46.12\%\\
 & \ding{51}
& v5& \textbf{14.87\%} & \textbf{26.72\%} \\ 
  &  \ding{51}
&v8& \textbf{26.04}\%& \textbf{35.85\%} \\
 
 \bottomrule
     \end{tabular}

     \label{abl:encodings}
 \end{table}
\begin{table}
     \centering
          \caption{\textcolor{red}{The ablation of skewness alignment. $\lambda_{\textrm{Atk}}$ is set by the deployment scenario. By controlling the skewness $\varsigma$ of the stealth distribution of generated PAEs in training, better extreme attack performance and custom stealthiness configuration can be achieved during deployment. \textcolor{red2}{Stealthiness is evaluated using SSIM (higher is better).}}}
     \resizebox{\linewidth}{!}{
  
     \begin{tabular}{lc|cc|cc|cc}
     
          \toprule
  && \multicolumn{2}{c|}{$ \varsigma_\text{target} = -1$}& \multicolumn{2}{c|}{$ \varsigma_\text{target} = +1$}&\multicolumn{2}{c}{$ \varsigma_\text{target} = 0$}\\
             &$\lambda_\text{Atk}$&  $AP_{50}$ \color{red}{↓} &\color{red2}{SSIM↑}&  $AP_{50}$ \color{red}{↓} &\color{red2}{SSIM↑}&$AP_{50}$ \color{red}{↓} &\color{red2}{SSIM↑}\\
      
 \midrule
   \multirow{5}{*}{\rotatebox[origin=c]{90}{Yolo-v5s}} &0.0& 77.9\%&0.99385&  77.8\%&0.99394& 78.2\%&0.99387
\\
  &0.2& 75.9\%&0.99380&  50.7\%&0.99326& 74.4\%&0.99385
\\

   &0.5& 50.3\%&0.99329&  24.2\%&0.99206& 43.8\%&0.99304
\\ 
   &0.8& 25.7\%&0.99205&  16.5\%&0.99031& 22.5\%&0.99172
\\  
             &1.0&  16.8\%&0.98913&\textbf{14.9}\%&0.98904& \textbf{14.9}\%&0.98905
\\
             \midrule
             
   \multirow{5}{*}{\rotatebox[origin=c]{90}{Yolo-v8n}} &0.0
& 85.7\%
&	
0.99385&  85.8\%
&0.99398
& 85.7\%
&0.99375
\\
  &0.2
& 78.0\%
&0.99374&  77.9\%
&0.99353
&

79.5\%
&0.99371
\\

   &0.5
& 68.4\%
&0.99347&  42.6\%
&0.99169
& 58.8\%
&0.99293
\\ 
   &0.8
& 34.4\%
&0.99120&  34.0\%
&0.99048
&

33.1\%
&0.99118
\\  
             &1.0&  29.4\%
&0.98918&  32.9\%
&0.98949
& \textbf{26.0\%}&0.98925
\\
            \bottomrule
     \end{tabular}
}
     \label{abl:skewness}
\end{table}

\noindent
$\bullet$ \textbf{Latent Regularization:}
\textcolor{red}{Due to the unique characteristic of \THEMODEL\ Training, additional regularization is applied on the latent encoding to prevent gradient explosion.}
We compare the differences in regularization in terms of the attack performance.
As shown in Table~\ref{abl:encodings}, the center-regularized $\boldsymbol z_{adv}$ achieves the optimal performance among different settings and solves the problem of gradient explosion when learning diverse PAEs through the auxiliary residual task.
Note that the PAE decoder $\mathrm{Dec}$ degenerates to a single static PAE without the auxiliary residual task.
We noticed classical VAE regularization still suffers the infinite gradient in the model of Yolo-v8n and the attack performance on Yolo-v5 is lower, indicating the widely-used Gaussian prior might not be a proper selection for the latent of explored PAEs.
\textcolor{red}{We further analyze the gradient explosion problem in supplementary materials.}

\noindent
\textcolor{red}{
$\bullet$ \textbf{Skewness Alignment:} The proposed \textit{skewness-aligned objective re-weighting} controls the stealthiness distribution of PAEs when different $\lambda_\mathrm{Atk}$ is set, and it is also beneficial for improving the attack performance. We perform the ablation study by altering the target skewness in Eq~\ref{eq:gradalpha}. The results are shown in Table~\ref{abl:skewness}, indicating that (1) neutral setting of target skewness ($ \varsigma_\text{target} = 0$) achieves the highest extreme attack performance when $\lambda_\mathrm{Atk} = 1$. This shows that balancing the attack and residual task is beneficial for exploring more aggressive PAE patterns, and the precise balance indicator is worth further study. (2) After deployment, the user can adjust the stealthiness by setting different $\lambda_\mathrm{Atk}$ as the input of the generator $\mathcal{G}$, and the generator can select the proper PAE pattern based on $\lambda_\mathrm{Atk}$}

\subsection{\textcolor{red1}{Applications and Defenses}}
\textcolor{red1}{This section discusses the use of the proposed attack method in adversarial test data generation, along with potential defense strategies.}
\label{sec:application}

\begin{figure}[h]
    \centering
    \vspace{-1em}    \includegraphics[width=\linewidth]{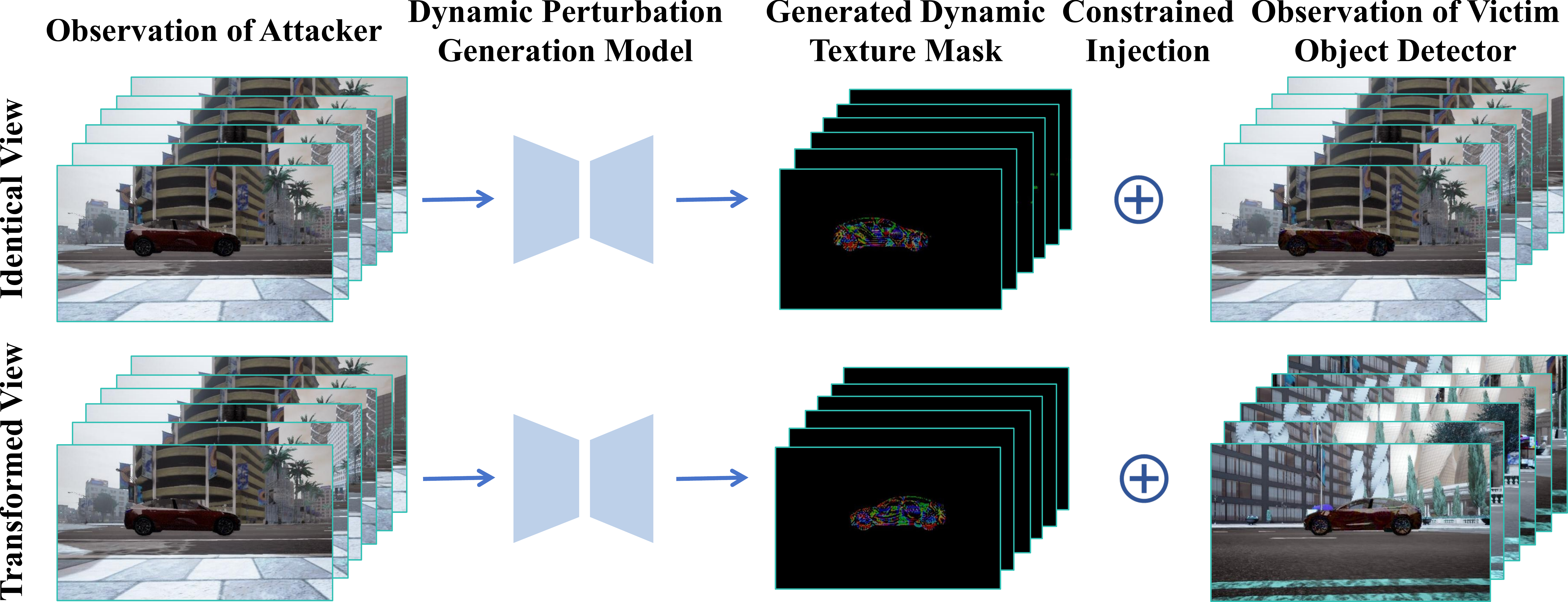}
    \vspace{-1em}
    \caption{Demo of end-to-end simulated adversarial evaluation of autonomous driving perception systems.}
    \vspace{-0.5em}
    \label{fig:carla}
\end{figure}

\noindent
$\bullet$ \textbf{Application on Simulated Testing:}
To evaluate the applicability of the proposed model in the simulated testing, we obtain the simulation results from different perspectives in the \textit{CARLA} autonomous driving simulation platform~\cite{carla}, divide them into training and testing datasets, and apply the adversarial example as a perturbation of the vehicle's texture.
Supplement to the digital environment benchmark, we compare the performance with the GAN-based perturbation generation framework \textit{AdvGAN}, since both our model and \textit{AdvGAN} are learning-based and real-time.
The adversarial attack scenario is shown in Figure~\ref{fig:carla}. In the \textit{Identical View} scenario, the attacker's observation, as the generative attacker's input, is identical to the victim model. In the \textit{Transformed View} scenario, the view of the victim model is on the other side. Following the \textit{Conditional-Uncertainty-Aligned Data Model}, we formulate the distribution $\Omega$ in Eq.~\ref{eq:omega} that generates $\mathbf{P}_\mathrm X$ as:
\begin{equation}
\textcolor{red}{
    \Omega(\boldsymbol\theta, \mathbf X) = [\mathcal{T}(\mathbf X, \mathrm{Cam}_{X}), \mathcal{T}(\boldsymbol m_\theta, \mathrm{Cam}_{X})], 
}
\vspace{-0.25em}
\end{equation}
where $\mathcal{T}$ is the view transformation performed by the simulation engine, $\mathrm{Cam}_{X}$ denotes the sampled camera state.

\textcolor{red2}{The results are in Table~\ref{tab:sim_eval}, showing that besides patch attacks, our model could also achieve strong perturbation attack performance, outperforming \textit{AdvGAN}.
To measure the stealthiness of the attack more accurately, we measure the SSIM similarity in addition to the $l_{\infty}$ constraint.
Interestingly, our method performs better in the transformed view than in the identical view when $\varepsilon = 1\%$, which may be attributed to the increased generalization due to the augmented transformation $\mathcal{T}$. This suggests our method is more suitable for complex physical attacks.}
\textcolor{red2}{We further test the \textit{robustness} of the perturbation under extreme weather conditions by adding a snow \& rain simulation provided by the \textit{Kornia library}~\cite{eriba2019kornia} during the test. The results show that (1) only adding extreme weather leads to a \textit{mild} decrease in detection precision. (2) \THEMODEL\ maintains \textit{significant} attack performance even after the perturbation is transformed by the simulation, and the performance may even increase due to the collaboration of multiple distortions. These results demonstrate \THEMODEL's cross-environment robustness, indicating its strong extensibility.
}

\begin{table}[h]
    \caption{Results on \textit{CARLA} simulation. The illustration of different view settings is shown in Figure~\ref{fig:carla}. \textcolor{red2}{\textit{Ex} indicates integrating extreme weather simulation.}}
    \centering
    \resizebox{\linewidth}{!}{
        \begin{tabular}{cc|ccc|ccc}
        \toprule
          &&  \multicolumn{3}{c|}{Identical View} &  \multicolumn{3}{c}{Transformed View}\\
          $\varepsilon$&Method&  AP$_{50}$\color{red}{↓}&  AP$_{01}$\color{red}{↓}& SSIM\color{red}{↑} &  AP$_{50}$\color{red}{↓}& AP$_{01}$\color{red}{↓}& SSIM\color{red}{↑} \\ 
        \midrule
          0\%&Clean& 43.23\% & 46.56\%   &  1.0000& 51.08\% & 52.56\%& 1.0000\\
 & \color{red2}Ex. weather& \color{red2}35.35\%& \color{red2}43.88\%& \color{red2}0.7833& \color{red2}44.35\%& \color{red2}50.35\%& \color{red2}0.7475\\
          \midrule
 1\%& \textit{AdvGAN} 
& 29.67\%  & 38.40\%& 0.9983 & 34.30\% & 44.17\% & \textbf{0.9994}\\
 1\%& \textit{\THEMODEL} & 5.35\%& 12.39\%& \textbf{0.9989} &\textbf{ 3.31\%}&\textbf{ 7.53\%}&\textbf{0.9994}\\
 
\color{red2} 1\%& \color{red2} \textit{\THEMODEL}-Ex & \color{red2}\textbf{1.56\%}& \color{red2}\textbf{10.43\%}& \color{red2}0.7823& \color{red2}3.95\%& \color{red2}8.11\%& \color{red2}0.7466\\
          \midrule
          5\%& \textit{AdvGAN} & 6.0e-5 & 7.0e-3& 0.9888& 2.9e-3&  1.1e-2& 0.9928    \\
          5\%&\textit{\THEMODEL} & $<$\textbf{1e-6}& 2.5e-6&  \textbf{0.9969}&   $<$\textbf{1e-6}& 8.5e-6& \textbf{0.9931} \\
          \color{red2} 5\%&\color{red2}  \textit{\THEMODEL}-Ex & \color{red2}$<$\textbf{1e-6}& \color{red2} $<$\textbf{1e-6}& \color{red2} 0.7752& \color{red2}  $<$\textbf{1e-6}& \color{red2}$<$\textbf{1e-6}& \color{red2} 0.7418\\
         \bottomrule
    \end{tabular}
    }

    \label{tab:sim_eval}
\end{table}

\noindent
\textcolor{red1}{$\bullet$ \textbf{Countermeasure on the Attack:}}
\textcolor{red1}{
Since this study aims to generate PAEs, which might threaten real-world deployed AI applications, it is necessary to provide possible countermeasures to make it responsible. We provide the experimental results by a typical adversarial defense approach, \emph{i.e.}, adversarial patch detection. Note that we believe that the adversarial training methods could also serve as a useful defense against dynamic adversarial patch attacks in principle, but we leave it as an open problem.}

\textcolor{red1}{Specifically, the adversarial patch detection methods try to train an object detector to locate the dangerous patches inside AEs. Therefore, we construct a test dataset based on COCO-val with PAEs injected.
We then introduce a SOTA method, namely, NAPGuard~\cite{Wu24NAPGuard}, as a defender with 2 different evaluation settings, \emph{i.e.}, the patch generated by attacking Yolo-v5s and Yolo-v8n.
We select the \textit{Ultralytics Yolov5s} as the backbone for NAPGuard.
For our framework, we provide 3 parameter settings for the evaluation, to wit, setting $\lambda = 0.5, 0.6, 1.0$ (\THEMODEL$_{A}$ in Table \ref{tab:main_results} adopts $\lambda=1.0$).
As for metrics, we report the AP$_{50}$ of the object detector and the adversarial patch detector (denoted as ``Attack AP$_{50}$'' \& ``Defense AP$_{50}$''), respectively.
The evaluation results can be witnessed in Table~\ref{tab:countermeasure}, where we can observe the following conclusions: 
\ding{182} compared with the \textit{GAN-NAP} and \textit{T-sea}, the NAPGuard could effectively detect \THEMODEL$_{\lambda=1.0}$, \emph{i.e.}, its Defense AP$_{50}$ achieves 0.9534, while that of the \textit{GAN-NAP} and \textit{T-sea} is respectively 0.6597 and 0.7099. That means the most aggressive settings($\lambda=1.0$) might be detected more easily, which shows that its greatest potential social risk is in control.
\ding{183} The tendency of the evaluation results in different settings shows that our proposed method could sacrifice aggressiveness for stealthiness, indicating its dynamic attacking ability in the physical world. Therefore, it is worth studying targeted defense approaches against this typical attack.
\ding{184} For different target models, \emph{i.e.}, Yolo-v5s and Yolo-v8n, our framework shows defense-evading ability, which may be attributed to the generalizability of the defense method, urging the study of more effective and efficient countermeasures.}

\begin{table}[h]
    \centering
    \caption{\textcolor{red1}{Performance of countermeasure methods.}} 
        \resizebox{\linewidth}{!}{
    \color{red1}
    \begin{tabular}{c|cc|cc}
    \toprule
 &   \multicolumn{2}{c|}{Attack AP$_{50}$\color{red}{↓}}&\multicolumn{2}{c}{Defense AP$_{50}$\color{red}{↑}}  \\
         Method&    Yolo-v5s&Yolo-v8n&Yolo-v5s&Yolo-v8n\\
    \midrule

 \textit{GAN-NAP}&   0.4014
&0.7103&0.6597& 0.9769\\
 \textit{T-sea}&   0.4825
&0.6043&0.7099& 0.9787\\
\THEMODEL$_{\lambda=0.5}$&     0.4375
&0.5875&0.6709&  0.2024\\\
\THEMODEL$_{\lambda=0.6}$&    0.3630
&0.4653&0.7537& 0.3232\\
 \THEMODEL$_{\lambda=1.0}$&   0.1487&0.2604&0.8307& 0.6609\\
    \bottomrule
    \end{tabular}
    }
    \label{tab:countermeasure}
    \vspace{-1em}
\end{table}

\section{Related Works}

\noindent
\textbf{Digital Adversarial Example Generation:} 
Classical type of AE research concentrated on generating $l_p-$constrained examples and developed techniques based on model gradient~\cite{DBLP:journals/corr/GoodfellowSS14}, projected gradient descent~\cite{madry2017towards}, SGD optimizer~\cite{carlini2017towards} and neural network~\cite{baluja2017adversarial}.
Further study focused on constructing more complex attack scenarios, such as the black-box attack~\cite{chen2017zoo}, data-manifold constrained AEs~\cite{song18aegen}, and the transfer attacks among models and scenarios.

\noindent
\textbf{Physical Adversarial Attacks:}
Based on the paradigm of optimizing static PAEs, research focuses on simulating the physical world~\cite{jan2019connecting} and the target AI system~\cite{zhu2023understanding}, and modeling PAEs under naturalness constrains~\cite{Hu2021NaturalisticPA} or new attack mediums~\cite{yang2023towards}.
Recent works began to focus on the challenge of generating dynamic PAEs.
\cite{guesmi2024dap} proposes the \textit{Dynamic Adversarial Patch} on the dynamically changing clothes, while the PAE data itself is still static.
\cite{chahe2023dynamic} proposes the adaptive PAE by manually clustering the attack scenario and optimizing the static PAE for each cluster, but it is not suitable for the open world.
\cite{sun2024embodied} proposes a physical adversarial attack by controlling a dynamic laser beam in the simulation using reinforcement learning.
Nevertheless, it is only capable of modeling a limited number of laser states but not the general space of PAE.
\textcolor{red}{
\cite{li2025uvattack} proposes a computational efficient optimization and rendering pipeline for 3D physical attacks. However, the dynamic adaptation to the current scene is left as an open problem.
}

\noindent
\textbf{Generative Models and Generative Adversarial Attacks:}
Motivated by learning representations~\cite{bengio13replearning}, classical generative neural networks adapt generative learning methods to train a model that maps the space of real data $\mathcal{X}$ to a latent space $\mathcal{Z}$, which has better mapping modeling and sampling efficiency.
With refined learning tasks and neural network construction~\cite{song19scoregen, tian2024visual}, generative models are able to generate complex data, \emph{e.g.}, model the relation of texts and natural images~\cite{textguidedgen}, molecular dynamics~\cite{zhang2018deep}, protein structure~\cite{morehead2024geometry} and images and corresponding digital perturbation AEs~\cite{xiao2018generating}.
In terms of adversarial attacks, NN-based generators have been applied in improving perturbation optimization~\cite{xiao2018generating}, generating naturalistic AEs by modeling the data manifold~\cite{song18aegen,liu2019perceptual,Hu2021NaturalisticPA, DiffPGD} and probabilistic modeling perturbations for black-box attacks~\cite{fei24blackbox}.
However, only \textit{unconditional} \& \textit{static} PAEs have been proposed under the generative framework~\cite{Hu2021NaturalisticPA, hu2022adversarial, DiffPGD}.
Except for some specific adversarial attack mediums, \emph{e.g.}, audio~\cite{mia22realtimevoice} and face subspace~\cite{yang2023towards}, the physical-digital gap~\cite{DBLP:conf/iclr/KurakinGB17a} is small, the randomized transformation may not be necessary, and therefore the digital AE generation network may be effective.
As an important step forward, we bridge the general gap between scene-aware dynamic PAE and generative NNs.

\noindent
\textbf{Multidisciplinary Optimization Techniques:}
Similar optimization problems also occurred in the application of neural networks~\cite{raissi2019physics} that learn complex tasks in the field of AI4Science, \emph{e.g.}, approximating complex fields~\cite{krishnapriyan2021characterizing} and training on noisy measurements~\cite{yuan2022pinn}.
We believe the relevant research will benefit dynamic PAE generation and vice versa.
In the field of reinforcement learning, research has been conducted on the problem of exploring the action space and utilizing inaccurate feedback~\cite{pmlr-v80-colas18a}.
However, the general dynamic PAE generation problem is defined in the complex state space, such as a patch with 10K+ pixels, which is intractable for reinforcement models.
Recently, language-driven models have been adapted for optimization~\cite{yang2024large}, but there is still a gap between language and low-level AE data~\cite{carlini2024aligned}, in which generative embedding models shall fill.

\section{\textcolor{red}{Discussions}}

\noindent
\textcolor{red}{
\textbf{Applications of the Framework:} Although we only constructed a prototype system for physical attacks and test cases in a simulated environment during the experiment, our framework is general-purpose.
Its applications include visual robustness evaluation of autonomous systems and building camouflage systems for interference or protecting key targets.
Specifically, when evaluating the visual robustness, DynamicPAE provides guidance and efficiently improves the coverage of the physical-world-aligned adversarial testing.
By incorporating existing technologies, such as dynamic materials (e.g., color-changing costumes~\cite{kim2021biomimetic}) and real-time 3D holography~\cite{shi2021towards}, DynamicPAE facilitates the construction of intelligent camouflage systems.
}

\noindent
\textcolor{red}{
\textbf{Limitations and Future Work:} Our work focuses on the key generating algorithm for PAEs. However, further exploration is necessary to investigate additional properties, including the transferability of PAE between target models, the robustness of PAE on flexible materials, and the efficiency of the generator on more resource-limited devices.
To achieve these properties, we can integrate existing black-box attacks, 3D simulations, and model light-weighting methods with the generative training method provided by DynamicPAE.
While the efficacy of DynamicPAE in challenging learning tasks and its adoption of end-to-end modular design suggest the potential for effective integration, the characteristics of specific downstream tasks are still worth investigation.
}
\textcolor{red1}{
For the generative training technology, given the complexity of the PAE generator training task, more accurate models are needed to characterize the hardness of the exploration problem. The incorporation of technologies from related fields, as previously discussed, holds the potential for further enhancing PAE exploration.
}

\section{Conclusion}
This study focuses on the dynamic physical adversarial examples (PAEs), a fundamental and largely unaddressed vulnerability of deep learning models in applications.
A highly effective and versatile method, \THEMODEL, is proposed for generating real-time scene-aware PAE, offering a significant advancement.
Extensive experimental results demonstrate that the proposed method exhibits superior dynamic attack performance in open and complex scenarios.

Future research directions can focus on technical improvements to the patch representation and training mechanism, exploring the application of the proposed framework, and facilitating the developed techniques in broader domains.
In terms of application, although we have presented the preliminary prototype in our experiments, work on generating adversarial test and training data and exploring real-world red team attacks for specific tasks is still worthy of further investigation.
We believe that our study has the broader potential to benefit the optimization of defense-insensitive adversarial attacks and noisy open-world tasks, and the model for the hardness of PAE optimization can bring insights to the defense of PAEs.

\section*{Acknowledgments}
This work is supported in part by Zhongguancun Laboratory, in part by State Key Laboratory of Complex \& Critical Software Environment (CCSE), and in part by the National Natural Science Foundation of China under Grant 62506347.
\bibliography{sample-base}

\vspace{-4em}

\begin{IEEEbiography}
[{\includegraphics[width=1in,height=1.25in,clip,keepaspectratio]{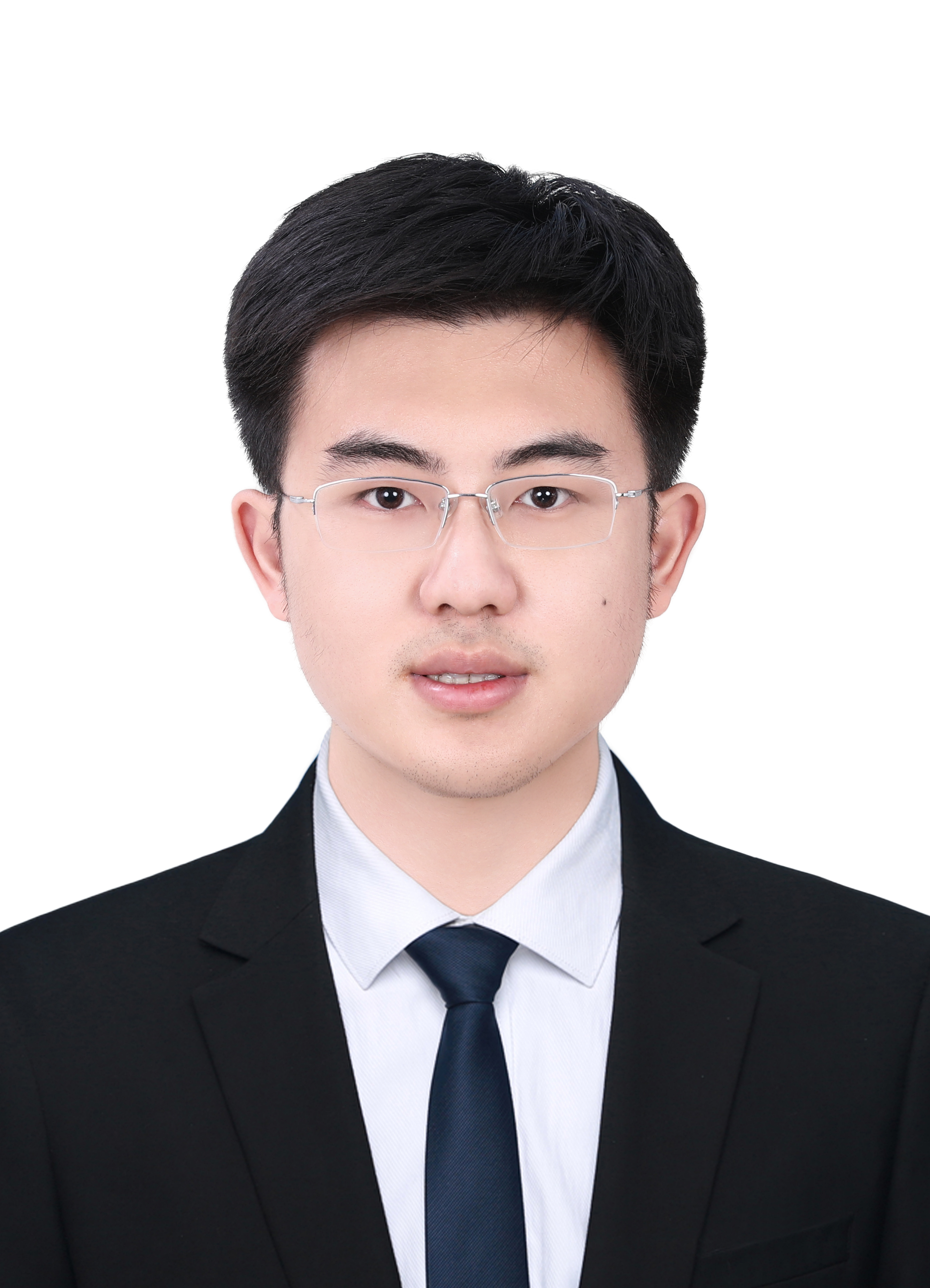}}]
{Jin Hu} is currently a Ph.D. Student at the State Key Laboratory of Complex \& Critical Software Environment, Beihang University and Zhongguancun Lab.
His research interests include adversarial machine learning, generative modeling, and trustworthy AI.

\end{IEEEbiography}

\vspace{-3em}

\begin{IEEEbiography}
[{\includegraphics[width=1in,height=1.25in,clip,keepaspectratio]{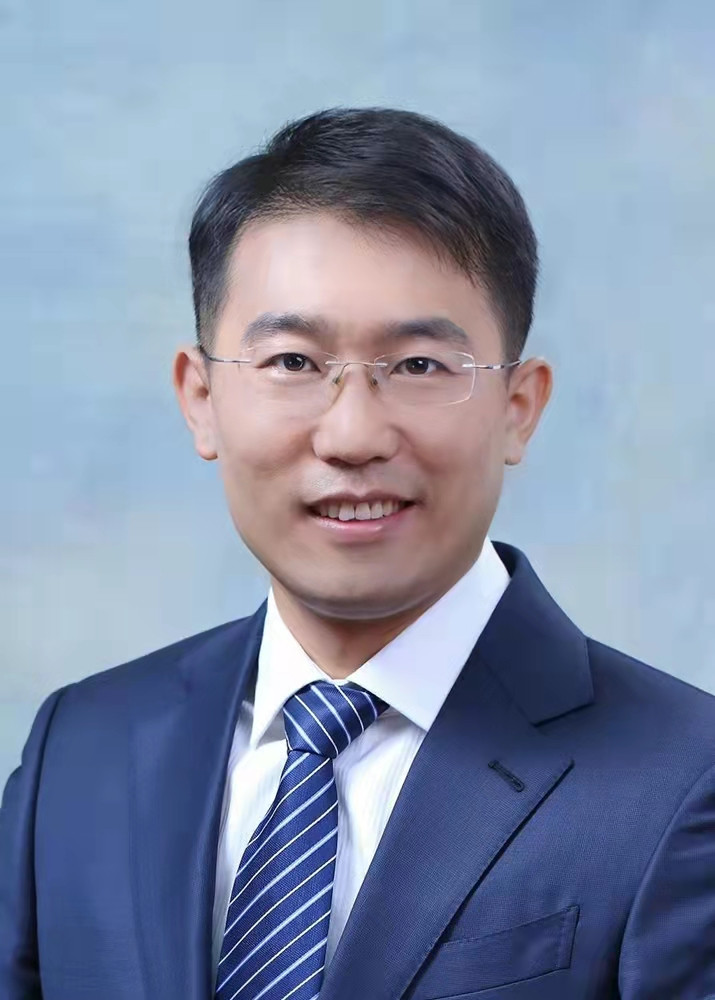}}]{Xianglong Liu} is a Full Professor in School of Computer Science and Engineering at Beihang University. He received BS and Ph.D degrees under supervision of Prof. Wei Li, and visited DVMM Lab, Columbia University as a joint Ph.D student supervised by Prof. Shih-Fu Chang. His research interests include fast visual computing (e.g., large-scale search/understanding) and robust deep learning (e.g., network quantization, adversarial attack/defense, few shot learning). He received NSFC Excellent Young Scientists Fund, and was selected into 2019 Beijing Nova Program, MSRA StarTrack Program, and 2015 CCF Young Talents Development Program.

\end{IEEEbiography}


\begin{IEEEbiography}
[{\includegraphics[width=1in,height=1.25in,clip,keepaspectratio]{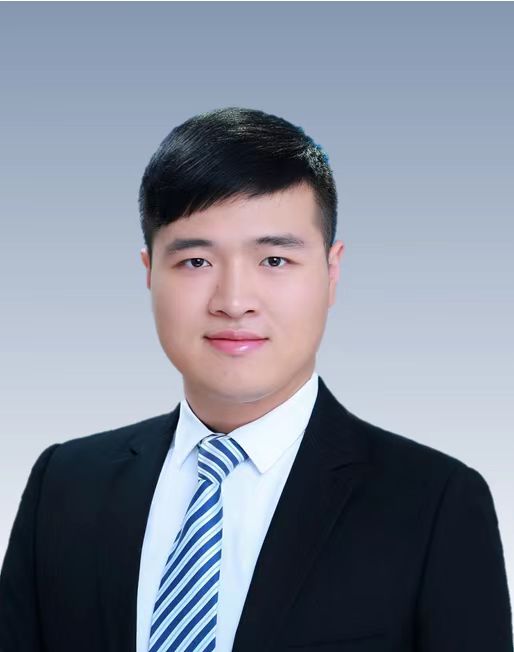}}]{Jiakai Wang} is now a Research Scientist in Zhongguancun Laboratory, Beijing, China. He received the Ph.D. degree in 2022 from Beihang University (Summa Cum Laude), supervised by Prof. Wei Li and Prof. Xianglong Liu. Before that, he obtained his BSc degree in 2018 from Beihang University (Summa Cum Laude). His research interests are Trustworthy AI in Computer Vision (mainly) and Multimodal Machine Learning, including Physical Adversarial Attacks and Defense and Security of Practical AI.
\end{IEEEbiography}

\vspace{-3em}
\begin{IEEEbiography}
[{\includegraphics[width=1in,height=1.25in,clip,keepaspectratio]{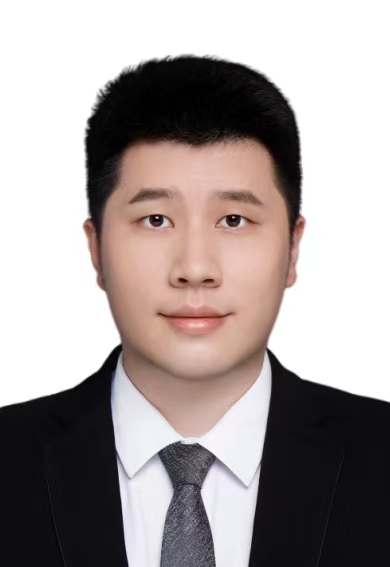}}]
{Junkai Zhang} is currently a research intern at State Key Laboratory of Complex \& Critical Software Environment, Beihang University, and a BEng student at the joint cultivation program of Beihang University and Beijing University of Technology. His research interests include computational linguistics and trustworthy AI.
\end{IEEEbiography}

\vspace{-3em}
\begin{IEEEbiography}
[{\includegraphics[width=1in,height=1.25in,clip,keepaspectratio]{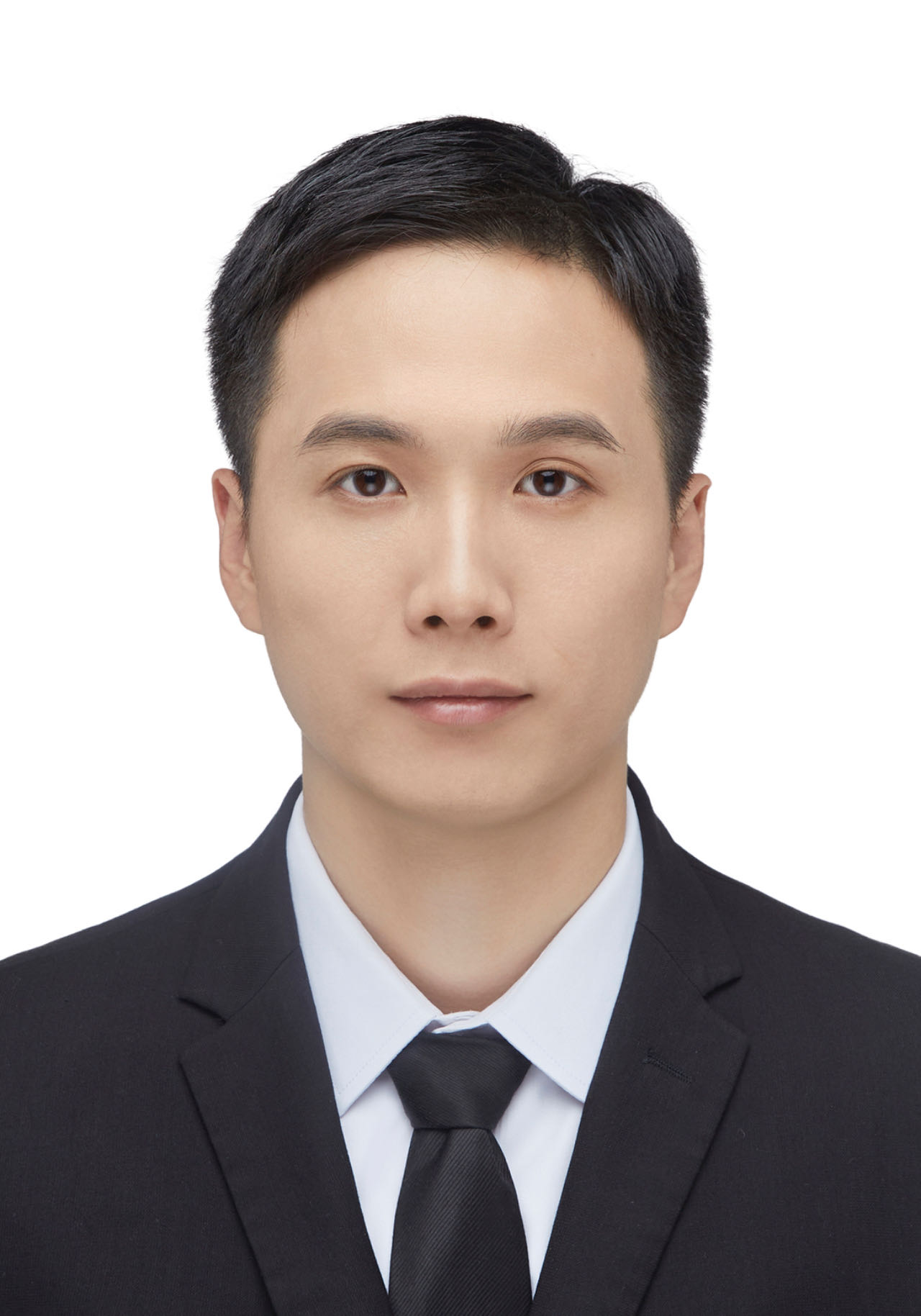}}]
{Xianqi Yang} is now an engineer at Zhongguancun Laboratory, Beijing, China. He received the master's degree in 2023 from Beihang University, supervised by Prof. Qing Gao and Prof. Kexin Liu. His research interests are blockchain(mainly) and industrial security.
\end{IEEEbiography}

\vspace{-3em}
\begin{IEEEbiography}
[{\includegraphics[width=1in,height=1.25in,clip,keepaspectratio]{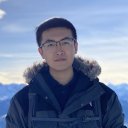}}]
{Haotong Qin} is a Postdoctoral Researcher at the Center for Project-Based Learning (PBL), ETH Zürich, Switzerland, working with PD Dr. Michele Magno. Previously, he received my Ph.D. degree from the State Key Laboratory of Complex \& Critical Software Environment (SKLCCSE), Beihang University in 2024, supervised by Prof. Wei Li and Prof. Xianglong Liu. He was a visiting PhD student at Computer Vision Lab, ETH Zürich. He obtained B.E. degree from SCSE, Beihang University in 2019. He interned at ByteDance AI Lab, Tencent WXG, and Microsoft Research Asia. His research interest broadly includes efficient deep learning.
\end{IEEEbiography}

\vspace{-3em}

\begin{IEEEbiography}
[{\includegraphics[width=1in,height=1.25in,clip,keepaspectratio]{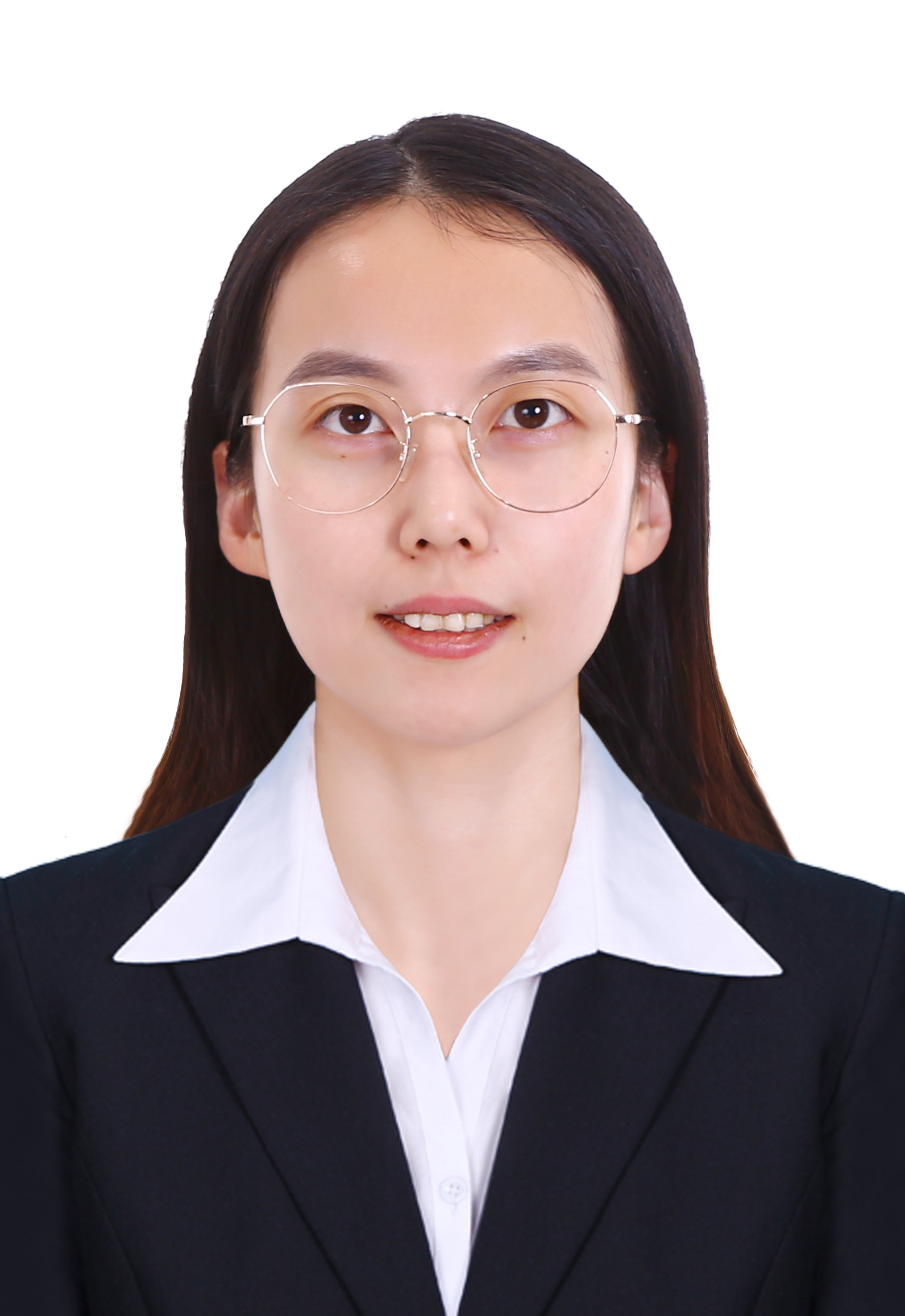}}]
    {Yuqing Ma} received the Ph.D degree in 2021 from Beihang University, China. She was a postdoctoral fellow at Beihang University from 2021 to 2023. She is currently an associate professor with the Institute of Artificial Intelligence at Beihang University. Her current research interests include computer vision, few-shot learning and open world learning. She has published nearly 30 research papers at top-tier conferences and journals.
\end{IEEEbiography}

\vspace{-3em}

\begin{IEEEbiography}
[{\includegraphics[width=1in,height=1.25in,clip,keepaspectratio]{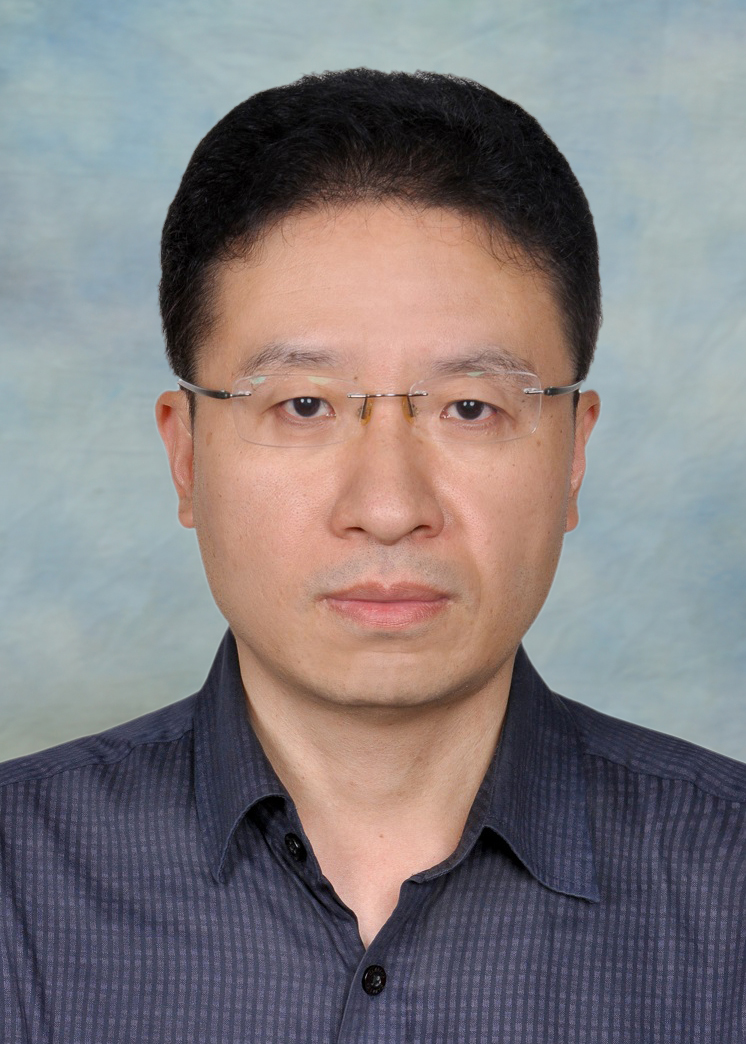}}]
{Ke Xu} received the BE, ME, and PhD degrees from Beihang University, in 1993, 1996, and 2000, respectively. He is a professor at the School of Computer Science and Engineering, Beihang University, China. His current research interests include phase transitions in NP-Complete problems, algorithm design, computational complexity, big spatio-temporal data analytics, crowdsourcing, and crowd intelligence. 
\end{IEEEbiography}

\end{document}